\documentclass[10pt,journal,compsoc]{IEEEtran}
\ifCLASSOPTIONcompsoc
  \usepackage[nocompress]{cite}
\else
  \usepackage{cite}
\fi
\ifCLASSINFOpdf
\else
\fi

\usepackage[symbol]{footmisc}
\usepackage[affil-it]{authblk}
\usepackage[english]{babel}
\usepackage{blindtext}
\usepackage[affil-it]{authblk}
\usepackage[english]{babel}
\usepackage{blindtext}
\usepackage{times}
\usepackage{epsfig}
\usepackage{graphicx}
\usepackage{amsmath}
\usepackage{amssymb}
\usepackage{algorithm}
\usepackage{algorithmic}
\usepackage{setspace}
\usepackage{url}
\usepackage{multirow}
\usepackage{bbm}

\usepackage{colortbl}
\definecolor{mygray}{gray}{.9}
\definecolor{mypink}{rgb}{.99,.91,.95}
\definecolor{mycyan}{cmyk}{.3,0,0,0}

\usepackage{color}
\usepackage{xcolor}
\definecolor{citecolor}{HTML}{0071bc} 
\definecolor{SeaGreen4}{RGB}{0,205,102} 
\definecolor{SlateBlue}{RGB}{106,90,205} 
\definecolor{DarkRed}{RGB}{178,34,34} 
\usepackage{newunicodechar}
\newunicodechar{≤}{\ensuremath{\leq}}
\usepackage[colorlinks, linkcolor=red,  anchorcolor=blue, citecolor=citecolor]{hyperref}
\DeclareUnicodeCharacter{2217}{\ensuremath{*}}

\begin{document}
\title{Pedestrian Attribute Recognition: A Survey}

\author{Xiao Wang, \emph{Member, IEEE}, Shaofei Zheng, Rui Yang, Aihua Zheng, Zhe Chen, \emph{Member, IEEE}, Jin Tang, and Bin Luo, \emph{Senior Member, IEEE}
\thanks{Xiao Wang, Shaofei Zheng, Rui Yang, Aihua Zheng, Jin Tang, and Bin Luo are with the School of Computer Science and Technology, Anhui University, Hefei 230601, China. Xiao Wang is also with Peng Cheng Laboratory, Shenzhen, China. \\ 
Zhe Chen is with UBTECH Sydney AI Centre, SIT, FEIT, University of Sydney, Australia. \\ 
Email: \{shaofeizheng, yangruiahu\}@foxmail.com, \{xiaowang, luobin, tangjin\}@ahu.edu.cn.} 
}

\markboth{Updated on August, 2023}%
{Shell \MakeLowercase{\textit{et al.}}: Bare Demo of IEEEtran.cls for Computer Society Journals}

\IEEEtitleabstractindextext{
\begin{abstract}
Recognizing pedestrian attributes is an important task in the computer vision community due to it plays an important role in video surveillance. Many algorithms have been proposed to handle this task. The goal of this paper is to review existing works using traditional methods or based on deep learning networks. Firstly, we introduce the background of pedestrian attribute recognition (PAR, for short), including the fundamental concepts of pedestrian attributes and corresponding challenges. Secondly, we introduce existing benchmarks, including popular datasets and evaluation criteria. Thirdly, we analyze the concept of multi-task learning and multi-label learning and also explain the relations between these two learning algorithms and pedestrian attribute recognition. We also review some popular network architectures which have been widely applied in the deep learning community.  Fourthly, we analyze popular solutions for this task, such as attributes group, part-based, etc. Fifthly, we show some applications that take pedestrian attributes into consideration and achieve better performance. Finally, we summarize this paper and give several possible research directions for pedestrian attribute recognition. We continuously update the following GitHub to keep tracking the most cutting-edge related works on pedestrian attribute recognition~\url{https://github.com/wangxiao5791509/Pedestrian-Attribute-Recognition-Paper-List}
\end{abstract}

}

\maketitle
\IEEEdisplaynontitleabstractindextext
\IEEEpeerreviewmaketitle

\begin{figure}[!htb]
\center
\includegraphics[width=3.5in]{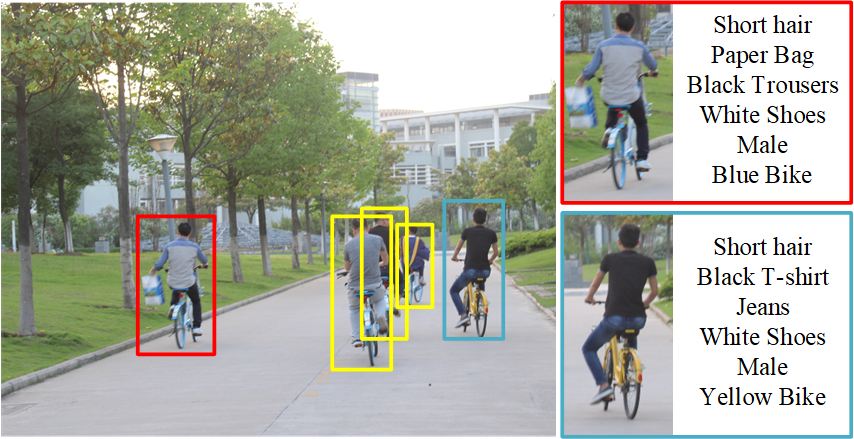}
\caption{Pedestrian attribute recognition is a key element in video surveillance. Given a person's image, pedestrian attribute recognition aims to predict a group of attributes to describe the characteristic of this person from a pre-defined attribute list. For example, the attributes of a man in the red bounding box are: short hair, with paper bag, black trousers, etc.} 
\label{fontImage}
\end{figure}

\IEEEraisesectionheading{\section{Introduction}\label{sec:introduction}}
\IEEEPARstart{P}{edestrian} attributes, are humanly searchable semantic descriptions and can be used as soft-biometrics in visual surveillance, with applications in person re-identification, face verification, and human identification. Pedestrian Attribute Recognition (PAR) aims at mining the attributes of target people when given a person image, as shown in Figure~\ref{fontImage}. Different from low-level features, such as HOG, LBP, or deep features, attributes can be seen as high-level semantic information which is more robust to viewpoint changes and viewing condition diversity. Hence, many tasks in computer vision integrate the attribute information into their algorithms to achieve better performance, such as person re-ID, and person detection. Although many works have been proposed on this topic, however, pedestrian attribute recognition is still an unsolved problem due to challenging factors, such as viewpoint change, low illumination, low resolution, and so on.

Traditional pedestrian attribute recognition methods usually focus on developing robust feature representation from the perspectives of hand-crafted features, powerful classifiers, or attribute relations. Some milestones including HOG~\cite{dalal2005histograms}, SIFT~\cite{lowe2004distinctive}, SVM~\cite{chang2011libsvm} or CRF model~\cite{lafferty2001conditional}. However, the reports on large-scale benchmark evaluations suggest that the performance of these traditional algorithms is far from the requirement of realistic applications. 
	
Over the past several years, deep learning has achieved an impressive performance due to its success in automatic feature extraction using multi-layer nonlinear transformation, especially in computer vision, speech recognition, and natural language processing. Several deep learning-based attribute recognition algorithms have been proposed based on these breakthroughs, such as \cite{sudowe2015person, acprli2015DeepMAR, abdulnabi2015multitaskCNN, bourdev2011describing, joo2013humanRAD, zhang2014panda, zhu2015multilabelCNN, Gkioxari_2015_ICCV, BMVC2016_ARAP, diba2016deepcamp, ICMEli2018pose, li2016humanDHContexts, liulocalizationBMVC2018, liu2017hydraplus, sarfraz2017deep, sarafianos2018deep, guo2017humanPRL, wang2016cnnRNN, wang2017attribute, ijcaizhao2018grouping, liu2018sequenceJCM, zhou2017weakly, mmhe2017adaptively, DCSA2012, park2018attributeAOG, vsgraaai2019, sudowe2016patchit, lu2017fullyafs, fabbri2017generative}.

Although so many papers have been proposed, until now, there exists no work to make a detailed survey, comprehensive evaluation, and insightful analysis of these attribute recognition algorithms. In this paper, we summarize existing works on pedestrian attribute recognition, including traditional methods and popular deep learning-based algorithms, to better understand this direction and help other researchers quickly capture the main pipeline and latest research frontier. Specifically speaking, we attempt to address the following several important issues: 
\textbf{Firstly}, what is the connection and difference between traditional and deep-learning-based pedestrian attribute recognition algorithms? We analyze traditional and deep learning-based algorithms from different classification rules, such as part-based, group-based, or end-to-end learning; 
\textbf{Secondly}, how do the pedestrian attributes help other related computer vision tasks? We also review some person attributes guided computer vision tasks, such as person re-identification, object detection, and person tracking, to fully demonstrate the effectiveness and wide applications in many other related tasks; 
\textbf{Thirdly}, how to better leverage deep networks for pedestrian attribute recognition and what is the future direction of the development of attribute recognition? By evaluating existing person attribute recognition algorithms and some top-ranked baseline methods, we make some useful conclusions and provide some possible research directions.

\begin{figure}[t]
\center
\includegraphics[width=3.5in]{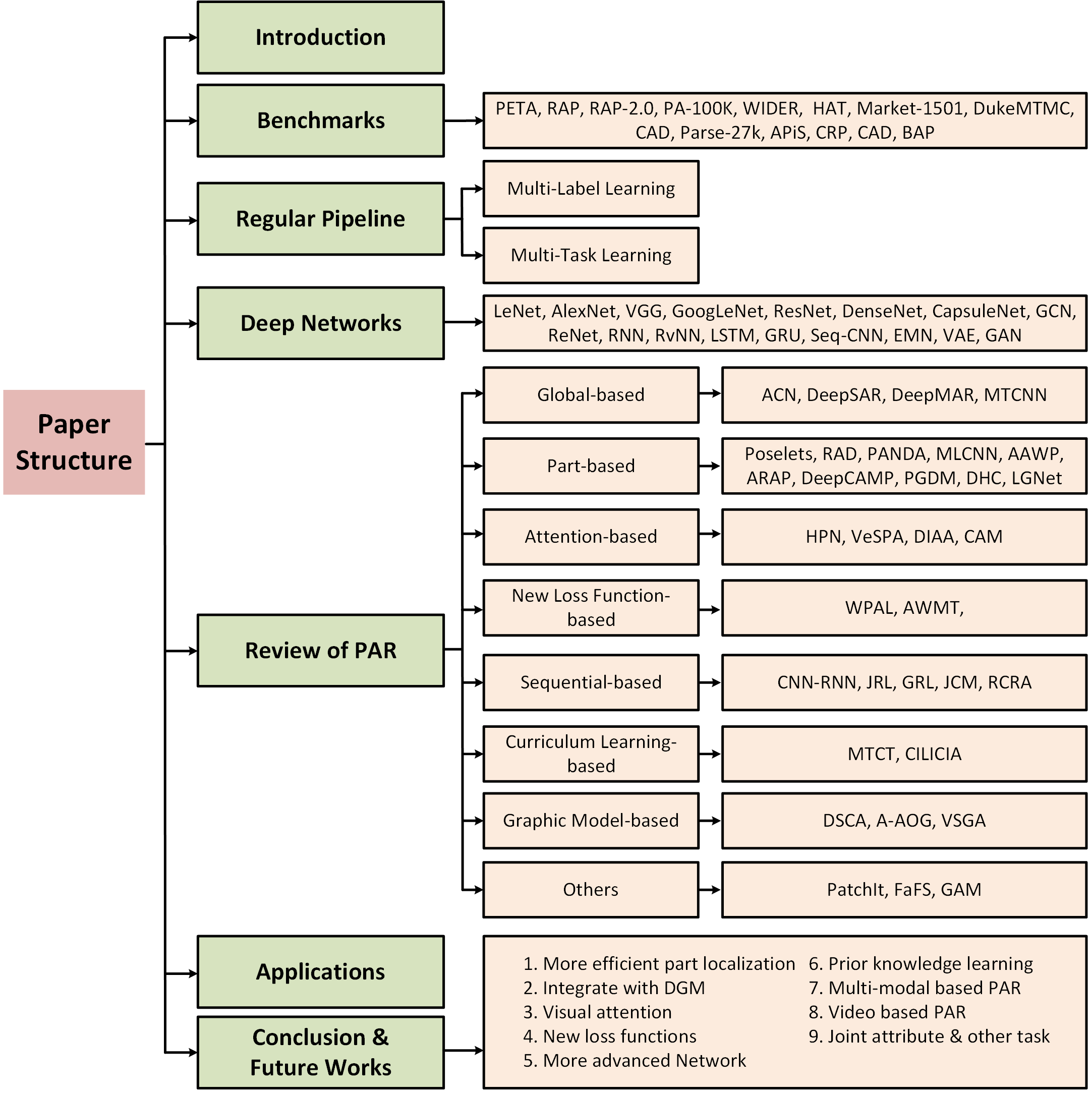}
\caption{The structure of this survey on pedestrian attribute recognition. }
\label{structure}
\end{figure}

\textbf{The rest of this paper is organized as follows:} In Section~\ref{problemformulationChallenges}, we briefly introduce the problem formulation of pedestrian attribute recognition and some challenging factors. In Section~\ref{Benchmarks}, we list some popular benchmarks for this task and report the corresponding recognition performance of baseline methods. After that, we review existing methods in Section~\ref{traditionalMethods} and Section~\ref{DeepAlgorithms} from different categories. We also divide these methods into eight domains, including global-based, local parts based, visual attention based, sequential prediction based, newly designed loss function based, curriculum learning based, graphic model based, and other algorithms. In Section~\ref{someApplications}, we show some examples that can integrate attributes into consideration and achieve better performance. Finally, we summarize this paper and provide some possible research points for this direction in Section~\ref{conAndFuture}. To better visualize and understand the structure of this paper, we give a figure as shown in Figure~\ref{structure}.

\section{Problem Formulation and Challenges} \label{problemformulationChallenges}

\begin{figure}[!htb]
\center
\includegraphics[width=3.3in]{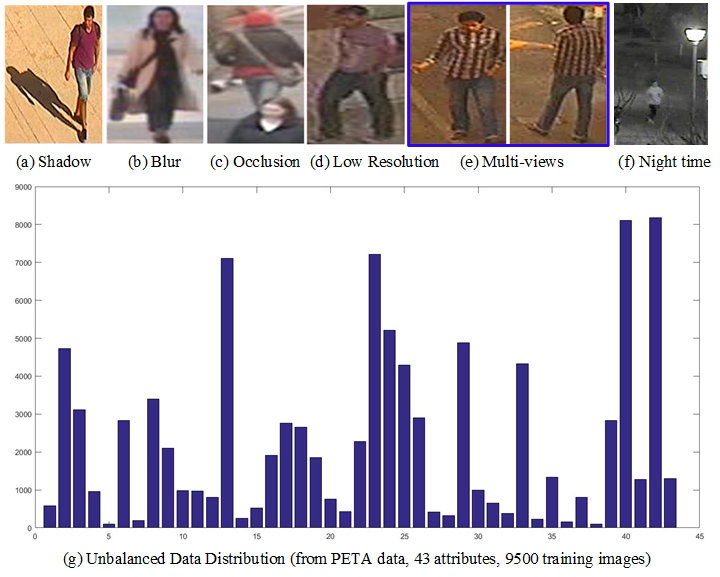}
\caption{Some challenging factors in pedestrian attribute recognition. }
\label{challenges}
\end{figure}

Given a person image $\mathcal{I}$, pedestrian attribute recognition aims to predict a group of attributes $a_i$ to describe the characteristics of this person from a pre-defined attribute list $\mathcal{A} = \{a_1, a_2, ..., a_L\}$, as shown in Figure~\ref{fontImage}. This task can be handled in different ways, such as multi-label classification, and binary classification. Also,  many algorithms and benchmarks have been proposed. However, this task is still challenging due to the large intra-class variations in attribute categories (appearance diversity and appearance ambiguity~\cite{deng2014pedestrian}). As shown in Figure~\ref{challenges}, we list challenging factors that may obviously influence the final recognized performance: 

\noindent 
\textbf{Multi-views.} The images taken from different angles by the camera lead to viewpoint issues for many computer vision tasks. The body of a human is not rigid, which further makes the person attribute recognition more complicated. 

\noindent 
\textbf{Occlusion.} Part of the human body are occluded by other person or things will increase the difficulty of a person's attribute recognition. Because the pixel values introduced by the occluded parts may make the model confused and lead to wrong predictions.

\noindent 
\textbf{Unbalanced Data Distribution.} Each person has different attributes, hence, the number of attributes is variable which leads to unbalanced data distribution. It is widely acknowledged that current machine learning algorithms may not perform optimally on these datasets. 

\noindent 
\textbf{Low Resolution.} In practical scenarios, the resolution of images is rather low due to the high-quality cameras being rather expensive. Hence, the person's attribute recognition needs to be done in this environment. 

\noindent 
\textbf{Illumination.} The images may taken from any time in 24 hours. Hence, the light condition may be different at different times. The shadow may also taken in the person's images and the images taken from nighttime may be totally ineffective. 

\noindent 
\textbf{Blur.} When a person is moving, the images taken by the camera may blur. How to recognize a person's attributes correctly in this situation is a very challenging task.


\section{Benchmarks}~\label{Benchmarks} 
Unlike other tasks in computer vision, for pedestrian attribute recognition, the annotation of the dataset contains many labels at different levels. For example, hairstyle and color, hat, glass, etc. are seen as specific low-level attributes and correspond to different areas of the images, while some attributes are abstract concepts, such as gender, orientation, and age, which do not correspond to certain regions, we consider these attributes as high-level attributes. Furthermore, human attribute recognition is generally severely affected by environmental or contextual factors, such as viewpoints, occlusions and body parts, In order to facilitate the study, some datasets provide perspective, parts bounding box, occlusion, etc. 

By reviewing related works in recent years, we have found and summarized several datasets which are used to research pedestrian attribute recognition, including PETA~\cite{deng2014pedestrian}, RAP~\cite{li2016richly}, RAP-2.0~\cite{li2019richlyRAP2}, PA-100K~\cite{liu2017hydraplus}, WIDER~\cite{li2016humanDHContexts}, Market-1501~\cite{lin2017improving}, DukeMTMC~\cite{lin2017improving}, Clothing Attributes Dataset~\cite{chen2012describing}, PARSE-27K~\cite{sudowe2015person}~\cite{Sudowe16BMVC}, APiS~\cite{zhu2013pedestrian}, HAT~\cite{sharma2011learning}, Berkeley-Attributes of People dataset~\cite{bourdev2011describing} and CRP dataset~\cite{hall2015fineCRP}. Details of the attribute labels of these datasets can be found on our project page due to the limited space in this paper.

\begin{table*}[htp!]
\center
\scriptsize
\newcommand{\tabincell}[2]{\begin{tabular}{@{}#1@{}}#2\end{tabular}}
\caption{An overview of datasets proposed for pedestrian attribute recognition (\# denotes the number of corresponding item).}	
\label{PersondatasetList}
\resizebox{\textwidth}{!}{
\begin{tabular}{l|l|l|l|l}
\hline
\hline
\textbf{Dataset}  	&\textbf{\#Pedestrians} &\textbf{\#Attributes}  &\textbf{Source}  &\textbf{Project Page}	\\
\hline 
PETA \cite{deng2014pedestrian}				& 19000 & \tabincell{l}{61 binary and 4 \\multi-class attributes} & outdoor \& indoor 	&	\href{http://mmlab.ie.cuhk.edu.hk/projects/PETA.html}{URL}	\\ 
\hline
RAP	\cite{li2016richly} 							& 41585 & \tabincell{l}{69 binary and 3 \\multi-class attributes} & indoor	&	\href{http://rap.idealtest.org/}{URL}	\\ 
\hline
RAP-2.0 \cite{li2019richlyRAP2}				& 84928 & \tabincell{l}{69 binary and 3 \\multi-class attributes} & indoor	&	\href{https://drive.google.com/file/d/1hoPIB5NJKf3YGMvLFZnIYG5JDcZTxHph/view}{URL}	\\ 
\hline
PA-100K \cite{liu2017hydraplus}				& 100000 & 26 binary attributes & outdoor 	&	\href{https://drive.google.com/drive/folders/0B5_Ra3JsEOyOUlhKM0VPZ1ZWR2M}{URL}	\\ 
\hline
WIDER \cite{li2016humanDHContexts}	& 13789 & 14 binary attributes &  WIDER images\cite{7298768}	&	\href{http://mmlab.ie.cuhk.edu.hk/projects/WIDERAttribute.html}{URL}	\\ 
\hline
Market-1501 \cite{lin2017improving}					& 32668 & \tabincell{l}{26 binary and 1\\ multi-class attributes} &  outdoor 	&	\href{https://github.com/vana77/Market-1501_Attribute}{URL}	\\ 
\hline
DukeMTMC \cite{lin2017improving}	& 34183 & 23 binary attributes &  outdoor 	&	\href{https://github.com/vana77/DukeMTMC-attribute}{URL}	\\ 
\hline
PARSE-27K	\cite{sudowe2015person, Sudowe16BMVC}  			& 27000 & \tabincell{l}{8 binary and 2\\ multi-class orientation attributes} & outdoor	&	\href{https://www.vision.rwth-aachen.de/page/parse27k}{URL}	\\ 
\hline
APiS	 \cite{zhu2013pedestrian} 			& 3661 & \tabincell{l}{11 binary and 2 \\multi-class attributes}  & \tabincell{l}{KITTI\cite{geiger2012we} , \\CBCL Street Scenes\cite{bileschi2006streetscenes},\\ INRIA\cite{dalal2005histograms} and SVS} 	&	\href{http://www.cbsr.ia.ac.cn/english/APiS-1.0-Database.html}{URL}	\\ 
\hline
HAT	\cite{sharma2011learning} 			& 9344 & \tabincell{l}{27 binary attributes} & image site Flickr	&	\href{https://jurie.users.greyc.fr/datasets/hat.html}{URL}	\\ 
\hline
CRP	\cite{hall2015fineCRP}					& 27454 & \tabincell{l}{1 binary attributes and \\13 multi-class attributes} &  outdoor  &	\href{http://www.vision.caltech.edu/~dhall/projects/CRP/}{URL}	\\ 
\hline
CAD	\cite{chen2012describing}							& 1856 & \tabincell{l}{23 binary attributes and 3 \\multi-class attributes} &  \tabincell{l}{image site \\Sartorialist\footnote{\url{http://www.thesartorialist.com}} and Flickr}	&	\href{https://purl.stanford.edu/tb980qz1002}{URL}	\\ 
\hline
BAP~\cite{bourdev2011describing} 		& 8035 & 9 binary attributes & \tabincell{l}{H3D\cite{bourdev2009poselets} dataset \\ PASCAL VOC 2010\cite{pascal-voc-2010}}	&	\href{https://www2.eecs.berkeley.edu/Research/Projects/CS/vision/shape/poselets/}{URL}	\\ 
\hline
MARS-Attributes~\cite{chen2019videoPAR} &20,478 tracklets (1,261 people) &20 attributes &MARS	&  \href{https://github.com/yuange250/MARS-Attribute}{URL} \\ 
\hline 
DukeMTMC-VID-Attributes~\cite{chen2019videoPAR} &4,832 tracklets (1,402 people) &18 attributes &DukeMTMC-VID	&  \href{https://github.com/yuange250/MARS-Attribute}{URL} \\ 
\hline 
UAV-Human~\cite{li2021uavHUMAN} &22,263    &7 attributes    &outdoor    &\href{https://github.com/SUTDCV/UAV-Human}{URL}    \\ 
\hline 
UPAR~\cite{specker2023upar} &-    &40 attributes    &PA100K, PETA, RAPv2, and Market1501    &\href{https://github.com/speckean/upar_challenge}{URL}    \\ 
\hline 
\end{tabular}} 
\end{table*}

\subsection{Datasets}

\noindent \textbf{PETA dataset}~\cite{deng2014pedestrian} is constructed from 10 publicly available small-scale datasets used to research person re-identification. This dataset consists of 19000 images, with resolution ranging from 17$\times$39 to 169$\times$365 pixels. Those 19000 images include 8705 persons, each annotated with 61 binary and 4 multi-class attributes, and it is randomly partitioned into 9,500 for training, 1,900 for verification, and 7,600 for testing. One notable limitation is that the samples of one person in PETA dataset are only annotated once by randomly picking one exemplar image and therefore share the same annotated attributes even though some of them might not be visible and some other attributes are ignored. Although this method is reasonable to a certain extent, it is not very suitable for visual perception.

\noindent \textbf{PARSE27K}~\cite{sudowe2015person, Sudowe16BMVC} dataset derives from 8 video sequences of varying lengths taken by a moving camera in a city environment. Every $15^{th}$ frame of the sequences was processed by the DPM pedestrian detector~\cite{felzenszwalb2010object}. It contains 27,000 pedestrians and has a training (50\%), validation (25\%), and test (25\%) split. Each sample is manually annotated with 10 attribute labels which include 8 binary attributes~\textit{such as is male?}, \textit{has bag on left shoulder?}, and two orientation attributes with 4 and 8 discretizations. In the PARSE-27K dataset, an attribute is called \textit{N/A} label when it can not be decided because of occlusion, image boundaries, or any other reason.

\noindent \textbf{RAP}~\cite{li2016richly} dataset is collected from real indoor surveillance scenarios and 26 cameras are selected to acquire images, it contains 41585 samples with resolution ranging from 36$\times$92 to 344$\times$554.  Specifically, there are 33268 images for training and the remains for testing. 72 fine-grained attributes (69 binary attributes and 3 multi-class attributes) are assigned to each image of this dataset. Three environmental and contextual factors, i.e., viewpoints, occlusion styles, and body parts, are explicitly annotated. Six parts (spatial-temporal information, whole body attributes, accessories, postures and actions, occlusion, and parts attributes) are considered for attribute annotations.

\noindent \textbf{RAP-2.0}~\cite{li2019richlyRAP2} dataset comes from a realistic High-Definition ($1280 \times 720$) surveillance network at an indoor shopping mall and all images are captured by 25 cameras scenes. This dataset contains 84928 images (2589 person identities) with resolution ranging from 33 $\times$ 81 to 415 $\times$ 583. Every image in this dataset has six types of labels, which are the same as the RAP dataset and have 72 attribute labels. All samples were divided into three parts, of which 50957 for training, 16986 for validation, and 16985 for testing.

\noindent \textbf{HAT}~\cite{sharma2011learning} dataset originates in the popular image sharing site Flickr. This dataset includes 9344 samples, of which 3500, 3500, and 2344 images are for training, validation, and testing respectively. Every image in this dataset has 27 attributes and shows a considerable variation in pose (standing, sitting, running, turned back, etc.), different ages (baby, teen, young, middle-aged, elderly, etc.), wearing different clothes (tee-shirt, suits, beachwear, shorts, etc.) and accessories (sunglasses, bag, etc.).

\noindent \textbf{APiS}~\cite{zhu2013pedestrian} dataset comes from four sources: KITTI \cite{geiger2012we} dataset, CBCL Street Scenes \cite{bileschi2006streetscenes} (CBCLSS for short) dataset, INRIA \cite{dalal2005histograms} database and SVS dataset (Surveillance Video Sequences at a train station). A pedestrian detection approach \cite{yan2012multi} was performed to automatically locate candidate pedestrian regions, false positives, and those too-small images were deleted, and finally, 3661 images were obtained each image larger than 90 pixels in height and 35 pixels in width. Each image is labeled with 11 binary attributes, such as \textit{male}, \textit{long hair}, and 2 multi-value attributes, including upper body color and lower body color. The \textit{ambiguous} indicates whether the corresponding attribute is uncertain or not. This dataset is separated into 5 equal-sized subsets, the performance is evaluated with 5-fold cross-validation, and the 5 results from the 5 folds are further averaged to produce a single performance report.

\noindent \textbf{Berkeley-Attributes of People (BAP)} \cite{bourdev2011describing} dataset comes from the H3D~\cite{bourdev2009poselets} dataset and the PASCAL VOC 2010 \cite{pascal-voc-2010} training and validation datasets for the person category, the low-resolution versions used in PASCAL are replaced by the full resolution equivalents on Flickr. All images were split into 2003 training, 2010 validation, and 4022 test images by ensuring that no cropped images of different sets come from the same source image and by maintaining a balanced distribution of the H3D and PASCAL images in each set. Each image was labeled with nine attributes. A label was considered as ground truth if at least 4 of the 5 annotators agreed on the value of the label. When an attribute is not determined to be present or absent, it is annotated as "unspecified".

\noindent \textbf{PA-100K} \cite{liu2017hydraplus} dataset is constructed by images captured from 598 real outdoor surveillance cameras, it includes 100000 pedestrian images with resolution ranging from 50 $\times$ 100 to 758 $\times$ 454 and is to-date the largest dataset for pedestrian attribute recognition. The whole dataset is randomly split into training, validation, and test sets with a ratio of 8:1:1. Every image in this dataset was labeled by 26 attributes, and the label is either 0 or 1, indicating the presence or absence of corresponding attributes respectively.

\noindent \textbf{WIDER} \cite{li2016humanDHContexts} dataset comes from the 50574 WIDER images~\cite{7298768} that usually contain many people and huge human variations, a total of 13789 images were selected. Each image was annotated with a bounding box but no more than 20 people (with top resolutions) in a crowd image, resulting in 57524 boxes in total and 4+ boxes per image on average. Each person is labeled with 14 distinct attributes, resulting in a total of 805336 labels. This dataset was split into 5509 training, 1362 validation, and 6918 test images.

\noindent \textbf{Market1501-attribute} \cite{DBLPAttIdent2017} dataset is collected by six cameras in front of a supermarket in Tsinghua University. There are 1,501 identities and 32,668 annotated bounding boxes in this dataset. Each annotated identity is present in at least two cameras. This dataset was split into 751 training and 750 test identities, corresponding to 12936 and 19732 images respectively. The attributes are annotated at the identity level, every image in this dataset is annotated with 27 attributes. Note that although there are 7 and 8 attributes for lower-body clothing and upper-body clothing, only one color is labeled as yes for an identity.

\noindent \textbf{DukeMTMC-attribute} \cite{DBLPAttIdent2017} is collected in Duke University. There are 1812 identities and 34183 annotated bounding boxes in the DukeMTMC-attribute dataset. This dataset contains 702 identities for training and 1110 identities for testing, corresponding to 16522 and 17661 images respectively. The attributes are annotated at the identity level, every image in this dataset is annotated with 23 attributes.

\noindent \textbf{CRP} \cite{hall2015fineCRP} is captured in the wild from a moving vehicle. The CRP dataset contains 7 videos and 27454 pedestrian bounding boxes. Each pedestrian was labeled with four types of attributes, age (5 classes), sex (2 classes), weight (3 classes), and clothing type (4 classes). This dataset is split into a training/validation set containing 4 videos, with the remaining 3 videos forming the test set.

\noindent \textbf{Clothing Attributes Dataset (CAD)} \cite{DCSA2012} was collected from Sartorialist~\footnote{\url{http://www.thesartorialist.com}} and Flickr. The dataset contains 1856 images, with 26 ground truth clothing attributes collected using Amazon Mechanical Turk. All labels are arranged in the order from image 1 to 1856. Some label entries are ``NaN", meaning the 6 human workers cannot reach an agreement on this clothing attribute. There are 26 attributes in total, including 23 binary-class attributes (6 for pattern, 11 for color, and 6 miscellaneous attributes) and 3 multi-class attributes (sleeve length, neckline shape, and clothing category). This dataset was split by leave-1-out for training and testing.

\noindent \textbf{UAV-Human}~\cite{li2021uavHUMAN} dataset is recorded using a UAV which involves 7 attributes, and 22,263 person images. This dataset also provides annotations for action recognition.

\noindent \textbf{UPAR}~\cite{specker2023upar} dataset is built by combining existing four well-known person attribute recognition datasets, including PA100K, PETA, RAPv2, and Market1501. 40 important binary attributes over 12 attribute categories are considered when providing 3,3M additional annotations for this dataset.

\noindent \textbf{Video-based PAR datasets.} There are two video-based PAR datasets proposed by Chen et al. in the year 2019, including MARS-Attributes and DukeMTMC-VID-Attributes~\cite{chen2019videoPAR}. As shown in Table~\ref{PersondatasetList}, the MARS-Attributes contains 20,478 tracklets from 1,261 people, which involves 20 attributes. The DukeMTMC-VID-Attributes contains 4,832 tracklets from 1,402 people and covers 18 attributes.

\subsection{Evaluation Criteria}
Zhu et al.\cite{zhu2013pedestrian} evaluate the performance of each attribute classification with the Receiver Operating Characteristic (ROC) and the Area Under the average ROC Curve (AUC) which are calculated by two indicators, the recall rate and false positive rate. The recall rate is the fraction of the correctly detected positives over the total amount of positive samples, and the false positive rate is the fraction of the misclassified negatives out of the whole negative samples. At various threshold settings, a ROC curve can be drawn by plotting the recall rate vs. the false positive rate. In addition, the Area Under the average ROC Curve (AUC) is also used by Zhu et al.~\cite{zhu2013pedestrian} so as to make a clearer performance comparison.

Deng at al.\cite{deng2014pedestrian} adopt the mean Accuracy (mA) to evaluate the attribute recognition algorithms. For each attribute, mA calculates the classification accuracy of positive and negative samples respectively and then gets their average values as the recognition result for the attribute. Finally, a recognition rate is obtained by taking average overall attributes. The evaluation criterion can be calculated through the following formula: 

\begin{equation}
\label{eq::objective}
\begin{aligned}
\mathit{mA} = \dfrac{1}{2N}\sum_{i=1}^{L}(\dfrac{TP_{i}}{P_{i}}+\dfrac{TN_{i}}{N_{i}})
\end{aligned}
\end{equation}
where $L$ is the number of attributes. $\mathit{TP_{i}}$ and $\mathit{TN_{i}}$ are the number of correctly predicted positive and negative examples respectively, $\mathit{P_{i}}$ and $\mathit{N_{i}}$ are the number of positive and negative examples respectively.

The above evaluation criteria treat each attribute independently and ignore the inter-attribute correlation which exists naturally in multi-attribute recognition problems. Li at al.~\cite{li2016richly} call the above solution as $\mathit{label}$-$\mathit{based}$ evaluation criteria and propose using the $\mathit{example}$-$\mathit{based}$ evaluation criteria inspired by a fact that example-based evaluation captures better the consistency of prediction on a given pedestrian image~\cite{zhang2014review}. The $\mathit{example}$-$\mathit{based}$ evaluation criteria which are widely used include four metrics: accuracy, precision, recall rate, and F1 score, as defined below:
\begin{equation}
\label{eq::objective}
\begin{aligned}
\mathit{Acc_{exam}} = \dfrac{1}{N}\sum_{i=1}^{N}\dfrac{|Y_{i} \cap f(x_{i})|}{|Y_{i} \cup f(x_{i})|}
\end{aligned}
\end{equation}

\begin{equation}
\label{eq::objective}
\begin{aligned}
\mathit{Prex_{exam}} = \dfrac{1}{2N}\sum_{i=1}^{N}\dfrac{|Y_{i} \cap f(x_{i})|}{|f(x_{i})|}
\end{aligned}
\end{equation}

\begin{equation}
\label{eq::objective}
\begin{aligned}
\mathit{Rec_{exam}} = \dfrac{1}{2N}\sum_{i=1}^{N}\dfrac{|Y_{i} \cap f(x_{i})|}{|Y_{i}|}
\end{aligned}
\end{equation}

\begin{equation}
\label{eq::objective}
\begin{aligned}
\mathit{F1} = \dfrac{2*Prec_{exam}*Rec_{exam}}{Prec_{exam}+Rec_{exam}}
\end{aligned}
\end{equation}
where $\mathit{N}$ is the number of examples, $\mathit{Y_{i}}$ is the ground truth positive labels of the $\mathit{i}$-th example, $\mathit{f(x)}$ returns the predicted positive labels for $\mathit{i}$-th example. $\mathit{|\cdot|}$ means the set cardinality.


\section{Regular Pipeline for PAR} \label{traditionalMethods}
The pedestrian attributes in practical video surveillance may contain dozens of categories, as defined in many popular benchmarks. Learning each attribute independently is one intuitive idea, but it will make the PAR redundant and inefficient. Therefore, the researchers prefer to estimate all the attributes in one model and treat each attribute estimation as one task. Due to the elegance and efficiency of multi-task learning, it draws more and more attention. On the other hand, the model takes the given pedestrian image as input and outputs corresponding attributes. PAR also belongs to the domain of multi-label learning. In this section, we will give a brief introduction on the regular pipeline for pedestrian attribute recognition from these two aspects, \emph{i.e.}, the multi-label learning and multi-task learning.

\subsection{Multi-task Learning} \label{MTLalgorithm}
To handle one specific task in the machine learning community, a traditional solution is to design an evaluation criterion, extract related feature descriptors, and construct single or ensemble models. It uses the feature descriptors to optimize the model parameters and achieve the best results according to evaluation criteria to improve the overall performance. This pipeline may achieve satisfying results on a single task, however, it ignores the other tasks which may bring further improvements for the evaluation criterion. 

In the real world, many things are correlated. The learning of one task may rely on or constrain the others. Even one task is decomposed, but the sub-tasks still have correlations to some extent. Processing a single task independently is prone to ignore such correlations, thus, the improvement of final performance may meet the bottleneck. Specifically, the pedestrian attributes are correlated with each other, such as gender and clothing style. On the other hand, supervised learning needs massive annotated training data which is hard to collect. Therefore, the most popular approach is to joint learning multi-tasks to mine the shared feature representation. It has been widely applied in multiple domains, such as natural language processing, and computer vision. 

\begin{figure}[!htb]
\center
\includegraphics[width=3.5in]{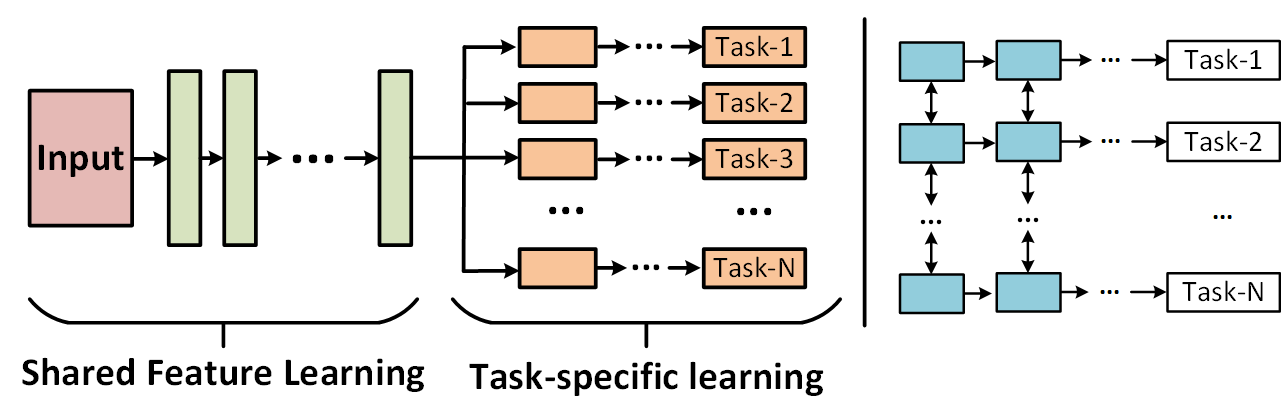}
\caption{The illustration of Hard (left sub-figure) and Soft (right sub-figure) parameter sharing for multi-task learning in deep neural networks \cite{ruder2017MTLoverview}.}
\label{MTLandMLC1}
\end{figure}

With the development of deep learning, many efficient algorithms are proposed by integrating multi-task learning and deep neural networks. To fully understand the reasons behind the efficiency of MTL, we need to analyze its detailed mechanism. According to the study of Ruder \emph{et al.} \cite{ruder2017MTLoverview}, the reasons can be concluded as following five points: implicit data augmentation, attention focusing, eavesdropping, representation bias, regularization. For the details of these reasons, please refer to their original paper. Generally speaking, there are two kinds of approaches in deep learning based multi-task learning, \emph{i.e.} the \emph{hard} and \emph{soft} parameter sharing. The hard parameter sharing usually takes the shallow layers as shared layers to learn the common feature representations of multiple tasks and treats the high-level layers as task-specific layers to learn more discriminative patterns. This model is the most popular framework in the deep learning community. The illustration of hard parameter sharing can be found in Figure \ref{MTLandMLC1} (left sub-figure). For the soft parameter sharing multi-task learning (as shown in Figure \ref{MTLandMLC1} (right sub-figure)), they train each task independently but make the parameters between different tasks similar via the introduced regularization constraints, such as $L_2$ distance \cite{duong2015lowl2norm} and trace norm \cite{yang2016traceNormMTL}.  

Therefore, it is rather intuitive to apply the multi-task learning for pedestrian attribute recognition and many algorithms are also proposed based on this framework \cite{sudowe2015person} \cite{acprli2015DeepMAR} \cite{acprli2015DeepMAR} \cite{abdulnabi2015multitaskCNN} \cite{bourdev2011describing} \cite{joo2013humanRAD} \cite{zhang2014panda} \cite{zhu2015multilabelCNN}.

\subsection{Multi-label Learning} \label{MLLalgorithm}
\begin{figure}[!htb]
\center
\includegraphics[width=3.5in]{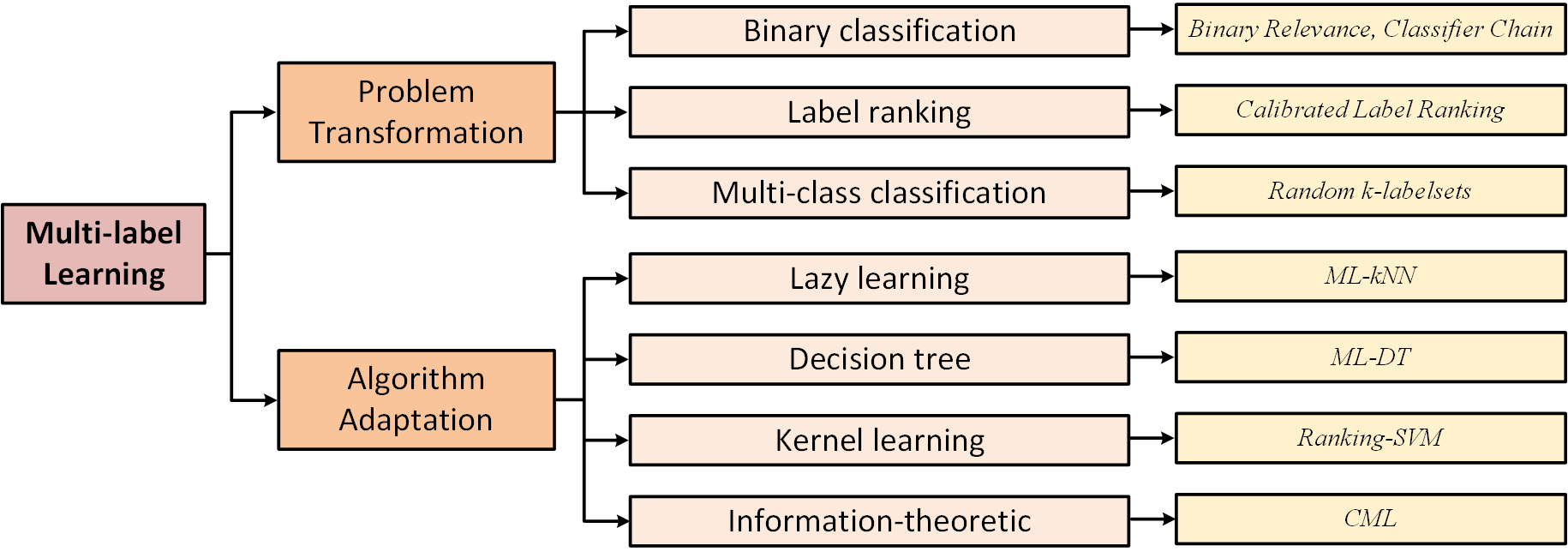}
\caption{Categorization of representative multi-label learning algorithms reviewed in \cite{zhang2014review}.}
\label{MTLandMLC2}
\end{figure}

For the multi-label classification algorithms, the following three kinds of learning strategy can be concluded as noted in \cite{zhang2014review}: 1). \emph{First-order strategy}, is the simplest form and could directly transform the multi-class into multiple binary-classification problems. Although it achieves better efficiency, this strategy can not model the correlations between multi-labels which leads to bad generic; 2). \emph{Second-order strategy}, takes the correlations between each label pair and achieves better performance than first-order strategy; 3). \emph{High-order strategy},  considers all the label relations and implements a multi-label recognition system by modeling the influence of each label on others. This approach is generic but with high complexity which may be weak at processing large-scale image classification tasks. Therefore, the following two approaches are frequently used for model construction: \emph{i.e.} the problem transformation and algorithm adaptation. The visualization of some representative multi-label learning algorithms can be found in Figure \ref{MTLandMLC2}, as noted in \cite{zhang2014review}. 

To simplify the multi-label classification using problem transformation, existing widely used frameworks can be adopted. Some representative algorithms are: 1). \emph{binary relevance algorithm} \cite{boutell2004learning}, which directly transforms the multi-label into multiple binary classification problems and finally fuses all the binary classifiers together for multi-label classification. This approach is simple and intuitive, but neglects the correlations between multiple labels; 2). \emph{classifier chain algorithm} \cite{read2011classifier}, the basic idea of this algorithm is to transform the multi-label learning problem into a binary classification chain. Each binary classifier is dependent on its previous one in the chain; 3). \emph{calibrated label ranking algorithm} \cite{furnkranz2008multilabel}, which considers the correlations between paired labels and transforms the multi-label learning into a label ranking problem; 4). \emph{random k-Labelsets algorithm} \cite{tsoumakas2011random}, it transforms the multi-label classification problem into sets of multiple classification problems, the classification task in each set is a multi-class classifier. And the categories the multi-class classifiers need to learn are the subset of all labels. 

Different from problem transformation, the algorithm adaptation directly improve existing algorithm and apply on multi-label classification problem, including: 1). \emph{multi-label k-nearest neighbor, ML-kNN} \cite{zhang2007mlMLKNN}, adopt the kNN techniques to handle the multi-class data and utilize the maximum a posteriori (MAP) rule to make prediction by reasoning with the labeling information embodied in the neighbors. 2).  \emph{multi-label decision tree, ML-DT} \cite{clare2001knowledge}, which attempts to deal with multi-label data with a decision tree, an information gain criterion based on multi-label entropy is utilized to build the decision tree recursively.  3). \emph{ranking support vector machine, Rank-SVM} \cite{elisseeff2002kernel}, adopt the maximum margin strategy to handle this problem, where a set of linear classifiers are optimized to minimize the empirical ranking loss and enabled to handle non-linear cases with kernel tricks. 4). \emph{collective multi-label classifier, CML} \cite{ghamrawi2005collective}, adapt maximum entropy principle to deal with multi-label tasks, where correlations among labels are encoded as constraints that the resulting distribution must satisfy. 

\begin{figure*}
\center
\includegraphics[width=7in]{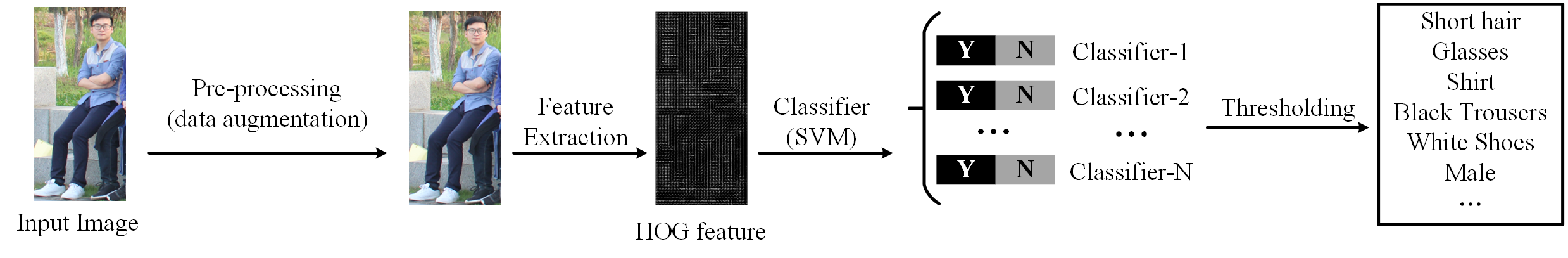}
\caption{The pipeline for regular pedestrian attribute recognition algorithm.}
\label{traditional-pipeline}
\end{figure*}

The regular pipeline of multi-label pedestrian attribute recognition can be found in Figure \ref{traditional-pipeline}. The machine learning model takes the human image (optionally pre-processed) as input and extracts its feature representation (such as HOG, SIFT, or deep features). Some commonly used pre-processing techniques are normalization, random cropping, whitening process, etc. It aims to improve the quality of the input image, suppress the unnecessary deformation, or augment the image features which may be important for subsequent operations, to improve the generic of the trained model. After that, they train a classifier based on extracted features to predict corresponding attributes. Existing deep learning-based PAR algorithm could jointly learn the feature representation and classifier in an end-to-end manner which significantly improve the final recognition performance.


\section{Deep Neural Networks}

In this subsection, we will review some well-known network architectures in the deep learning community that are already or may be used for the pedestrian attribute recognition task.

\begin{figure}[htb!]
\center
\includegraphics[width=3.5in]{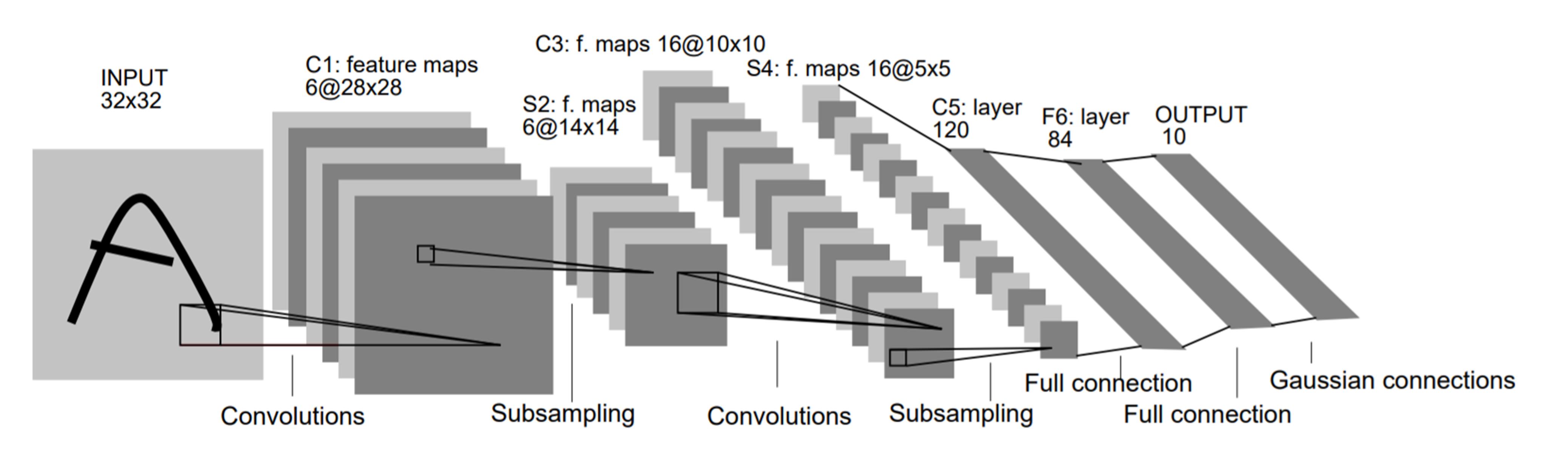}
\caption{Architecture of LeNet-5, a Convolutional Neural Network, here for digits recognition. Each plane is a feature map, i.e. a set of units whose weights are constrained to be identical.}
\label{Network-LeNet}
\end{figure}

\textbf{LeNet} \cite{lecun1998gradient} is first proposed by Yann LeCun \emph{et al.} in 1998. It is first designed for handwritten and machine-printed character recognition, as shown in the website \footnote{\url{http://yann.lecun.com/exdb/lenet/}}. The architecture of LeNet can be found in Figure \ref{Network-LeNet}. It takes $32 \times 32$ single channel images as inputs and uses 2 groups of convolutional + max pooling layers to extract its features. The classification is done by 2 fully-connected layers and outputs the distribution of numbers.

\begin{figure}[htb!]
\center
\includegraphics[width=3.3in]{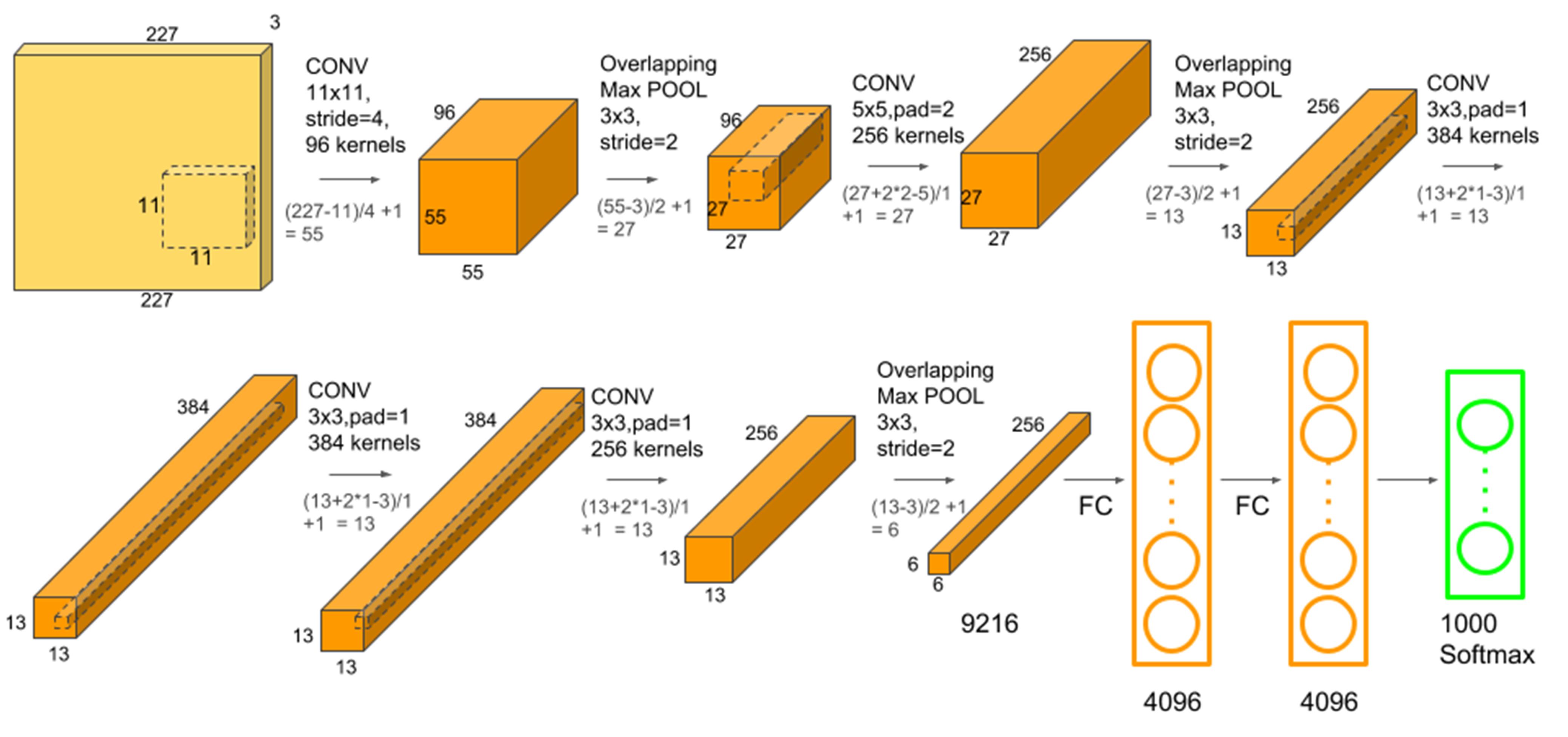}
\caption{Architecture of AlexNet for ImageNet classification. This figure is taken from: \url{https://www.learnopencv.com/understanding-alexnet/}.}
\label{Network-AlexNet}
\end{figure}

\textbf{AlexNet} \cite{krizhevsky2012imagenet} is a milestone in deep learning history which was proposed by Alex \emph{et al.} in 2012 and won the ILSVRC-2012 with a TOP-5 test accuracy of 84.6\%. AlexNet was much larger than previous CNNs used for computer vision tasks, such as LeNet. It has 60 million parameters and 650,000 neurons, as shown in Figure \ref{Network-AlexNet}. It consists of 5 convolutional layers, max-pooling layers, Rectified Linear Units (ReLUs) as non-linearities, three fully connected layers, and a dropout unit.

\textbf{VGG}\cite{simonyan2014very} is a CNN model proposed by the Visual Geometry Group (VGG) from the University of Oxford. This network uses more convolutional layers (16, 19 layers) and also achieves good results on ILSVRC-2013. Many subsequent neural networks all follow this network. It first uses a stack of convolutional layers with small receptive fields in the first layers while previous networks adopt layers with large receptive fields. The smaller receptive fields can reduce the parameters by a wide margin and more non-linearities, making the learned features more discriminative and also more efficient for training. 

\begin{figure}[htb!]
\center
\includegraphics[width=3in]{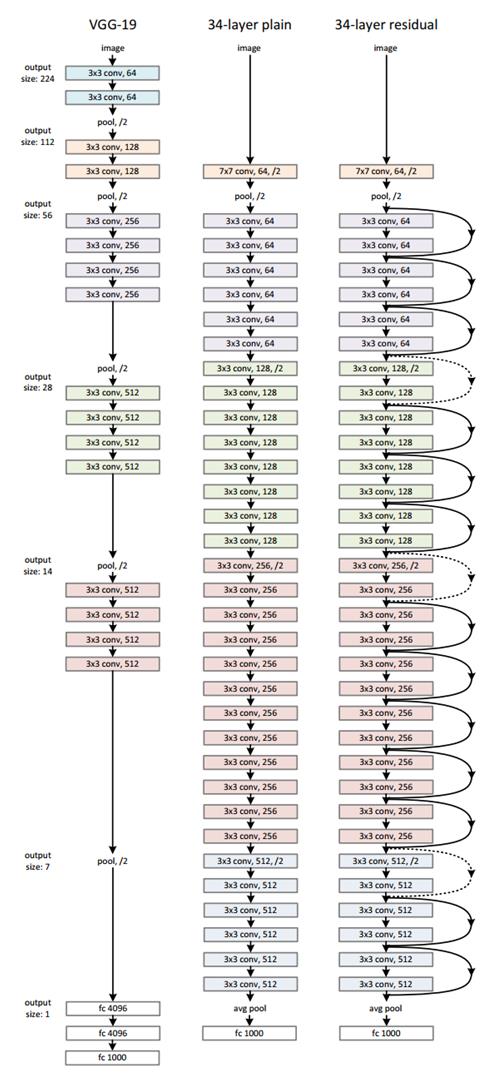}
\caption{Architecture of VGG-19, 34-layer plain and Residual Network-34 for ImageNet classification. This figure is taken from the paper of Residual Network \cite{he2016deep}.}
\label{Network-VGG-Residual}
\end{figure}	

\begin{figure}[htb!]
\center
\includegraphics[width=3.3in]{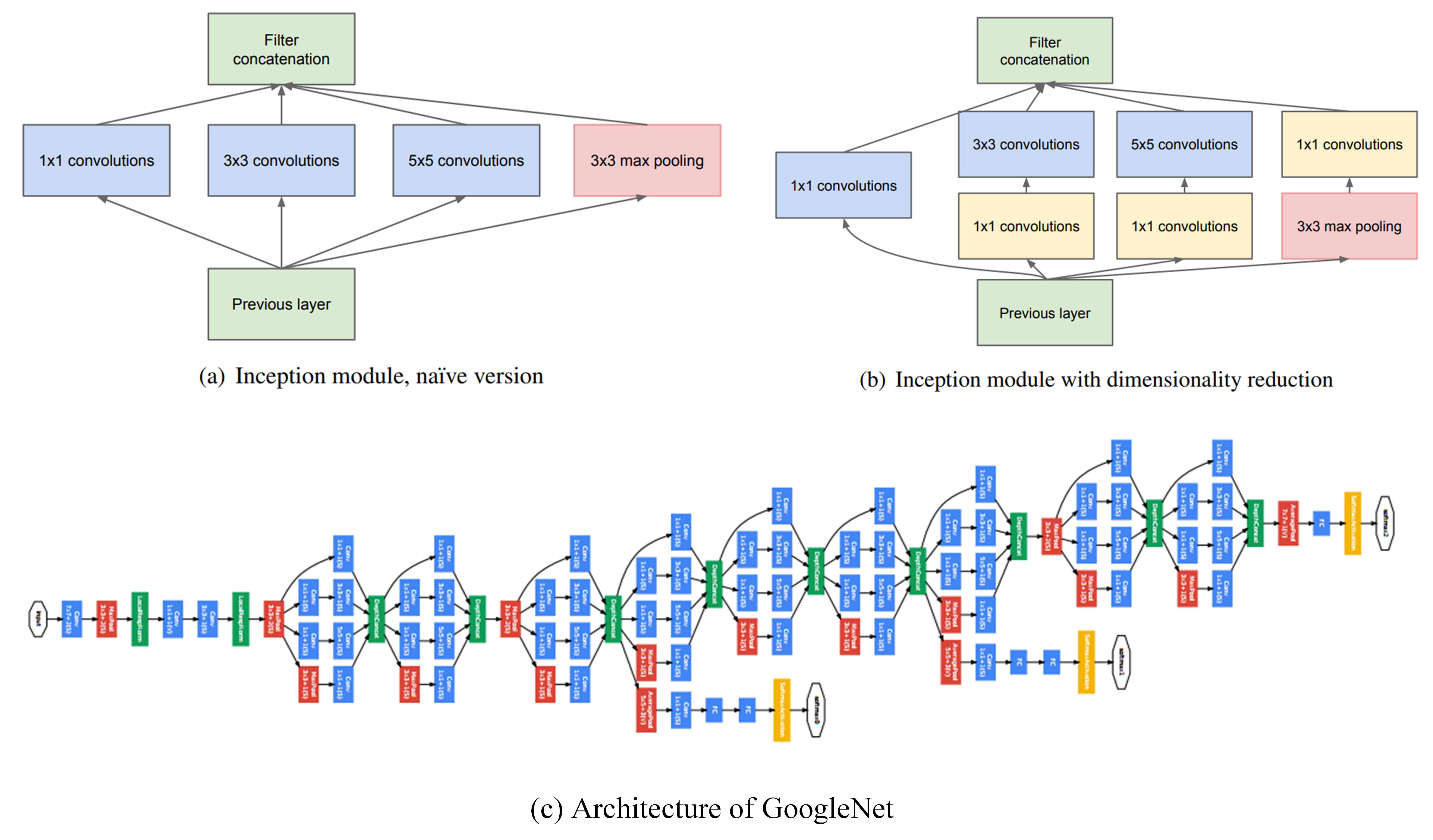}
\caption{Architecture of GoogleNet. This figure is taken from the paper of GoogleNet \cite{szegedy2015going}.}
\label{Network-GoogleNet}
\end{figure}

\textbf{GoogleNet}\cite{szegedy2015going} is another popular network architecture in the deep learning community (22 layers). Different from the traditional sequential manner, this network first introduced the concept of \emph{inception} module and won the competition of ILSVRC-2014. See Figure \ref{Network-GoogleNet} for details. The GoogleNet contains a Network in Network (NiN) layer \cite{lin2013network}, a pooling operation, a large-sized convolution layer, and a small-sized convolution layer. These layers can be computed in parallel and followed by $1\times1$ convolution operations to reduce dimensionality.

\begin{figure}[htb!]
\center
\includegraphics[width=2.5in]{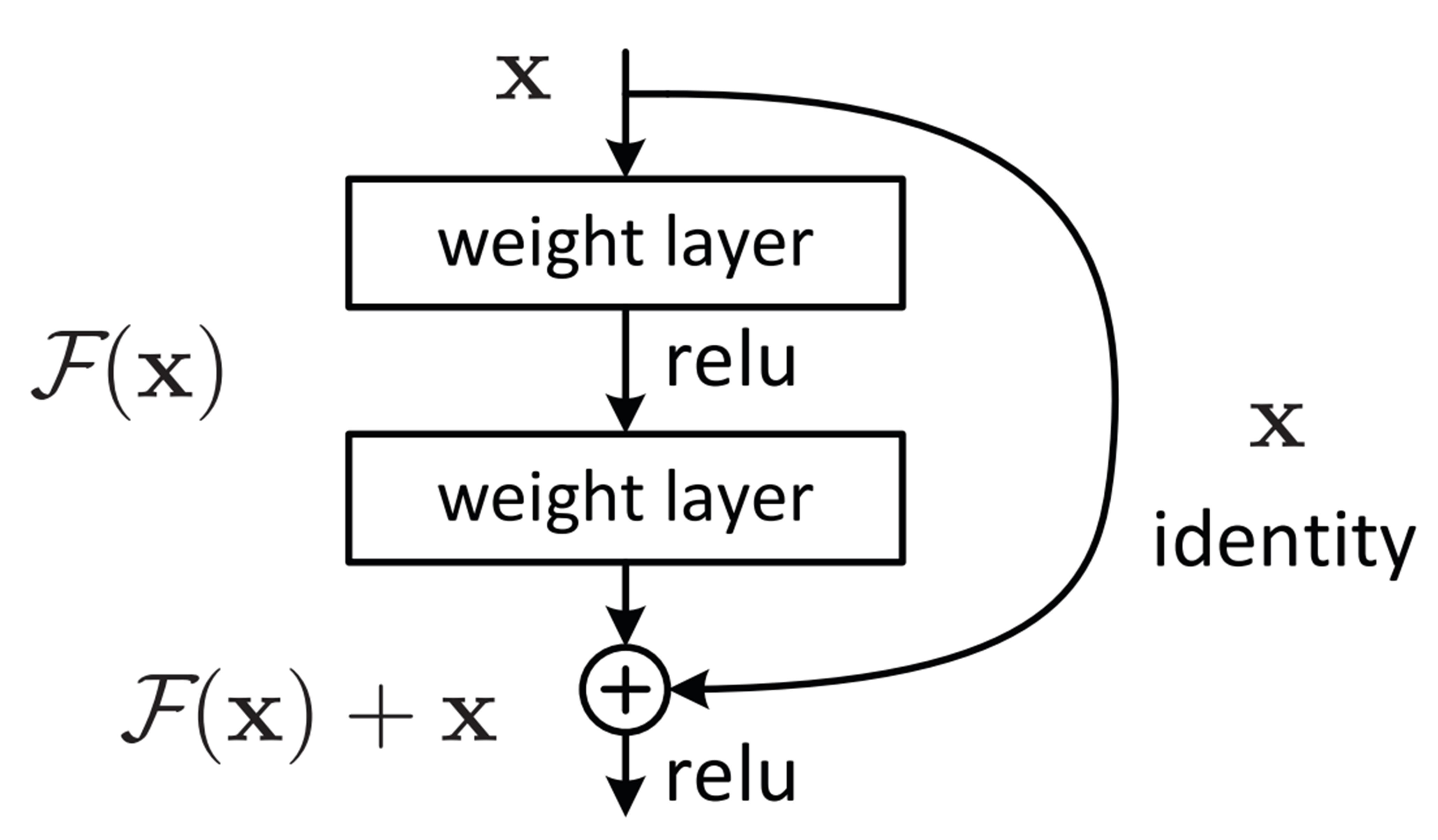}
\caption{Residual learning: a building block. This figure is taken from the paper of Residual Network \cite{he2016deep}.}
\label{Network-residual}
\end{figure}	

\textbf{Residual Network} \cite{he2016deep} is first well known for its super deep architecture (more than 1k layers) while previous networks are rather ``shallow'' by comparison. The key contribution of this network is the introduction of residual blocks, as shown in Figure \ref{Network-residual}. This mechanism can address the problem of training a really deep architecture by introducing identity skip connections and copying their inputs to the next layer. The vanishing gradients problem can be handled to a large extent with this method.

\textbf{Dense Network} \cite{huang2017densely} is proposed by Huang \emph{et al.} in 2017. This network further extends the idea of residual network and has better parameter efficiency, one big advantage of DenseNets is their improved flow of information and gradients throughout the network, which makes them easy to train. Each layer has direct access to the gradients from the loss function and the original input signal, leading to implicit deep supervision \cite{lee2015deeply}. This helps in training deeper network architectures. 

\begin{figure}[htb!]
\center
\includegraphics[width=3in]{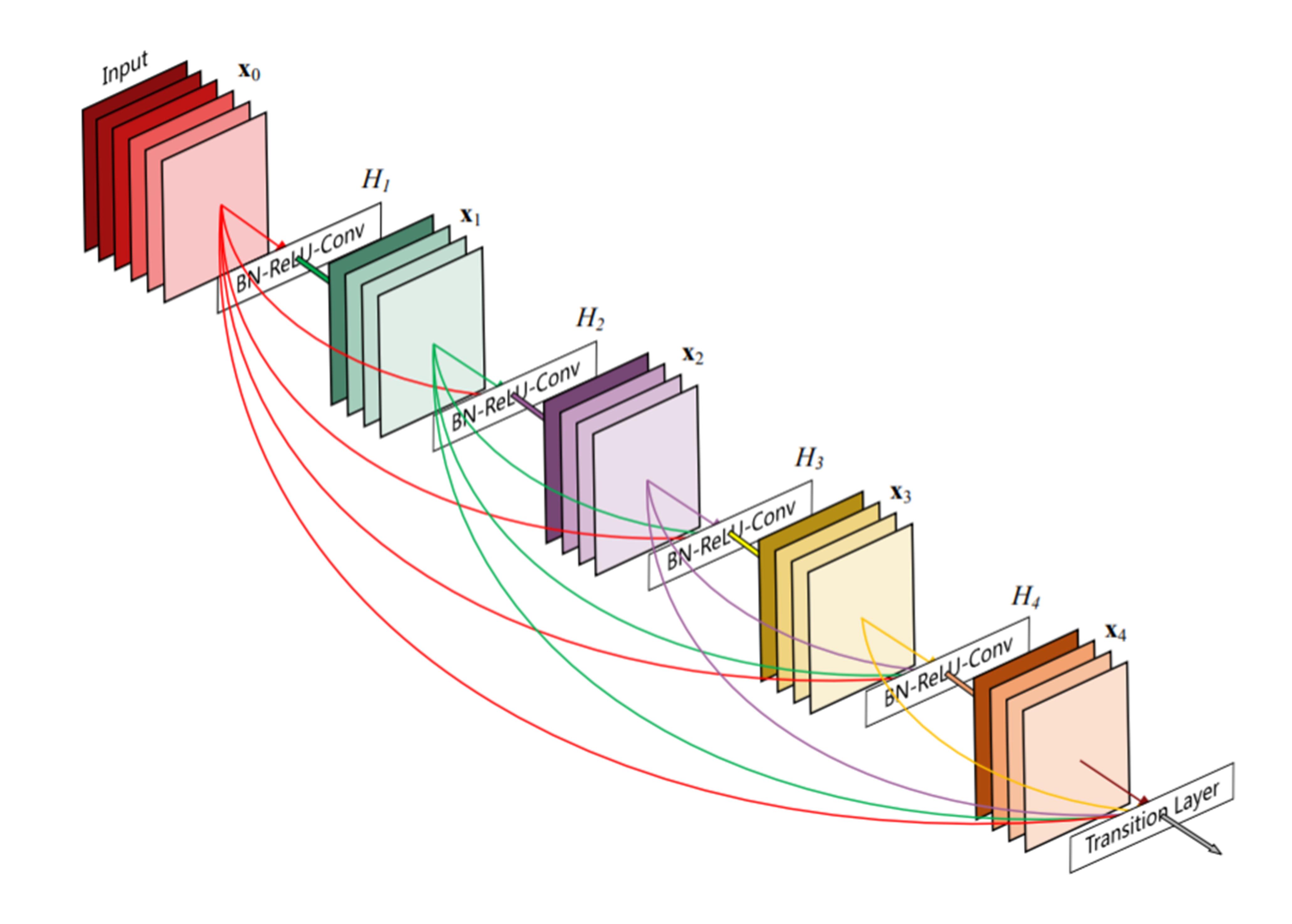}
\caption{A 5-layer dense block with a growth rate of k = 4. Each layer takes all preceding feature maps as input. This figure is taken from the paper of DenseNet \cite{huang2017densely}.}
\label{Network-DenseNet}
\end{figure}

\begin{figure}[htb!]
\center
\includegraphics[width=3.5in]{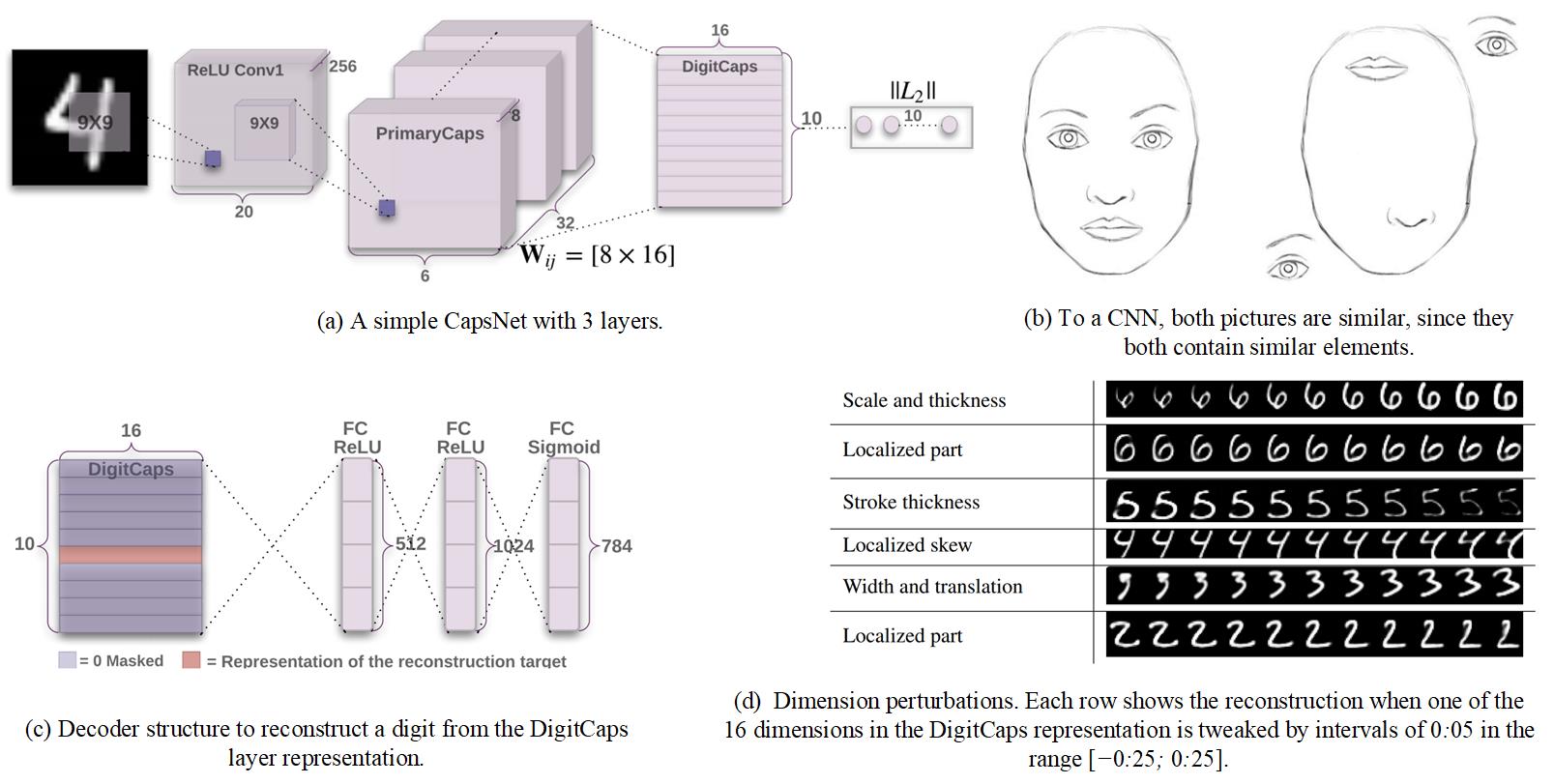}
\caption{The motivation and illustration of Capsule Network. This figure is taken from the paper of Capsule Network \cite{sabour2017dynamic} and the blog {\url{https://medium.com/ai\%C2\%B3-theory-practice-business/understanding-hintons-capsule-networks-part-i-intuition-b4b559d1159b}}.}
\label{Network-CapsuleNetwork}
\end{figure}

\textbf{Capsule Network} \cite{sabour2017dynamic} \cite{hinton2018matrix} is introduced in 2017 by Hiton \emph{et al.} to handle the limitations of standard CNN. As we all know, the use of max pooling layers in standard CNN reduced the dimension of feature maps generated from previous convolutional layers and made the feature learning process more efficient. As shown in Figure \ref{Network-CapsuleNetwork} (b), the two face images are similar to a CNN due to they both contain similar elements. However, standard CNN can not capture the difference between the two face images of the usage of max pooling layers. The capsule network is proposed to handle this issue by abandoning the max pooling layers and use a \emph{capsule} to output a vector instead of a value for each neuron. This makes it possible to use a powerful dynamic routing mechanism (``routing-by-agreement'') to ensure that the output of the capsule gets sent to an approximate parent in the layer above. The utilization of margin loss and re-construction loss for the training of the capsule network validated its effectiveness. Some ablation studies also demonstrate the attributes of each digit can be encoded in the output vectors by the capsule network.

\begin{figure}[htb!]
\center
\includegraphics[width=3.5in]{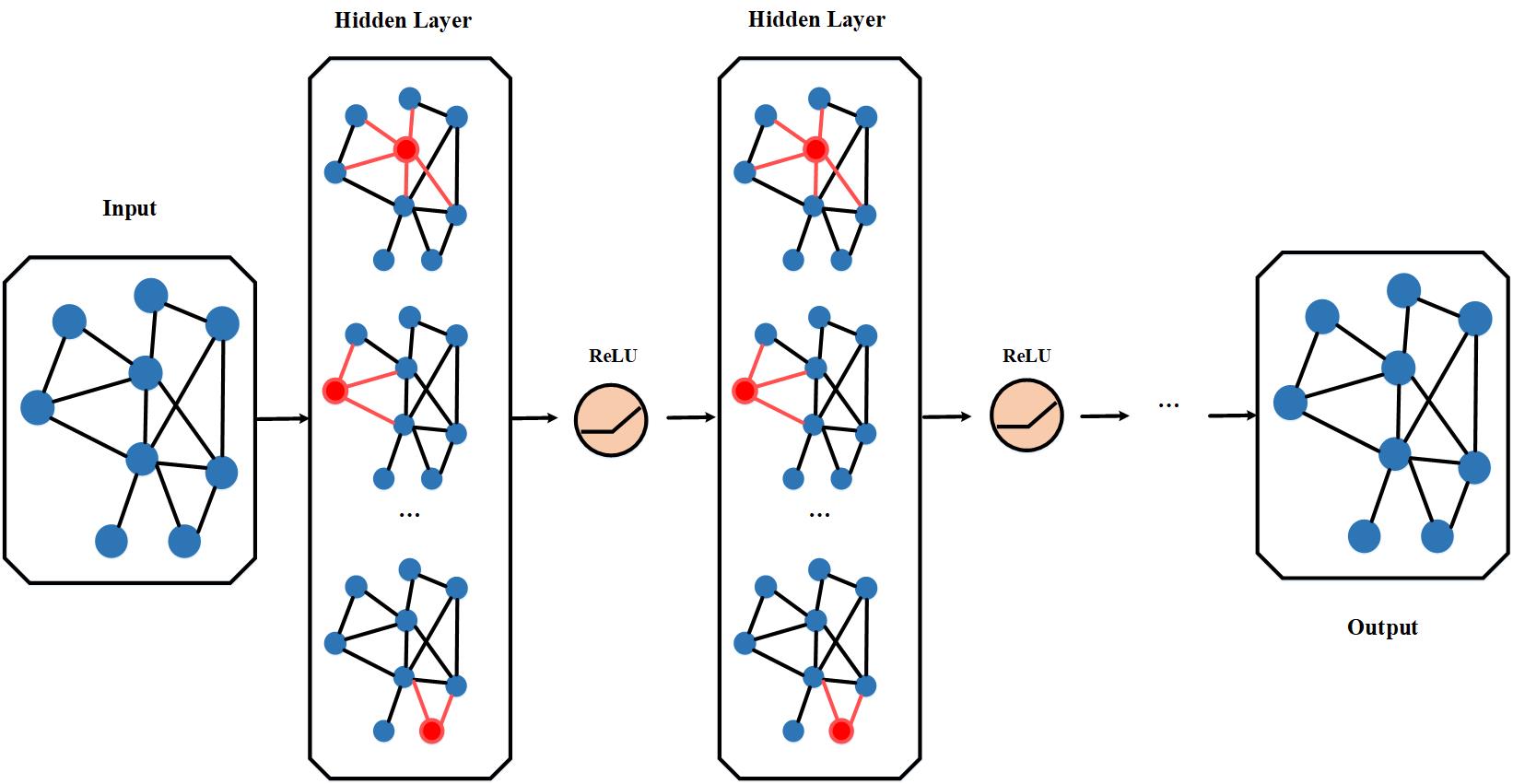}
\caption{The illustration of Graph Convolutional Network (GCN). This figure is rewritten based on the blog {\url{http://tkipf.github.io/graph-convolutional-networks/}}}
\label{Network-GCN}
\end{figure}

\textbf{Graph Convolutional Network} \cite{kipf2016semi} attempt to extend the CNN into non-grid data due to the standard convolution operation on images/videos can not be directly used in graph-structured data. The goal of GCN is to learn a function of signals/features on a graph $G = (V, E)$ which takes feature description $x_i$ for each node $i$ and representative description of the adjacency matrix $\textbf{A}$ as input and produces a node-level output $\textbf{Z}$. The overall architecture of GCN can be found in Figure \ref{Network-GCN}.

\begin{figure}[htb!]
\center
\includegraphics[width=2.5in]{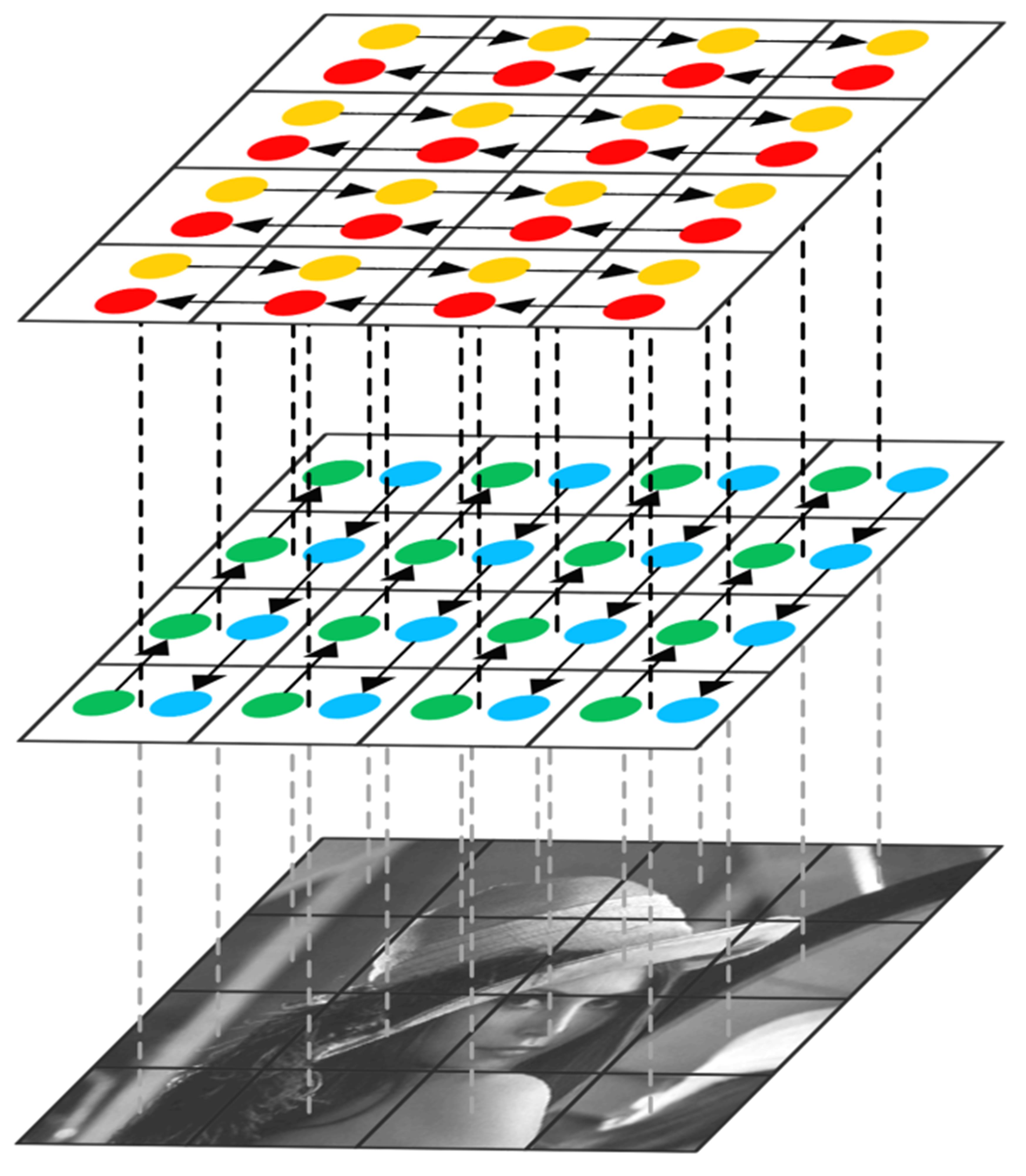}
\caption{One layer of ReNet architecture modeling vertical and horizontal spatial dependencies. This figure is adopted from \cite{visin2015renet}.}
\label{Network-ReNet}
\end{figure}

\textbf{ReNet} \cite{visin2015renet} In order to extend Recurrent Neural Networks (RNNs) architectures to multi-dimensional tasks, Graves et al. \cite{graves2009offline} proposed a Multi-dimensional Recurrent Neural Network (MDRNN) architecture which replaces each single recurrent connection from standard RNNs with $d$ connections, where $d$ is the number of spatio-temporal data dimensions. Based on this initial approach, Visin et al. \cite{visin2015renet} proposed ReNet architecture in which instead of multidimensional RNNs, they have been using usual sequence RNNs. In this way, the number of RNNs is scaled linearly at each layer regarding the number of dimensions $d$ of the input image (2d). In this approach, each convolutional layer (convolution + pooling) is replaced with four RNNs sweeping the image vertically and horizontally in both directions as we can see in Figure \ref{Network-ReNet}.

\textbf{Recurrent Neural Network, RNN.} Traditional neural network is based on the assumption that all inputs and outputs are independent of each other, however, the assumption may not be true in many tasks, such as sentence translation. Recurrent Neural Network (RNN) is proposed to handle the task that involves sequential information. RNNs are called \emph{recurrent} because they perform the same task for every element of a sequence, with the output being dependent on the previous computations. Another way to think about RNNs is that they have a "memory" that captures information about what has been calculated so far. In theory, RNNs can make use of information in arbitrarily long sequences, but in practice, they are limited to looking back only a few steps. 
\begin{figure}[htb!]
\center
\includegraphics[width=3.5in]{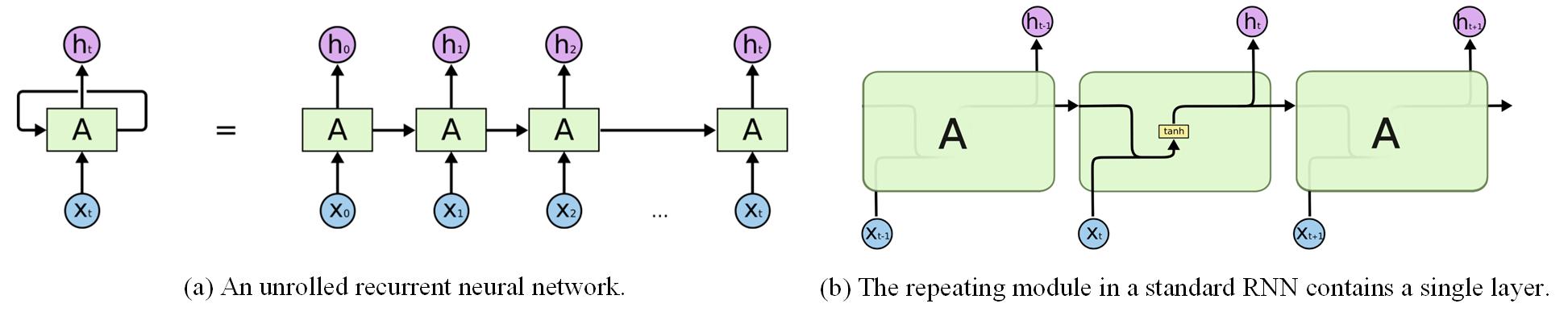}
\caption{The illustration of RNN. This figure is adapted from {\url{http://colah.github.io/posts/2015-08-Understanding-LSTMs/}}.}
\label{Network-RNN}
\end{figure}

\begin{figure}[htb!]
\center
\includegraphics[width=3.5in]{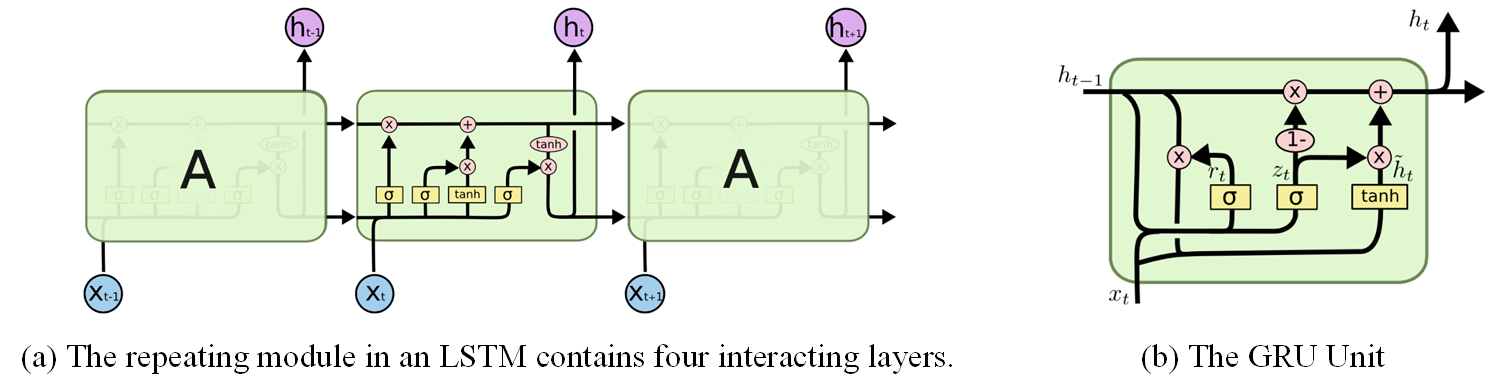}
\caption{The illustration of LSTM and GRU Unit. This figure is adapted from {\url{http://colah.github.io/posts/2015-08-Understanding-LSTMs/}}.}
\label{Network-LSTM-GRU}
\end{figure}

\textbf{Long Short-term Memory, LSTM} is introduced to handle the issue of gradient vanish or explosion of RNN. An LSTM has three of these gates, to protect and control the cell state, \emph{i.e.} the forget gate, input gate, and output gate. Specifically, we denote the input sequences as $\textbf{X} = (\textbf{x}_1, \textbf{x}_2, ... , \textbf{x}_N)$. At each position $k, k \in [1, N]$, there is a set of internal vectors, including an input gate $\textbf{i}_k$, a forget gate $\textbf{f}_k$, an output gate $\textbf{o}_k$ and a memory cell $\textbf{c}_k$. The hidden state $\textbf{h}_k$ can be computed by all these vectors, as follows:
\begin{align}
\label{lstmUpdatefunction}
\textbf{i}_k & = \sigma (\textbf{W}^i \textbf{x}_k + \textbf{V}^i\textbf{h}_{k-1} + \textbf{b}^i), \\
\textbf{f}_k & = \sigma (\textbf{W}^f \textbf{x}_k + \textbf{V}^f\textbf{h}_{k-1} + \textbf{b}^f), \\
\textbf{o}_k & = \sigma (\textbf{W}^o \textbf{x}_k + \textbf{V}^o\textbf{h}_{k-1} + \textbf{b}^o), \\
\textbf{c}_k & =  \textbf{f}_k \odot \textbf{c}_{k-1} + \textbf{i}_k \odot \tanh(\textbf{W}^c \textbf{x}_k + \textbf{V}^c\textbf{h}_{k-1} + \textbf{b}^c), \\
\textbf{h}_k & = \textbf{o}_k \odot \tanh(\textbf{c}_k)
\end{align}
where $\sigma$ is the sigmoid function, $\odot$ is the element-wise multiplication of two vectors, and all $\textbf{W}^*$, $\textbf{V}^*$, $\textbf{b}^*$ are weight matrices and vectors to be learned. Please see Figure \ref{Network-LSTM-GRU} (a) for detailed information on LSTM. 

\textbf{GRU.} A slightly more dramatic variation on the LSTM is the Gated Recurrent Unit, or GRU, introduced by \cite{cho2014learning}. It combines the forget and input gates into a single “update gate.” It also merges the cell state and hidden state and makes some other changes. The resulting model is simpler than standard LSTM models and has been growing increasingly popular. The detailed GRU Unit can be found in Figure \ref{Network-LSTM-GRU} (b).

\begin{figure}[htb!]
\center
\includegraphics[width=3.3in]{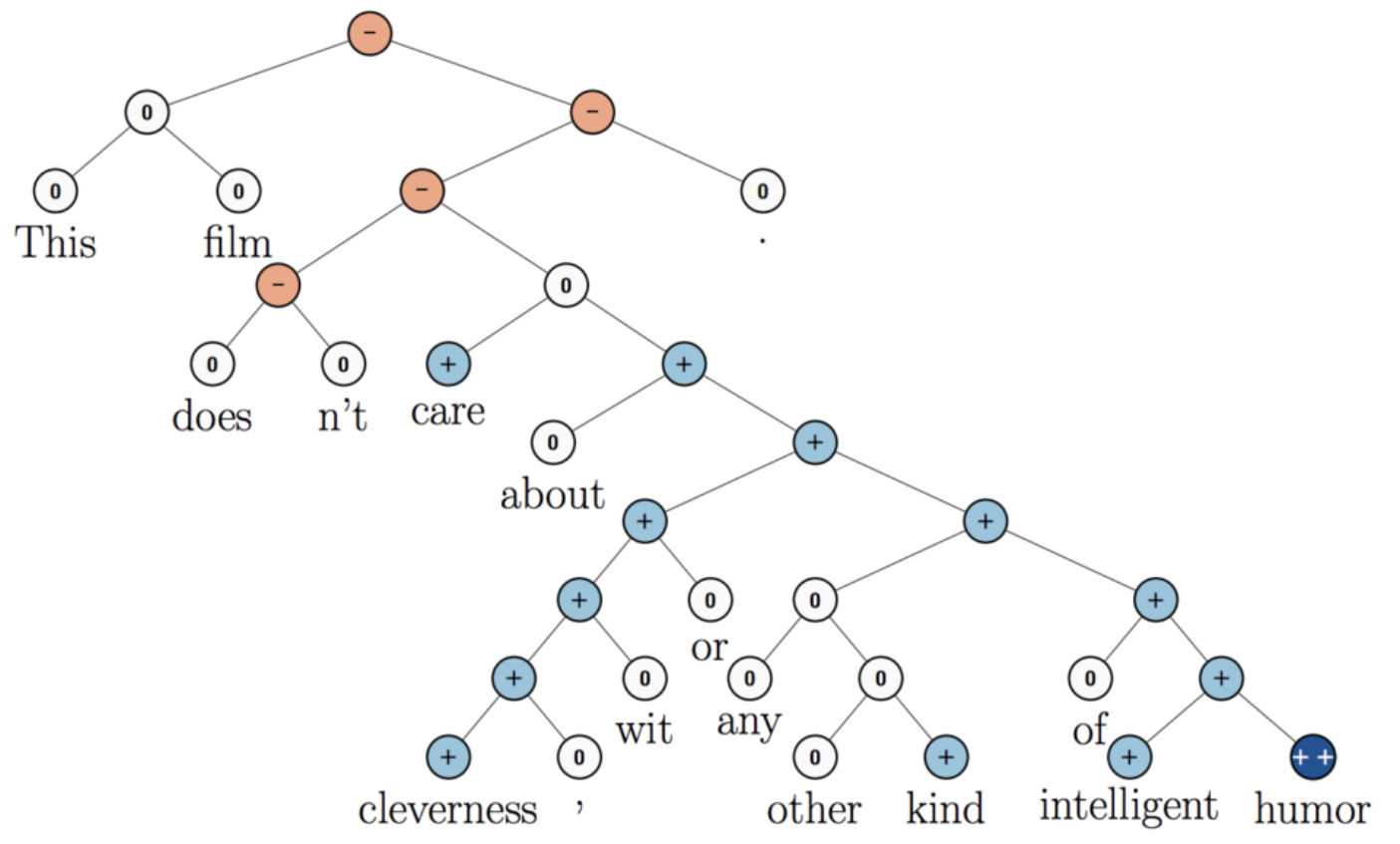}
\caption{The illustration of Recursive Neural Network (RvNN).}
\label{Network-RvNN}
\end{figure}

\textbf{Recursive Neural Network (RvNN)} \cite{irsoy2014deeprecursive} 
As noted in \footnote{\url{https://en.wikipedia.org/wiki/Recursive_neural_network}}, a recursive neural network is a kind of deep neural network created by applying the same set of weights recursively over a structured input, to produce a structured prediction over variable-size input structures, or a scalar prediction on it, by traversing a given structure in topological order. RvNNs have been successful, for instance, in learning sequence and tree structures in natural language processing, mainly phrase and sentence continuous representations based on word embedding. The illustration of RvNN can be found in Figure \ref{Network-RvNN}. 

\textbf{Sequential CNN.} \cite{gehring2017convolutional, johnson2015effective}
Different from regular works that use RNN to encode the time series inputs, the researchers also study CNN to achieve more efficient operation. With the sequential CNN, the computations over all elements can be fully parallelized during training to better exploit the GPU hardware and optimization is easier since the number of non-linearities is fixed and independent of the input length \cite{gehring2017convolutional}. 

\begin{figure}[htb!]
\center
\includegraphics[width=3.5in]{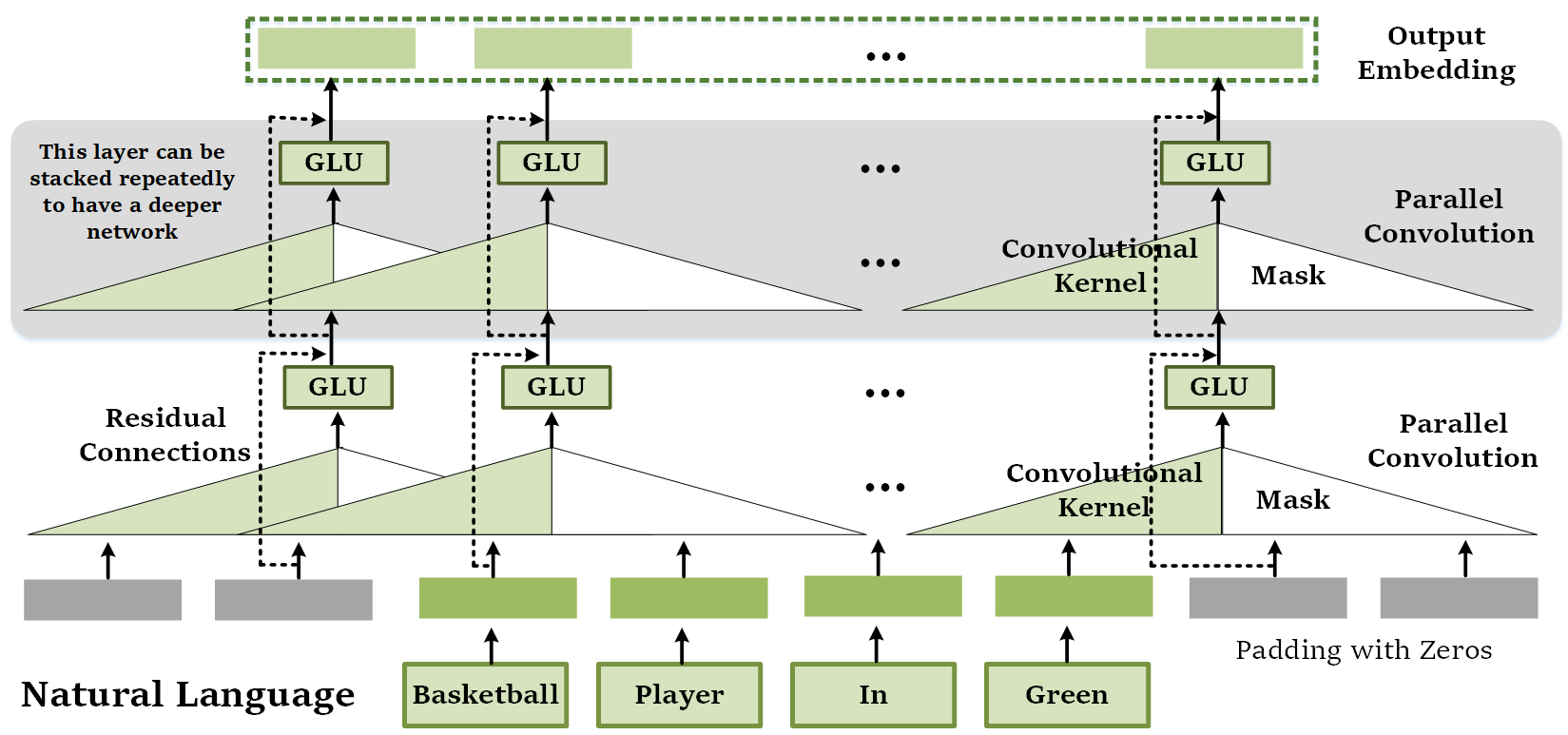}
\caption{The illustration of all convolutional networks for sequential modeling. This figure is adapted from \cite{wang2018describe}.}
\label{Network-CNNforsequential}
\end{figure}

\textbf{External Memory Network.} \cite{graves2016hybrid}
The visual attention mechanism can be seen as a kind of short-term memory that allocates attention over input features they have recently seen, while an external memory network could provide long-term memory through the read-and-write operation. It has been widely used in many applications such as visual tracking \cite{Yang_2018_ECCV_DMN}, visual question answering \cite{xiong2016dynamicVQA, Ma_2018_CVPR_VQA}. 

\begin{figure}[htb!]
\center
\includegraphics[width=3.5in]{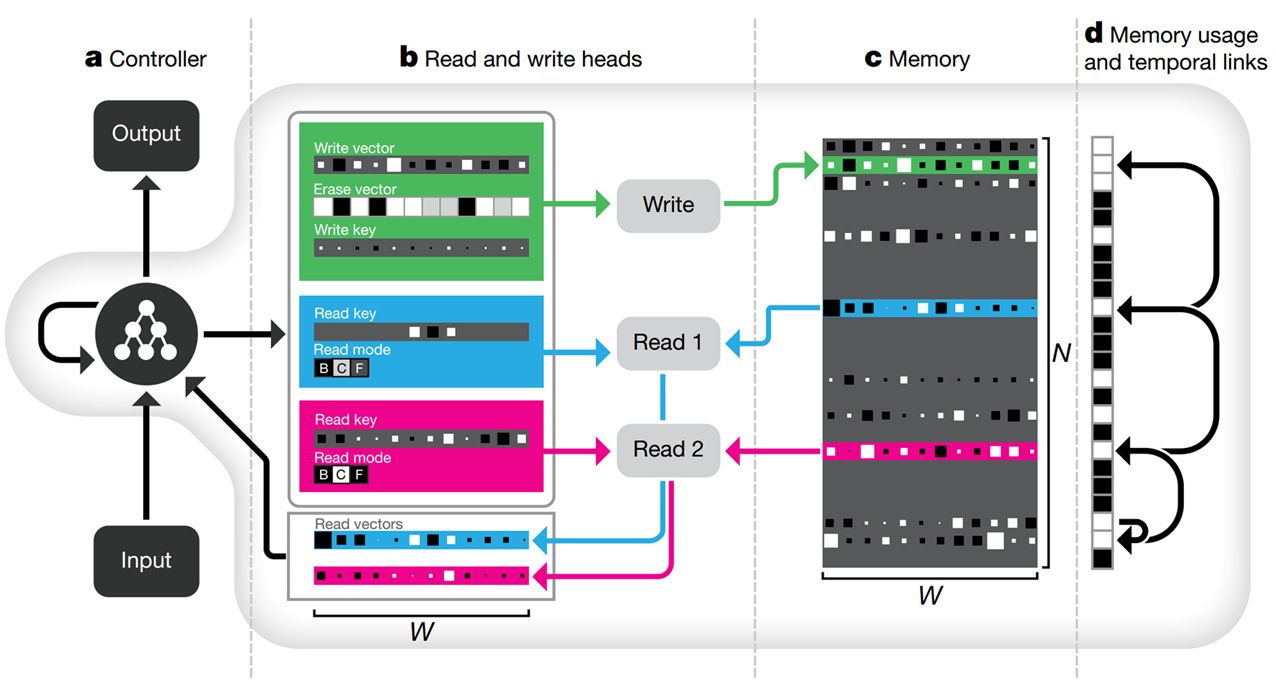}
\caption{The illustration of external memory network. This figure is adapted from DNC \cite{graves2016hybrid}.}
\label{Network-externalMemoryNet}
\end{figure}	

\begin{figure*}[htb!]
\center
\caption{The illustration of VAE, GAN, and CGAN.} \label{VAEGANCGAN}
\includegraphics[width=6.5in]{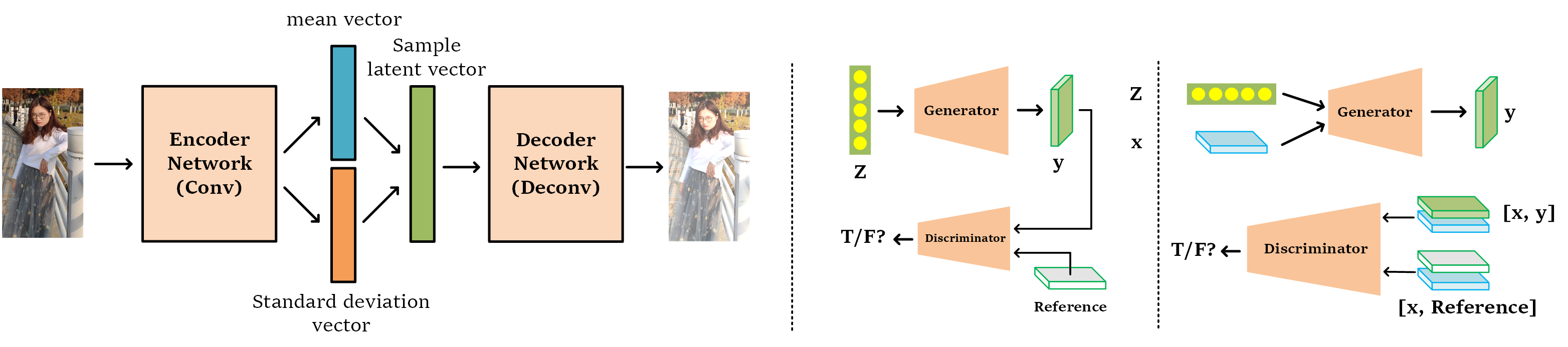}
\end{figure*}

\textbf{Deep Generative Model.}  In recent years, deep generative models achieved great development and many popular algorithms have been proposed, such as VAE (variational auto-encoder)~\cite{doersch2016tutorial}, GAN (generative adversarial networks)~\cite{goodfellow2014generative}, CGAN (conditional generative adversarial network)~\cite{mirza2014conditional}. The illustration of these three models can be found in Fig. \ref{VAEGANCGAN}. We think the strategy of attribute-based pedestrian image generation can handle the issue of low resolution, and unbalanced data distribution and augment the training dataset significantly. 

\section{The Review of PAR Algorithms}\label{DeepAlgorithms}
In this section, we will review the deep neural network-based PAR algorithms from the following eight aspects: global-based, local parts-based, visual attention-based, sequential prediction-based, newly designed loss function-based, curriculum learning-based, graphic model-based, and others algorithms.

\subsection{Global Image-based Models}
In this section, we will review the PAR algorithms which consider global image only, such as ACN \cite{sudowe2015person}, DeepSAR \cite{acprli2015DeepMAR}, DeepMAR \cite{acprli2015DeepMAR}, MTCNN \cite{abdulnabi2015multitaskCNN}. 

\subsubsection{ACN (ICCVW-2015) \cite{sudowe2015person}}

\begin{figure}[htb!]
\center
\includegraphics[width=3.5in]{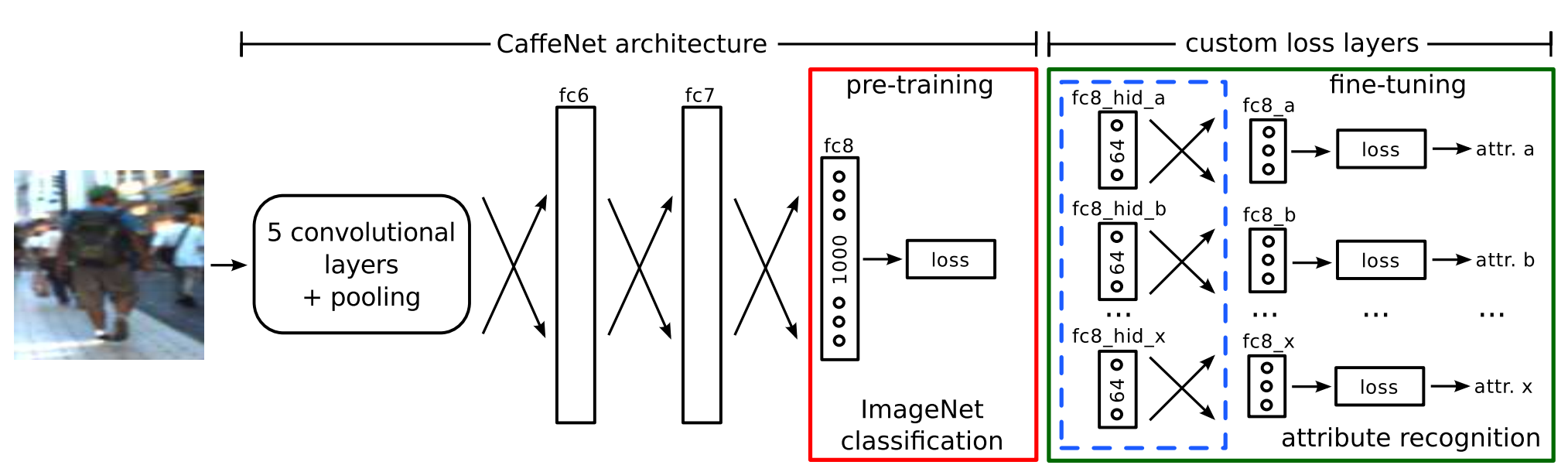}
\caption{The illustration of ACN \cite{sudowe2015person}.}
\label{DeepAlgorithms-ACN}
\end{figure}	

This paper proposes a multi-branch classification layer for each attribute learning with the convolutional network. As shown in Fig. \ref{DeepAlgorithms-ACN}, they adopt a pre-trained AlexNet as a basic feature extraction sub-network and replace the last fully connected layer with one loss per attribute using the KL-loss (Kullback-Leibler divergence-based loss function). The specific formulation can be described as follows: 
\begin{equation}
KL(P||Q) = \sum_{i}^{N} P(x_i) log \frac{Q(x_i)}{P(x_i)} 
\end{equation}
where $Q$ is the neural network's prediction and $P$ is the binary attribute's state in reality. 

In addition, they also propose a new dataset named PARSE-27k to support their evaluation. This dataset contains 27000 pedestrians and is annotated with 10 attributes. Different from the regular person attribute dataset, they propose a new category annotation, \emph{i.e., not decidable (N/A)}. Because for most input images, some attributes are not decidable due to occlusion, image boundaries, or any other reason.

\subsubsection{DeepSAR and DeepMAR (ACPR-2015) \cite{acprli2015DeepMAR}}

\begin{figure}[htb!]
\center
\includegraphics[width=3.5in]{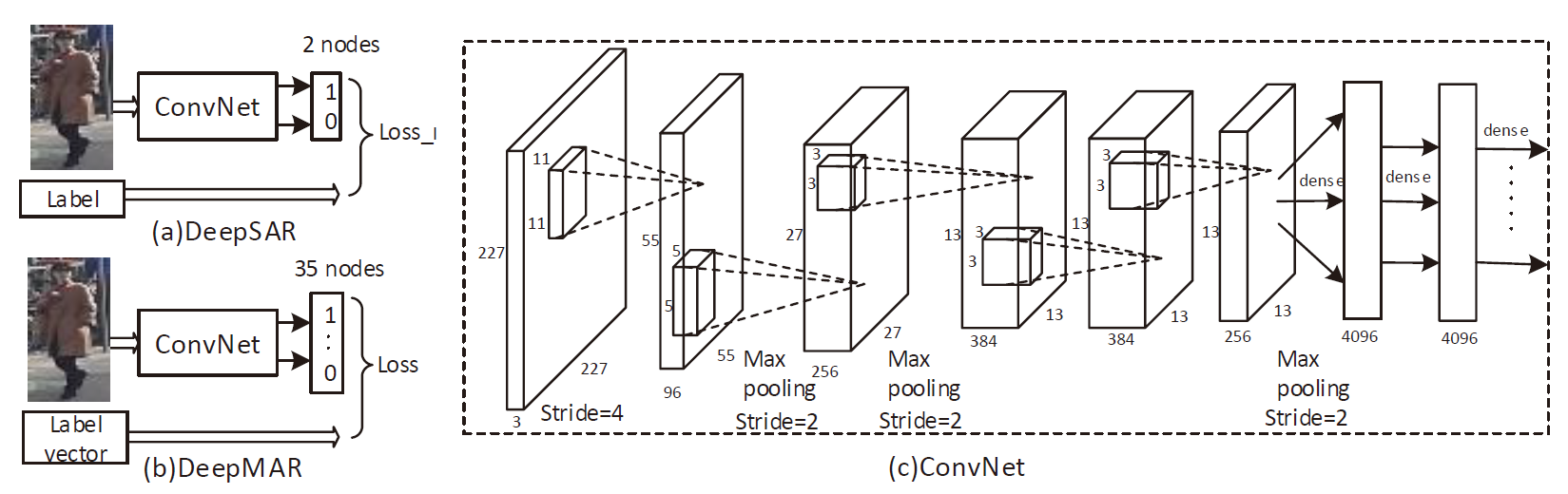}
\caption{The overall pipeline of DeepSAR and DeepMAR \cite{acprli2015DeepMAR}.}
\label{DeepAlgorithms-DeepMAR}
\end{figure}	

This paper introduces a deep neural network for person attribute recognition and attempts to handle the following two issues that exist in traditional methods: 
1). Hand-crafted features used in existing methods, like HOG, color histograms, and LBP (local binary patterns); 
2). Correlations between attributes are usually ignored. The authors propose two algorithms \emph{i.e.} DeepSAR and DeepMAR in this paper, as shown in Fig. \ref{DeepAlgorithms-DeepMAR}. They adopt AlexNet as their backbone network and obtain the DeepSAR by changing the output category defined in the last dense layer into two. The softmax loss is adopted to compute the final classification loss.  

Although the DeepSAR can use deep features for binary classification. However, it did not model the correlations between human attributes which may be the key to further improving the overall recognition performance. Therefore, they propose the DeepMAR which takes a human image and its attribute label vectors simultaneously and jointly considers all the attributes via sigmoid cross entropy loss: 
\begin{equation}
\label{sigmoidCELossFunction}
L_{ce} = -\frac{1}{N} \sum_{i=1}^{N} \sum_{l=1}^{L} y_{il} log(\hat{P_{il}}) + (1-y_{il}) log (1-\hat{p_{il}}) 
\end{equation}
\begin{equation}
\hat{p_{il}} = \frac{1}{1+exp(-x_{il})}
\end{equation}
where $\hat{p_{il}}$ is the estimated score for the $l$'th attribute of sample $x_i$. $y_{il}$ is the ground truth label.

In addition, they also consider the unbalanced label distribution in practical surveillance scenarios and propose an improved loss function as follows: 
\begin{equation}
\label{weightedsigmoidCELossFunction}
L_{wce} = -\frac{1}{N} \sum_{i=1}^{N} \sum_{l=1}^{L} w_l (y_{il} log(\hat{P_{il}}) + (1-y_{il}) log (1-\hat{p_{il}})) 
\end{equation}
\begin{equation}
w_l = exp(-p_l / \sigma^2)
\end{equation}
where $w_l$ is the loss weight for the $l^{th}$ attribute. $p_l$ denote the positive ratio of $l^{th}$ attribute in the training dataset. $\sigma$ is a hyperparameter.

\subsubsection{MTCNN (TMM-2015) \cite{abdulnabi2015multitaskCNN}}

\begin{figure}[htb!]
\center
\includegraphics[width=3.5in]{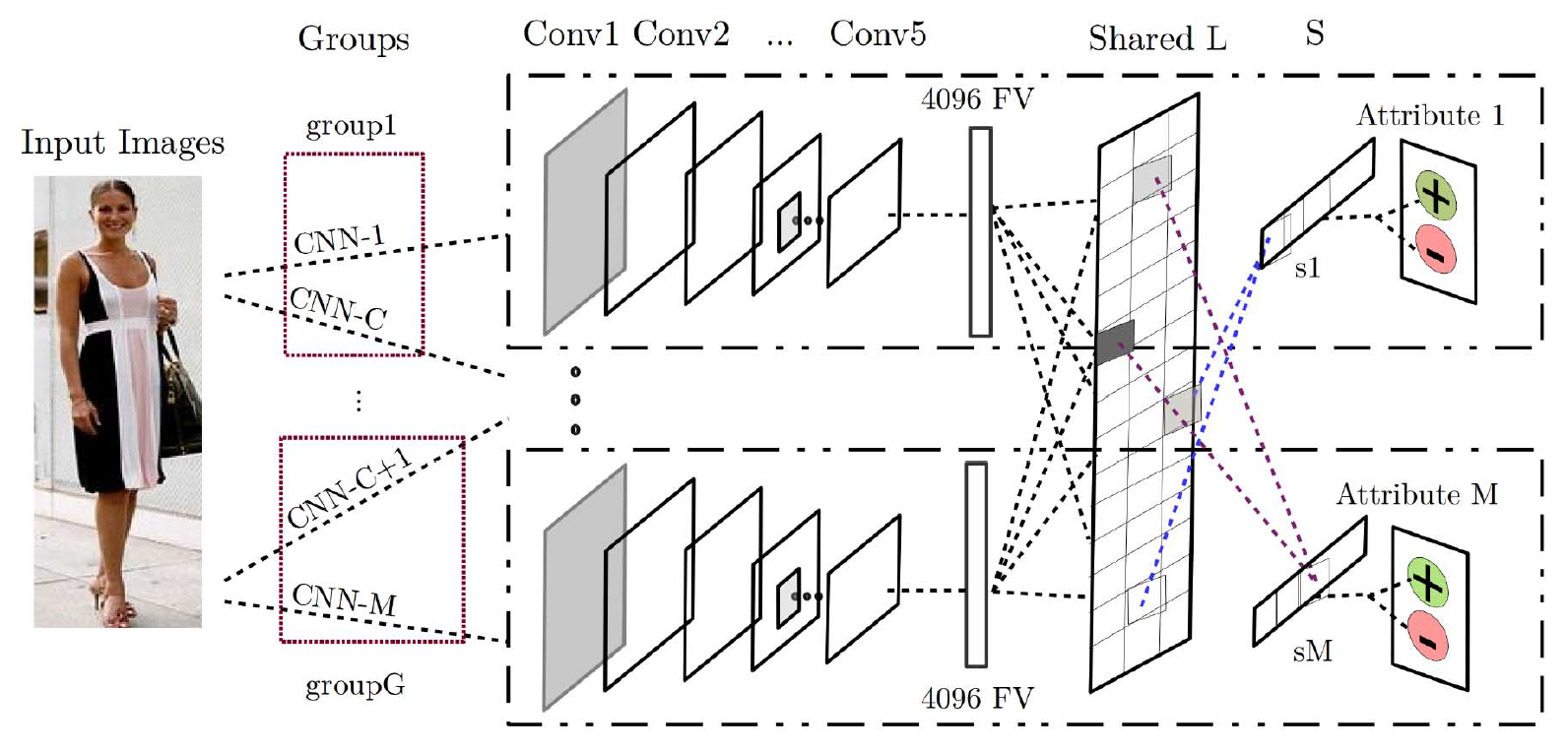}
\caption{The pipeline of Mutli-task CNN \cite{abdulnabi2015multitaskCNN}.}
\label{DeepAlgorithms-MTCNN}
\end{figure}	

This paper proposes a joint multi-task learning algorithm for attribute estimation using CNN, named MTCNN, as shown in Fig. \ref{DeepAlgorithms-MTCNN}. The MTCNN lets the CNN models share visual knowledge among different attribute categories. They adopt multi-task learning on the CNN features to estimate corresponding attributes. In their MTL framework, they also use rich information groups because knowing any a priori information about the statistical information of features will definitely aid the classifiers. They use the decomposition method to obtain shareable latent task matrix $L$ and combination matrix $S$ from total classifier weights matrix $W$, and thus flexible global sharing and competition between groups through learning localized features, \emph{i.e.}, $W = LS$. Therefore, the objective function (named \emph{MTL squared maxing hinge loss}) is formulated as follows: 
\begin{equation}
\label{MTCNN_objectFunction} 
\begin{small}
\begin{aligned}
\min_{L, S} \sum_{m=1}^{M} \sum_{i=1}^{N_m} \frac{1}{2} [max(0, 1 - Y^{i}_{m} (Ls^m)^T X^i_m)]^2 + \\ 
						\mu \sum_{k=1}^{K} \sum_{g=1}^{G} ||s^g_k||_2 + \gamma ||L||_1 + \lambda ||L||_F^2
\end{aligned}
\end{small}
\end{equation}
where $(X^i_m,  Y^{i}_{m})_{i=1}^{N_m}$ is the training data, $N_m$ is the number of training samples of the $m^{th}$ attribute. $K$ is the total latent task dimension space. The model parameter of $m^{th}$ attribute category is denoted as $Ls^m$. 

They employ the Accelerate Proximal Gradient Descent (APG) algorithm to optimize both $L$ and $S$ in an alternating manner. Therefore, the overall model weight matrix $W$ can be obtained after obtaining the $L$ and $S$.

\textbf{Summary:} According to the aforementioned algorithms \cite{sudowe2015person} \cite{acprli2015DeepMAR} \cite{acprli2015DeepMAR} \cite{abdulnabi2015multitaskCNN}, we can find that these algorithms all take the whole images as input and conduct multi-task learning for PAR. They all attempt to learn more robust feature representations using feature sharing, end-to-end training or multi-task learning squared maxing hinge loss. The benefits of these models are simple, intuitive, and highly efficient which are very important for practical applications. However, the performance of these models is still limited due to the lack of consideration of fine-grained recognition.

\subsection{Part-based Models}\label{partbasedPAR}
In this subsection, we will introduce the part-based algorithms that could jointly utilize local and global information for more accurate PAR. The algorithms are including: Poselets \cite{bourdev2011describing}, RAD \cite{joo2013humanRAD}, PANDA \cite{zhang2014panda}, MLCNN \cite{zhu2015multilabelCNN}, AAWP \cite{Gkioxari_2015_ICCV},  ARAP \cite{BMVC2016_ARAP}, DeepCAMP \cite{diba2016deepcamp},  PGDM \cite{ICMEli2018pose}, DHC \cite{li2016humanDHContexts}, LGNet \cite{liulocalizationBMVC2018}.

\subsubsection{Poselets (ICCV-2011) \cite{bourdev2011describing}} 
The motivation of this paper is that we can train attribute classifiers simply if we can isolate image patches corresponding to the same body part from the same viewpoint. However, directly using object detectors was not reliable for body parts localization at that time (before the year 2011) due to its limited ability. Therefore, the authors adopt the \emph{poselets} \cite{bourdev2009poselets} to decompose the image into a set of parts, each capturing a salient pattern corresponding to a given viewpoint and local pose. This provides a robust distributed representation of a person from which attributes can be inferred without explicitly localizing different body parts. 

\begin{figure}[htb]
\center
\includegraphics[width=3.5in]{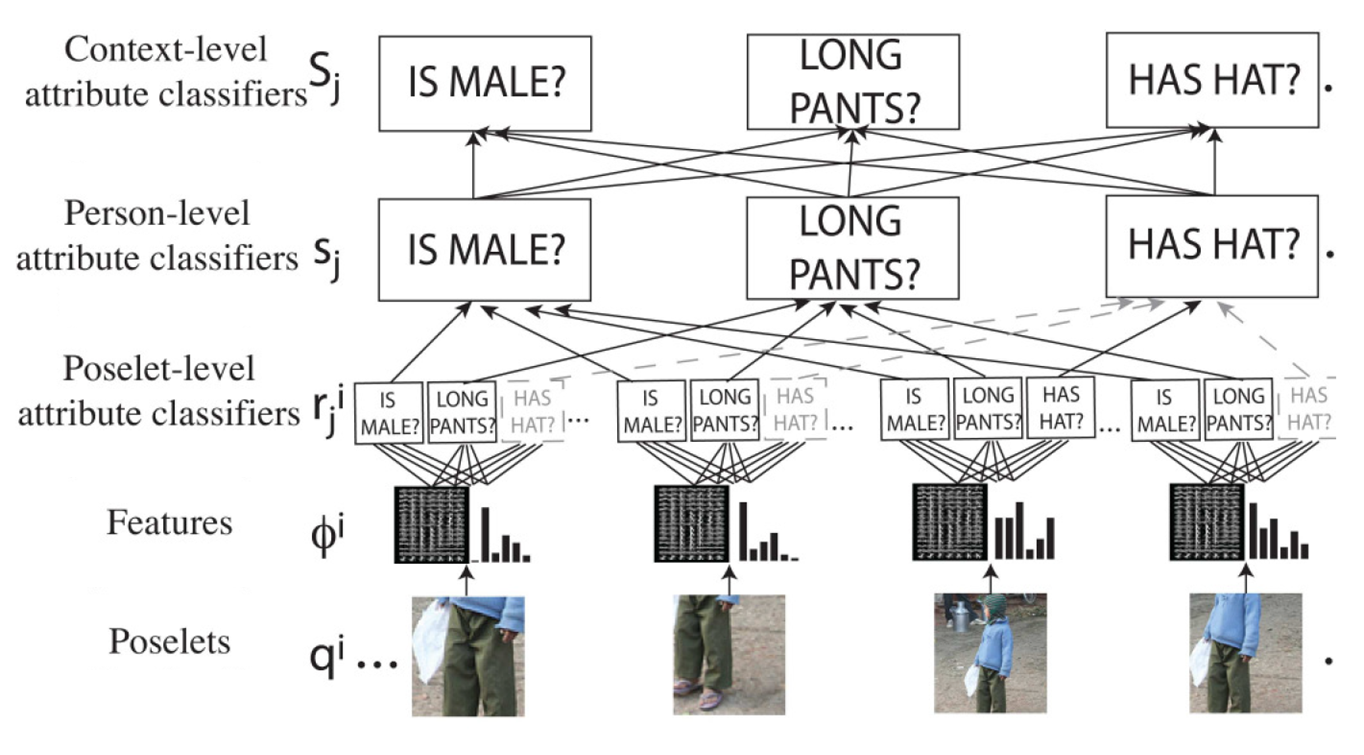}
\caption{The overview of Poselets at test time \cite{bourdev2011describing}.}
\label{DeepAlgorithms-Poselets}
\end{figure}

The Fig. \ref{DeepAlgorithms-Poselets} illustrates the overall pipeline of Poselets. Specifically speaking, they first detect the poselets on a given image and obtain their joint representations by concatenating the HOG, color histogram, and skin-mask features. Then, they train multiple SVM classifiers which are used for \emph{poselet-level, person-level, context-level} attribute classification, respectively. The poselet-level classifiers target to determine the presence of an attribute from a given part of the person under a given viewpoint. The person-level classifiers are used to combine the evidence from all body parts and the context-level classifiers take the output of all person-level classifiers as input and try to exploit the correlations between the attributes. Their attribute prediction results are the output of context-level classifiers. 

The idea of \emph{poselets} is also extended by combining it with a deep neural network \cite{bourdev2014deepposelet}, named \emph{deep poselets}. It can be used for human body parts localization-based tasks, such as human detection tasks \cite{jammalamadaka2015human}.

\subsubsection{RAD (ICCV-2013) \cite{joo2013humanRAD}}
\begin{figure}[htb]
\center
\includegraphics[width=3.5in]{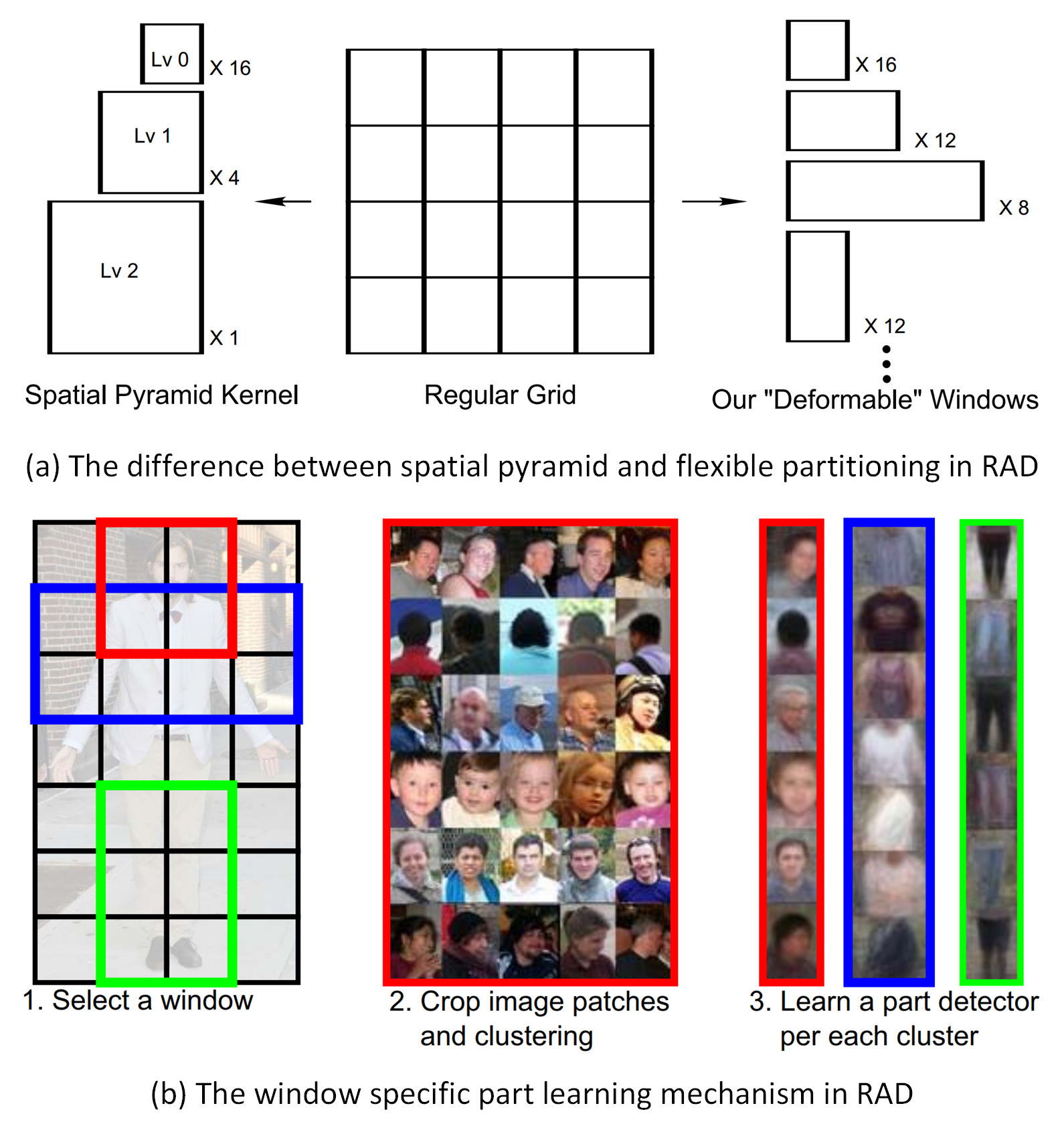}
\caption{(a) The difference between RAD and spatial pyramid; (b) The window specific part learning in RAD.}
\label{DeepAlgorithms-RAD}
\end{figure}

This paper proposes a part learning algorithm from the perspective of appearance variance while previous works focus on handling geometric variation which requires manual part annotation, such as poselet \cite{bourdev2011describing}. They first divide the image lattice into a number of overlapping subregions (named \emph{window}). As shown in Fig. \ref{DeepAlgorithms-RAD} (a), a grid of size $W \times H$ is defined and any rectangle on the grid containing one or more number of cells of the grid forms a window. The proposed method is more flexible in shape, size, and location of the part window while previous works (such as spatial pyramid matching structure, SPM \cite{lazebnik2006beyondSPM}) recursively divide the region into four quadrants and make all subregions squares that do not overlap with each other at the same level. 

With all these windows, they learn a set of part detectors that are spatially associated with that particular window. For each window, all corresponding image patches are cropped from training images and represented by HOG \cite{dalal2005histograms} and color histogram feature descriptors. Then, K-means clustering is conducted based on the extracted features. Each obtained cluster denotes a specific appearance type of a part. They also train a local part detector for each cluster by logistic regression as an initial detector and iteratively refine it by applying it to the entire set again and updating the best location and scale to handle the issue of noisy clusters. 

After learning the parts at multi-scale overlapping windows, they follow the method for attribute classification proposed in the Poselet-based approach \cite{bourdev2011describing}. Specifically, they aggregate the scores from these local classifiers with the weights given by part detection scores for final prediction.

\subsubsection{PANDA (CVPR-2014) \cite{zhang2014panda}} 
\begin{figure}[htb]
\center
\includegraphics[width=3.3in]{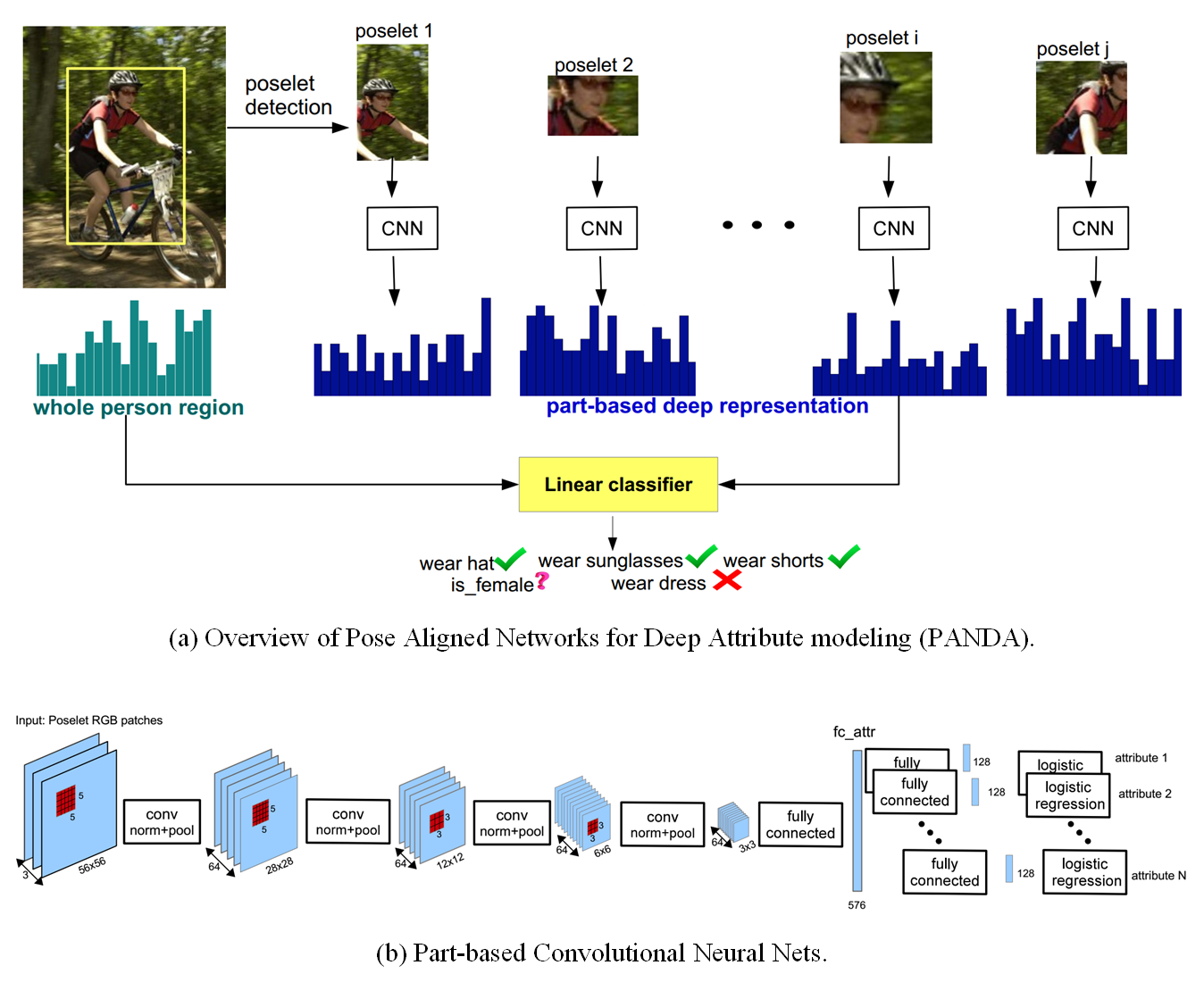}
\caption{Overview of Pose Aligned Networks for Deep Attribute Modeling (PANDA). One convolutional neural net is trained on semantic part patches for each poselet and then the top-level activations of all nets are concatenated to obtain a pose-normalized deep representation. The final attributes are predicted by a linear SVM classifier using the pose-normalized representations. This figure is adapted from PANDA \cite{zhang2014panda}.}
\label{DeepAlgorithms-panda}
\end{figure}

Zhang \emph{et al.} find the signal associated with some attributes is subtle and the image is dominated by the effects of pose and viewpoint. For the attribute of \emph{wear glasses}, the signal is weak at the scale of the full person and the appearance varies significantly with the head pose, frame design, and occlusion by the hair. They think the key to accurately predicting the underlying attributes lies in locating object parts and establishing their correspondences with model parts. They propose to jointly use global image and local patches for person attribute recognition and the overall pipeline can be found in Fig. \ref{DeepAlgorithms-panda} (a) and (b).

As shown in Fig. \ref{DeepAlgorithms-panda} (a), they first detect the poselets \cite{bourdev2011describing} and obtain parts of the person. Then, they adopt the CNN to extract the feature representations of the local patches and the whole human image. If the poselet is not detected, they simply leave the feature to zero. Hence, their model could leverage both the power of convolutional nets for learning discriminative features from data and the ability of poselets to simplify the learning task by decomposing the objects into their canonical poses. They directly feed the combined local and global features into the linear classifier which is an SVM (Support Vector Machine) for multiple attributes estimation.

Fig. \ref{DeepAlgorithms-panda} (b) illustrates the detailed architecture, this network takes the poselet RGB patch $56 \times 56 \times 3$ as input and outputs the response score of each attribute with corresponding fully connected layer (fc layer). The feature extraction module contains four groups of convolutional/pooling/normalization layers and the output of these groups are $28 \times 28 \times 64$, $12 \times 12 \times 64$, $6 \times 6 \times 64$ and $3 \times 3 \times 64$, respectively. Then the input image is mapped into a feature vector whose dimension is 576-D with a fully connected layer. They set a fc layer for each attribute whose output dimension is $128$. 

The advantage of this work, it adopts deep features rather than shallow low-level features which can obtain more powerful feature representation than previous works. In addition, it also processes the human image from the perspective of local patches and global images, which can mine more detailed information than those works that only consider the whole image. These two points all improve the person's attribute recognition significantly. However, we think the following issues may limit the final performance of their procedure: 1). the parts localization, i.e. the accuracy of poselets, which may be the bottleneck of their results; 2). they do not use an end-to-end learning framework for the learning of deep features; 3). Their poselet also contains background information, which may also influence the feature representation.

\subsubsection{MLCNN (ICB-2015) \cite{zhu2015multilabelCNN}} 

\begin{figure}[htb]
\center
\includegraphics[width=3.5in]{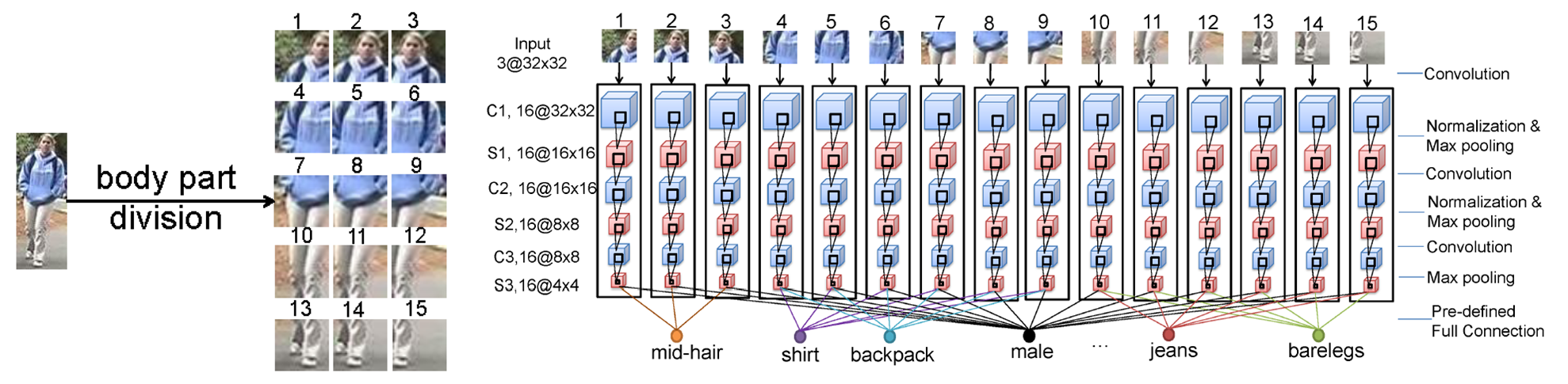}
\caption{The pipelines of MLCNN \cite{zhu2015multilabelCNN}.}
\label{DeepAlgorithms-MLCNN}
\end{figure}	

This paper proposes a multi-label convolutional neural network to predict multiple attributes together in a unified framework. The overall pipeline of their network can be found in Fig. \ref{DeepAlgorithms-MLCNN}. They divide the whole image into 15 overlapping patches and use a convolutional network to extract its deep features. They adopt corresponding local parts for specific attribute classification, such as the patch $1, 2, 3$ are used for hairstyle estimation. They utilize the softmax function for each attribute prediction. 

In addition, they also use the predicted attributes to assist person re-identification. Specifically, they fuse the low-level feature distance and attribute-based distance as the final fusion distance to discriminate whether given images have the same identity.

\subsubsection{AAWP (ICCV-2015) \cite{Gkioxari_2015_ICCV}} 
\begin{figure}[htb]
\center
\includegraphics[width=3.5in]{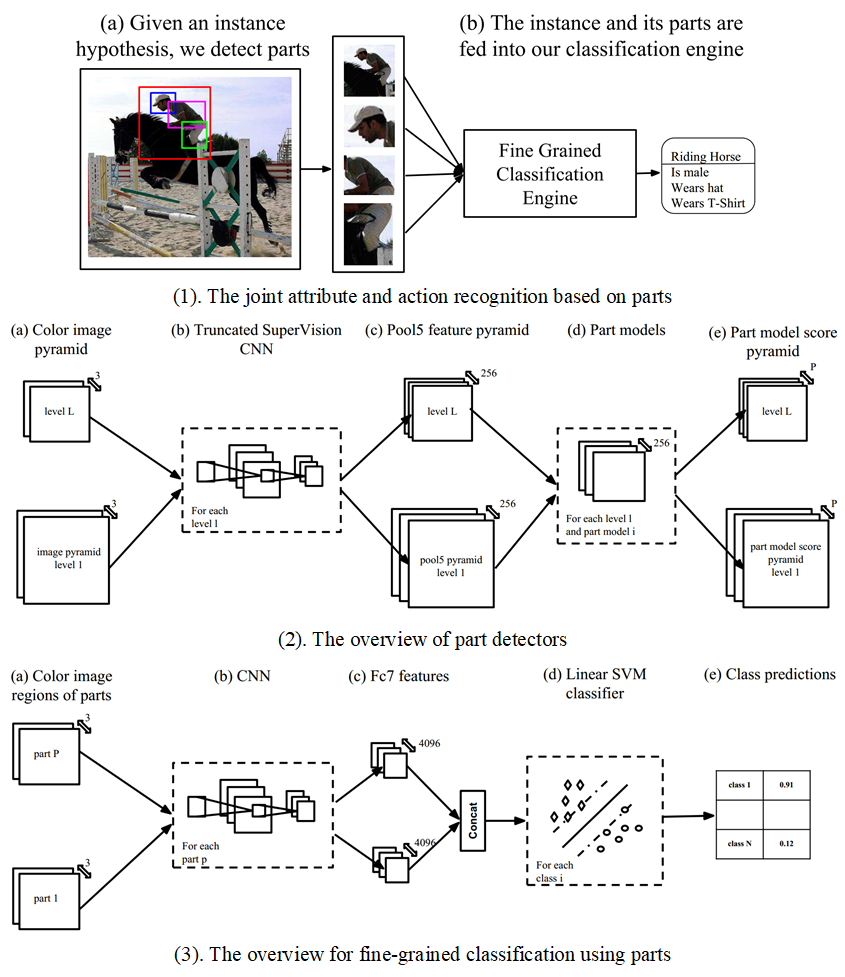}
\caption{The pipelines of joint action and attribute recognition using parts, part detector, and fine-grained classification module, respectively. This figure is adapted from AAWP \cite{Gkioxari_2015_ICCV}.}
\label{DeepAlgorithms-AAWP}
\end{figure}

The AAWP is introduced to validate whether parts could bring improvements in both action and attribute recognition. As shown in Fig. \ref{DeepAlgorithms-AAWP} (1), the CNN features are computed on a set of bounding boxes associated with the instance to classify, \emph{i.e.} the whole instance, the oracle or person detector provided and poselet-like part detector provided. The authors define three human body parts (head, torso, and legs) and cluster the key points of each part into several distinct poselets. This part detectors are named \emph{deep} version of poselets due to the utilization of deep feature pyramid, rather than low-level gradient orientation features used in traditional poselets \cite{bourdev2010detecting, bourdev2011describing}. In addition, the authors also introduce task-specific CNN fine-tuning and their experiments show that a fine-tuned holistic model (\emph{i.e.} no parts) could already achieve comparable performance with a part-based system like PANDA \cite{zhang2014panda}. Specifically, the whole pipeline can be divided into two main modules, \emph{i.e.} the part detector module and fine-grained classification module, as shown in Fig. \ref{DeepAlgorithms-AAWP} (2) and (3) respectively.

For the part detector module, they design their network by following the object detection algorithm RCNN \cite{girshick2015deformable} which contains two stages, \emph{i.e.} the feature extraction and part classification. They adopt a multi-scale fully convolutional network to extract the image features. More specifically, they first construct a color image pyramid and obtain the pool5 feature for each pyramid level. Then, they adopt part models to obtain the corresponding score, as shown in Fig. \ref{DeepAlgorithms-AAWP} (2). Therefore, the key problem lies in how to achieve accurate part localization given these feature map pyramids. To handle the localization of parts, the authors designed three body areas (head, torso, and legs) and trained part detectors with linear SVMs. The positive training data is collected from PASCAL VOC 2012 with a clustering algorithm. In the testing phase, they keep the highest scoring part within a candidate region box in an image. 

For the task of part-based classification discussed in their paper, \emph{i.e.} the action and attribute recognition. They consider four different approaches to understanding which design factors are important, \emph{i.e.} no parts, instance fine-tuning, joint fine-tuning, and 3-way split. The detailed pipeline for the fine-grained classification can be found in Fig. \ref{DeepAlgorithms-AAWP} (3). Given the image and detected parts, they use CNN to obtain fc7 features and concatenate them into one feature vector as its final representation. Therefore, the action or attribute category can be estimated with a pre-trained linear SVM classifier. Their experiments on the PASCAL VOC action challenge and Berkeley attributes of people dataset \cite{bourdev2011describing} validated the effectiveness of part. In addition, they also find that as more powerful convolutional network architectures are engineered, the marginal gain from explicit parts may vanish. They think this might be because of the already high performance achieved by the holistic network.

This work further expands and validates the effectiveness and necessity of parts in a wider way. It also shows more insights into deep learning-based human attributes recognition.

\subsubsection{ARAP (BMVC2016) \cite{BMVC2016_ARAP}}

\begin{figure}[htb]
\center
\includegraphics[width=3.5in]{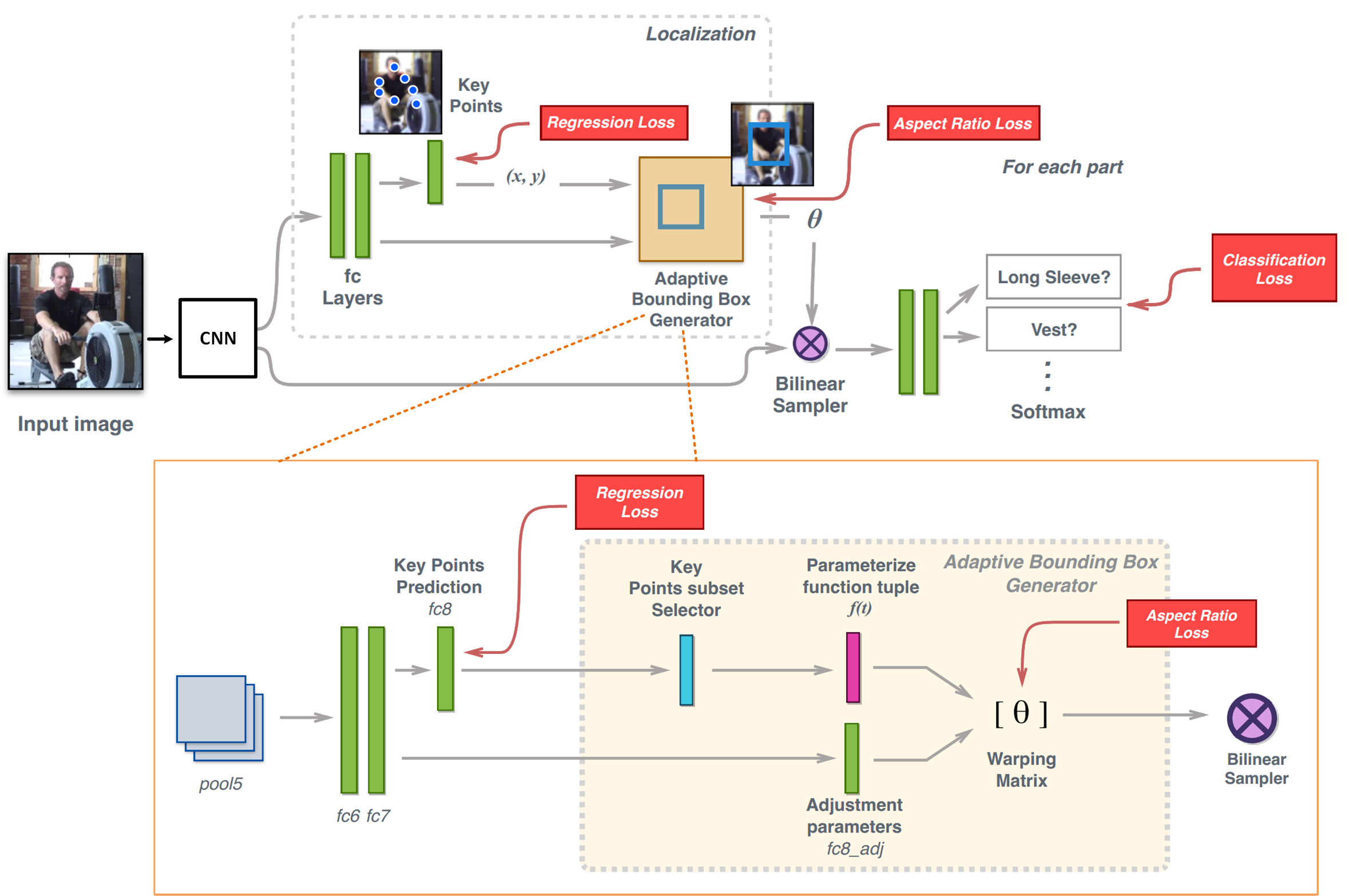}
\caption{The pipelines of ARAP \cite{BMVC2016_ARAP}.}
\label{DeepAlgorithms-ARAP}
\end{figure}

This paper adopts an end-to-end learning framework for joint part localization and multi-label classification for person attribute recognition. As shown in Fig. \ref{DeepAlgorithms-ARAP}, the ARAP contains the following sub-modules: initial convolutional feature extraction layers, a key point localization network, an adaptive bounding box generator for each part, and the final attribute classification network for each part. Their network contains three loss functions, \emph{i.e.} the regression loss, aspect ratio loss, and classification loss. 

Specifically speaking, they first extract the feature map of the input image, then conduct key points localization. Given the key points, they divide the human body into three main regions (including hard, torso, and legs) and obtain an initial part bounding box. On the other hand, they also take the previous fc7 layer's features as input and estimate the bounding box adjustment parameters. Given these bounding boxes, they adopt a bilinear sampler to extract corresponding local features. Then, the features are fed into two FC layers for multi-label classification.

\subsubsection{DeepCAMP (CVPR-2016) \cite{diba2016deepcamp}} 

\begin{figure*}[htb]
\center
\includegraphics[width=7.2in]{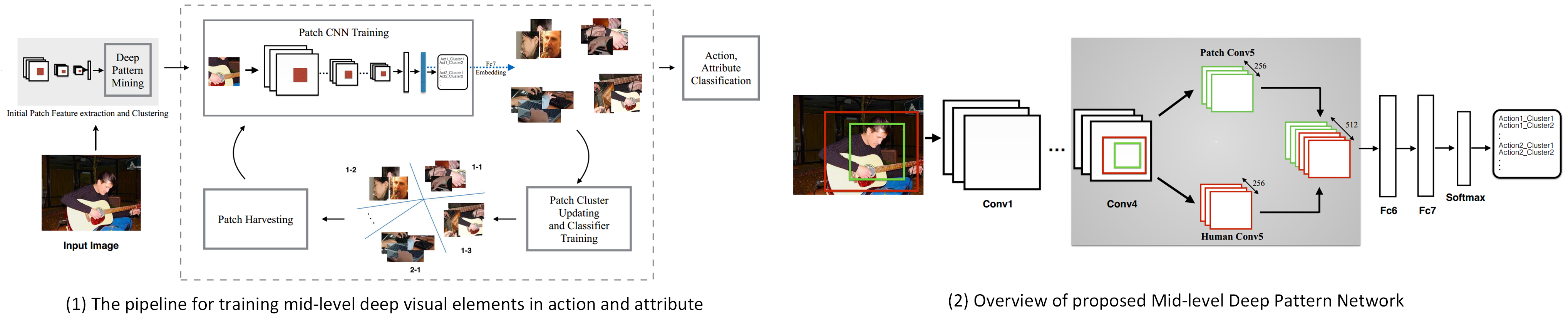}
\caption{The overview of DeepCAMP \cite{diba2016deepcamp}.}
\label{DeepAlgorithms-DeepCAMP}
\end{figure*}

This paper proposes a novel CNN that mines mid-level image patches for fine-grained human attribute recognition. Specifically, they train a CNN to learn discriminative patch groups, named \emph{DeepPattern}. They utilize regular contextual information (see Fig. \ref{DeepAlgorithms-DeepCAMP} (2)) and also let an iteration of feature learning and patch clustering purify the set of dedicated patches, as shown in Fig. \ref{DeepAlgorithms-DeepCAMP} (1).

The main insight of this paper lies in that a better embedding can help improve the quality of the clustering algorithm in pattern mining algorithm \cite{li2015midlevelpattern}. Therefore, they propose an iteration algorithm where in each iteration, they train a new CNN to classify cluster labels obtained in the previous iteration to help improve the embedding. On the other hand, they also concatenate features from both the local patch and global human bounding box to improve the clusters of mid-level elements.

\subsubsection{PGDM (ICME-2018) \cite{ICMEli2018pose}} 

\begin{figure}[htb]
\center
\includegraphics[width=3.3in]{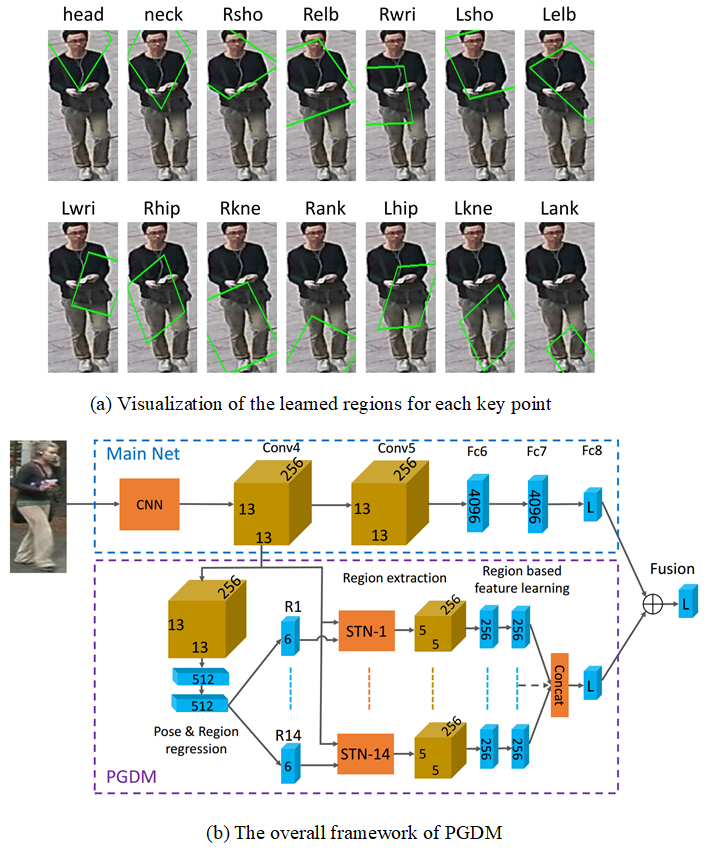}
\caption{The visualization of pose estimation and pipeline of PGDM. This figure is adapted from PGDM \cite{ICMEli2018pose}.}
\label{DeepAlgorithms-PGDM}
\end{figure}

The PGDM is the first work that attempts to explore the structure knowledge of the pedestrian body (\emph{i.e.} pedestrian pose) for person attributes learning.  They first estimate the key points of given human image using a pre-trained pose estimation model. Then, they extract the part regions according to these key points. The deep features of part regions and whole images are all extracted and used for attribute recognition independently. These two scores are then fused together to achieve final attribute recognition. The visualization of pose estimation and the whole pipeline of PGDM can be found in Fig. \ref{DeepAlgorithms-PGDM} (a) and (b) respectively.

As shown in Fig. \ref{DeepAlgorithms-PGDM} (b), the attribute recognition algorithm contains two main modules: \emph{i.e.} the main net and PGDM. The main net is a modification of AlexNet \cite{krizhevsky2012imagenet}, the fc8 layer is set as same with attribute number. It takes attribute recognition as a multi-label classification problem and adopts the improved cross-entropy loss \cite{acprli2015multi} as its objective function.

The PGDM module, its target is to explore the deformable body structure knowledge to assist pedestrian attribute recognition. The authors resort to deep pose estimation models rather than re-annotating human pose information of training data. And they embed existing pose estimation algorithms into their attribute recognition model instead of using it as an external one. They directly train a regression network to predict the pedestrian pose with coarse ground truth pose information obtained from existing pose estimation model \footnote{The human key points list used in PGDM are: head, neck, right shoulder, right elbow, right wrist, left shoulder, left elbow, left wrist, right hip, right knee, right ankle, left hip, left knee, left ankle.}. Once the pose information is obtained, they transform the key points into informative regions using spatial transformer network (STN) \cite{jaderberg2015spatial}. Then, they use an independent neural network for feature learning from each key point-related region. They jointly optimize the main net, PGDM, and pose regression network.

\subsubsection{DHC (ECCV-2016) \cite{li2016humanDHContexts}}

\begin{figure*}[htb]
\center
\includegraphics[width=6.5in]{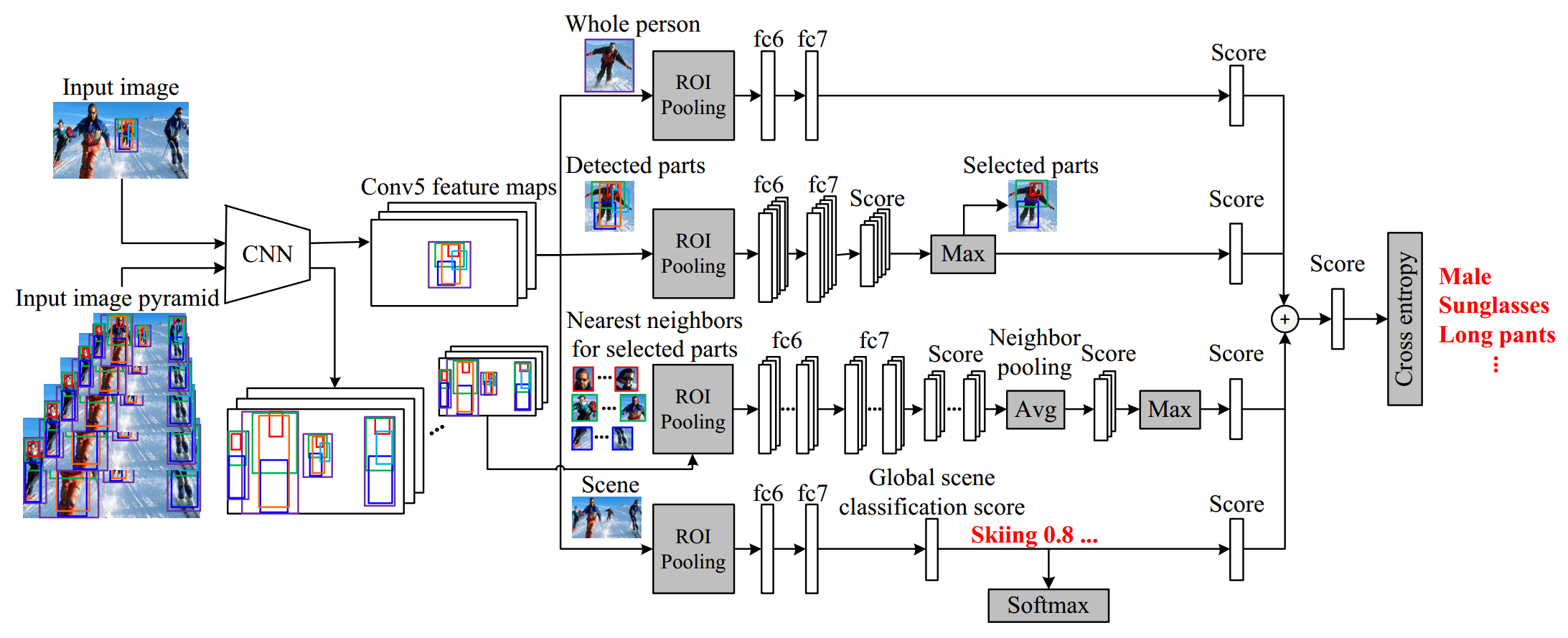}
\caption{The pipeline of DHC-network. This figure is adapted from DHC \cite{li2016humanDHContexts}.}
\label{DeepAlgorithms-DHC}
\end{figure*}

This paper proposes to use \emph{deep hierarchical contexts} to help person attribute recognition due to the background would sometimes provide more information than the target object only. Specifically, the \emph{human-centric context} and \emph{scene context} are introduced in their network architecture. As shown in Fig. \ref{DeepAlgorithms-DHC}, they first construct an input image pyramid and pass them all through CNN (the VGG-16 network is used in this paper) to obtain multi-scale feature maps. They extract features of four sets of bounding box regions, \emph{i.e.} the whole person, detected parts of the target object, nearest neighbor parts from the image pyramid, and global image scene. The first two branches (the whole person and parts) are the regular pipelines for the person attribute recognition algorithm. The main contributions of this paper lie in the latter two branches, \emph{i.e.} the human-centric and scene-level contexts to help improve the recognition results. Once the scores of these four branches are obtained, they sum up all the scores as the final attribute score.

Due to the use of context information, this neural network needs more external training data than regular pedestrian attribute recognition tasks. For example, they need to detect the part of the human body (head, upper, and bottom body regions) and recognize the style/scene of a given image. They propose a new dataset named \emph{WIDER}, to better validate their ideas. Although the human attribute recognition results can be improved significantly via this pipeline, however, this model looks a little more complex than other algorithms.

\subsubsection{LGNet (BMVC-2018) \cite{liulocalizationBMVC2018}} 
\begin{figure}[htb]
\center
\includegraphics[width=3.3in]{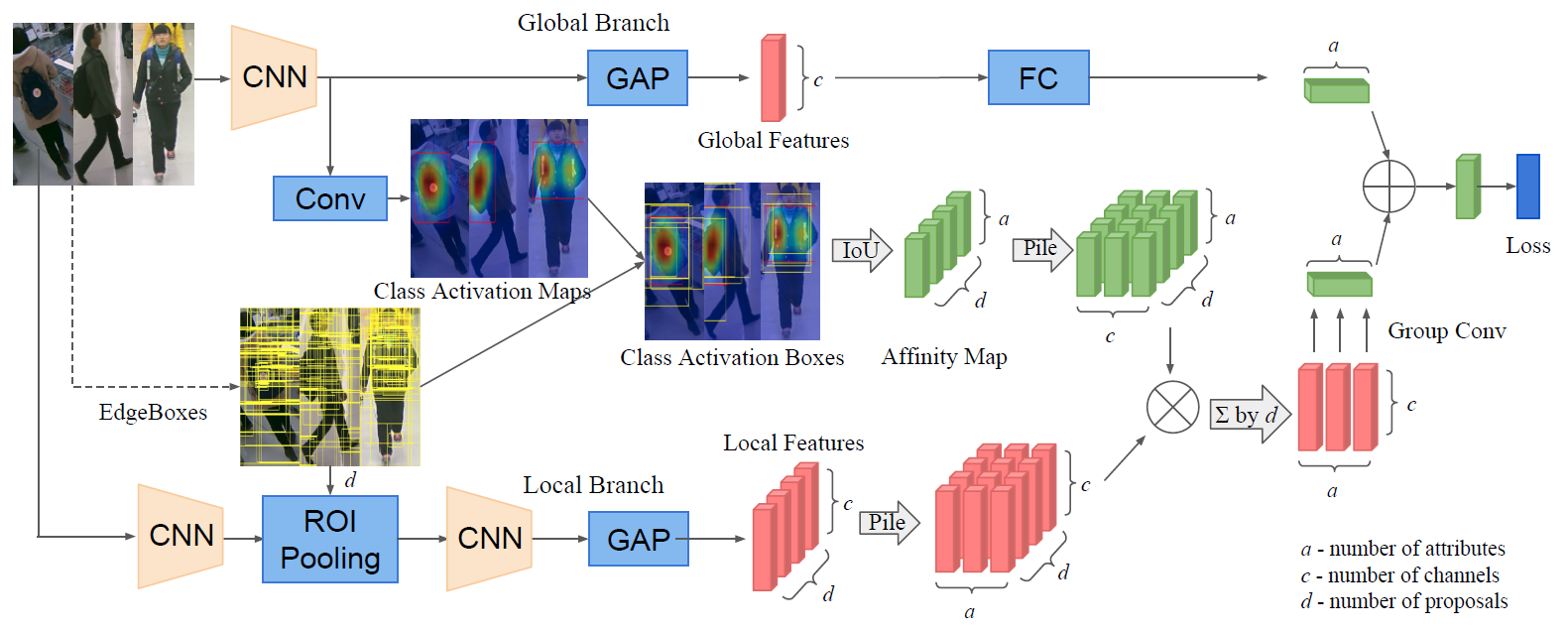}
\caption{The visualization of pose estimation and pipeline of LGNet. This figure is adapted from LGNet \cite{liulocalizationBMVC2018}.}
\label{DeepAlgorithms-LGNet}
\end{figure}

This paper proposes a Localization Guide Network (named LGNet) which could localize the areas corresponding to different attributes. It also follows the local-global framework, as shown in Fig. \ref{DeepAlgorithms-LGNet}. Specifically, they adopt Inception-v2 \cite{ioffe2015batchNormalization} as their basic CNN model for feature extraction. For the global branch, they adopt the global average pooling layer (GAP) to obtain its global features. Then, a fully connected layer is utilized to output its attribute predictions. For the local branch, they use the $1 \times 1$ convolution layer to produce $c$ class activation maps for each image, where $c$ is the number of attributes in the used dataset. Given the class activation maps, they can capture an activation box for each attribute by cropping the high-response areas of the corresponding activation map. They also use EdgeBoxes \cite{zitnick2014edgeBoxes} to generate region proposals to obtain local features from the input image. In addition, they also consider the different contributions of extracted proposals and different attributes should focus on different local features. Therefore, they use the class active map for each attribute to serve as a guide to determine the importance of the local features to different attributes. More specifically, they compute the spatial affinity map between the proposals and class activation boxes according to the Interaction over Union (IoU) and linearly normalized to weight the local feature vectors for further predictions. Finally, the global and attended local features are fused together by element-wise sum for pedestrian attribute prediction. 

\textbf{Summary:} On the basis of the reviewed papers \cite{bourdev2011describing} \cite{joo2013humanRAD} \cite{zhang2014panda}  \cite{zhu2015multilabelCNN} \cite{Gkioxari_2015_ICCV} \cite{BMVC2016_ARAP} \cite{diba2016deepcamp} \cite{ICMEli2018pose} \cite{li2016humanDHContexts} \cite{liulocalizationBMVC2018}, it is easy to find that these algorithms all adopt the joint utilization of global and fine-grained local features. The localization of body parts is achieved via an external part localization module, such as part detection, pose estimation, poselets, or proposal generation algorithm. The use of part information improves the overall recognition performance significantly. At the same time, it also brings some shortcomings as follows: Firstly, as an operation in the middle phase, the final recognition performance heavily relies on the accuracy of part localization. In other words, the inaccurate part detection results will bring the wrong features for final classification. Secondly, it will also need more training or inference time due to the introduction of human body parts. Thirdly, some algorithms need manual annotated labels about part locations which further increases the cost of manpower and money.

\subsection{Attention-based Models}

In this section, we will talk about person attribute recognition algorithms using attention mechanisms, such as HydraPlus-Net \cite{liu2017hydraplus}, VeSPA \cite{sarfraz2017deep}, DIAA \cite{sarafianos2018deep}, CAM \cite{guo2017humanPRL}. 

\subsubsection{HydraPlus-Net (ICCV-2017) \cite{liu2017hydraplus}}
\begin{figure}[htb]
\center
\includegraphics[width=3.3in]{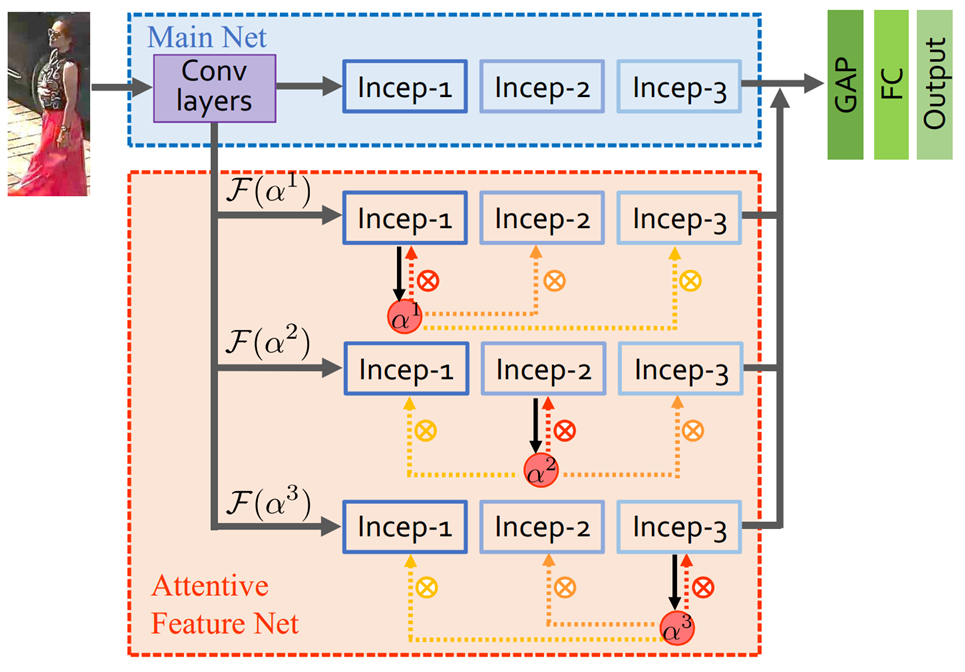}
\caption{Some visualizations of HydraPlus-Net. This figure is adapted from HydraPlus-Net \cite{liu2017hydraplus}.}
\label{DeepAlgorithms-HPNet}
\end{figure}

HPNet is introduced to encode multi-scale features from multiple levels for pedestrian analysis using multi-directional attention (MDA) modules. As shown in Fig. \ref{DeepAlgorithms-HPNet} (2), it contains two main modules \emph{i.e.} the Main Net (M-net) which is a regular CNN, and the Attentive Feature Net (AF-net) which includes multiple branches of multi-directional attention modules applied to different semantic feature levels. The AF-net and M-net share the same basic convolution architectures and their outputs are concatenated and fused by global average pooling (GAP) and fc layers. The output layer can be the attribute logits for attribute recognition or feature vectors for person re-identification. The authors adopt inception-v2 \cite{ioffe2015batch} as their basic network.

A specific illustration of AF-net can be found in Fig. \ref{DeepAlgorithms-HPNet} (4). Given the feature maps of black 1, 2, and 3, they conduct $1 \times 1$ convolution operation on feature map 2 and obtain its attention map $\alpha^2$. It is worth noting that, this attention module is different from previous attention-based models which only push the attention map back to the same block. They not only use this attention map to attend to feature map 2 but also attend to the adjacent features, such as feature maps 1 and 3. Applying one single attention map to multiple blocks naturally lets the fused features encode multi-level information within the same spatial distribution, as shown in Fig. \ref{DeepAlgorithms-HPNet} (3).

The HP-net is trained in a stage-wise manner, in other words, the M-net, AF-net, and remaining GAP and fc layers are trained in a sequential way. The output layer is used to minimize the cross-entropy loss and softmax loss for person attribute recognition and person re-identification respectively.

\subsubsection{VeSPA (ArXiv-2017) \cite{sarfraz2017deep}} 

\begin{figure}[htb]
\center
\includegraphics[width=3.3in]{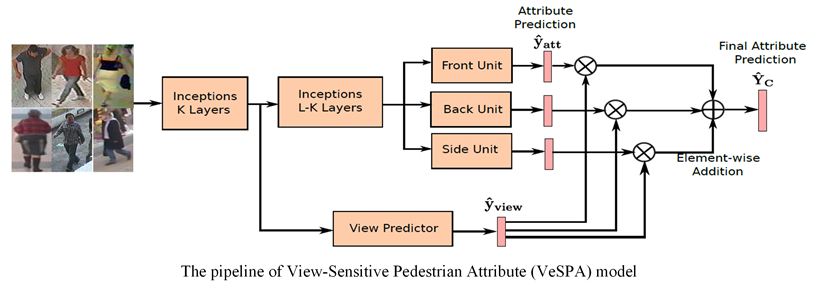}
\caption{The pipeline of VeSPA model. This figure is adapted from the VeSPA model \cite{sarfraz2017deep}.}
\label{DeepAlgorithms-VeSPA}
\end{figure}	

The VeSPA takes the view cues into consideration to better estimate the corresponding attribute. The authors find that the visual cues hinting at attributes can be strongly localized and inference of personal attributes such as hair, backpack, shorts, etc., are highly dependent on the acquired view of the pedestrian. As shown in Fig. \ref{DeepAlgorithms-VeSPA}, the image is fed into the Inceptions (K layers) and obtains its feature representation. The view-specific unit is introduced to mapping the feature maps into coarse attribute prediction $\hat{y}_{att} = [y^1, y^2, ..., y^c ]^T$. Then, a view predictor is used to estimate the view weights $\hat{y}_{view}$. The attention weights are used to multiply view-specific predictions and obtain the final multi-class attribute prediction $\hat{Y}_c = [ y^1, y^2, ..., y^C ]^T$.

The view classifier and attribute predictors are trained with separate loss functions. The whole network is a unified framework and can be trained in an end-to-end manner.

\subsubsection{DIAA (ECCV-2018) \cite{sarafianos2018deep}} 

\begin{figure}[htb]
\center
\includegraphics[width=3.3in]{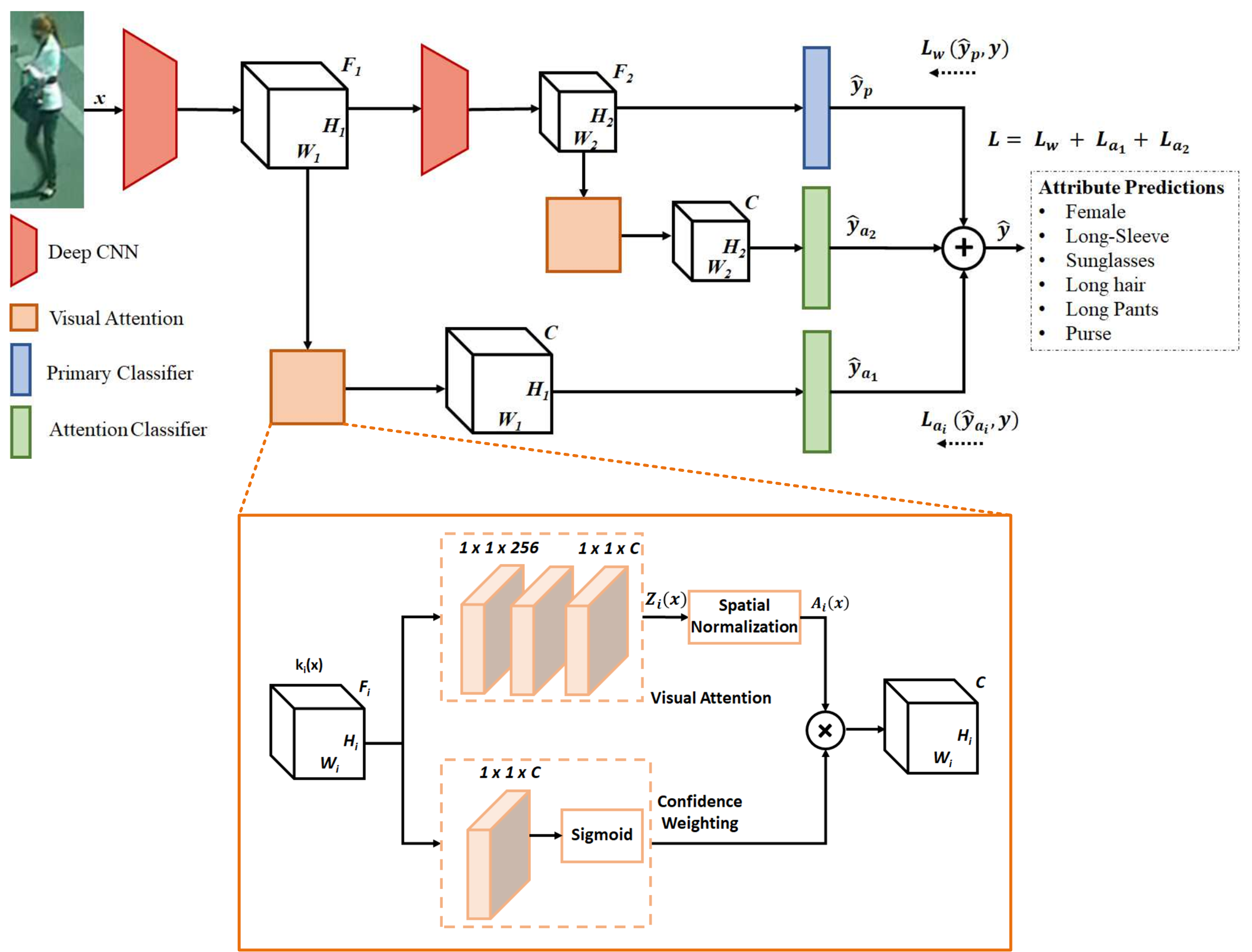}
\caption{The pipeline of DIAA model. This figure is adapted from the DIAA model \cite{sarafianos2018deep}.}
\label{DeepAlgorithms-DIAA}
\end{figure}

The DIAA algorithm can be seen as an ensemble method for person attribute recognition. As shown in Fig. \ref{DeepAlgorithms-DIAA}, their model contains the following modules: multi-scale visual attention and weighted focal loss for deep imbalanced classification. For the multi-scale visual attention, as we can see from Fig. \ref{DeepAlgorithms-DIAA}, the authors adopt feature maps from different layers. They propose the weighted focal loss function \cite{lin2018focalLoss} to measure the difference between predicted attribute vectors and ground truth:

\begin{equation}
\label{weightedFocalLossfunction}
\begin{small}
\begin{aligned}
\mathcal{L}_w (\hat{y}_p, y) = - \sum_{c=1}^{C} w_c ( (1-\sigma(\hat{y}_p^c))^\gamma log(\sigma(\hat{y}_p^c)) y^c + \\
                                                         \sigma(\hat{y}_p^c)^\gamma log(1-\sigma(\hat{y}^c_p)) (1-y^c)),
\end{aligned}
\end{small}
\end{equation}
where $\gamma$ is a parameter that is used to control the instance-level weighting based on the current prediction giving emphasis to the hard misclassified samples. $w_c = e^{-a_c}$ and $a_c$ is the prior class distribution of $c^{th}$ attribute following \cite{sarfraz2017deep}.

In addition, they also propose to learn the attention maps in a weakly supervised manner (only the attribute labels, no specific bounding box annotation) to improve the classification performance by guiding the network to focus its resources to those spatial parts that contain information relevant to the input image. As shown in the right part of Fig. \ref{DeepAlgorithms-DIAA}, the attention sub-network takes the feature map $W_i \times H_i \times F_i$ as input and output an attention mask with dimension $W_i \times H_i \times C$. The output is then fed to the attention classifier to estimate the pedestrian attributes. Due to limited supervised information for the training of the attention module, the authors resort to the prediction variance by following \cite{chang2017active}. Attention mask predictions with high standard deviation across time will be given higher weights in order to guide the network to learn those uncertain samples. They collect the history $H$ of the predictions for the $s^{th}$ sample and compute the standard deviation across time for each sample in a mini-batch. Hence, the loss for the attention map with attribute-level supervision for each sample $s$ can be obtained by:
\begin{equation}
\label{attenitonLossDIAA}
\mathcal{L}_{a_i} (\hat{y}_{a_i}, y) = (1 + std_s(H)) \mathcal{L}_b(\hat{y}_{a_i}, y)
\end{equation}
where $\mathcal{L}_b(\hat{y}_{a_i}, y)$ is binary cross entropy loss and $std_s(H)$ is the standard deviation.
Therefore, the total loss used to train this network end-to-end is the sum of loss from the primary network and the two attention modules.

\subsubsection{CAM (PRL-2017) \cite{guo2017humanPRL}} 

\begin{figure}[htb]
\center
\includegraphics[width=3.3in]{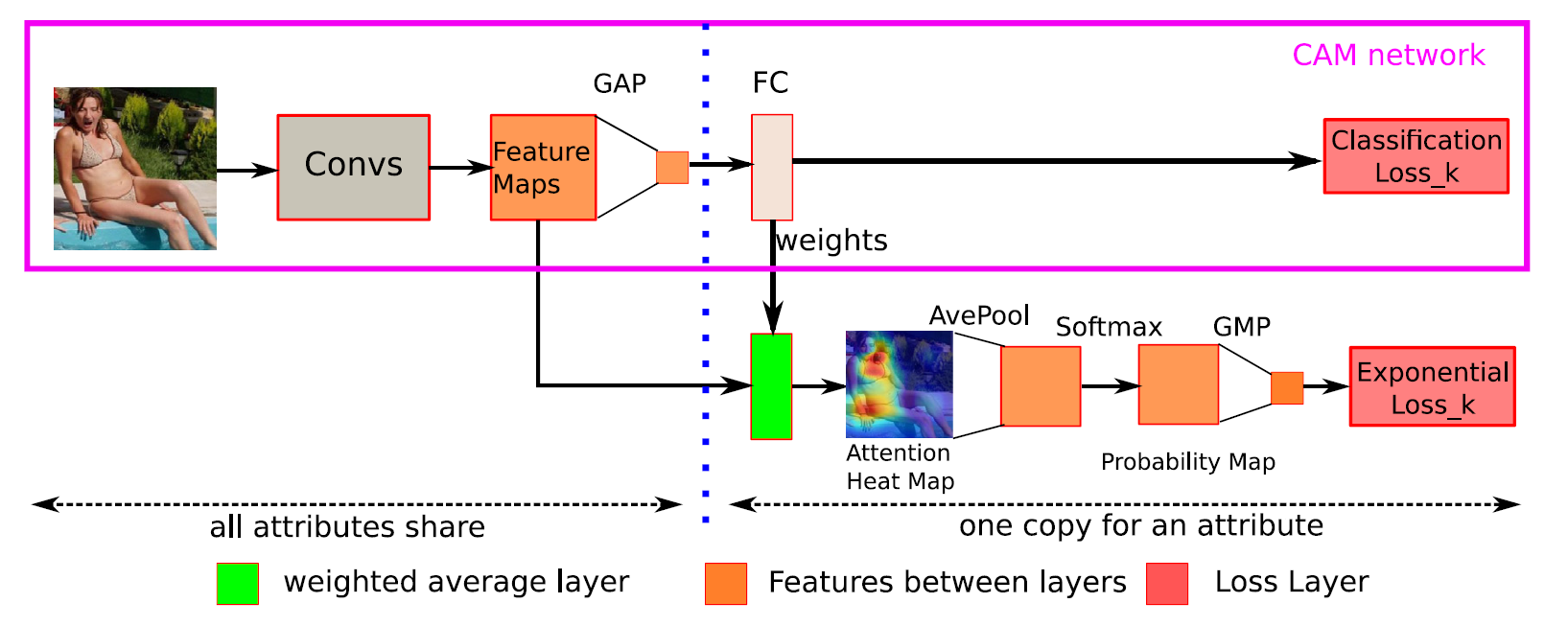}
\caption{The pipeline of CAM-network \cite{guo2017humanPRL}.}
\label{DeepAlgorithms-CAM}
\end{figure}	

In this paper, the authors propose to use and refine attention maps to improve the performance of person attribute recognition. As shown in Fig. \ref{DeepAlgorithms-CAM}, their model contains two main modules, \emph{i.e.} the multi-label classification sub-network and attention map refinement module. The adopted CAM net \cite{zhou2014learningCAMnet} also follows the category-specific framework, in other words, different attribute classifiers have different parameters for the fully connected (FC) layer. They use the parameters in the FC layer as weights to linearly combine the feature maps in the last convolutional layer to obtain the attention map of each object category. However, this naive implementation of the attention mechanism could not focus on the right regions all the time due to low resolution, over-fitting training, \emph{et al}. 

To handle the aforementioned issues, they explore refining the attention map by tuning the CAM network. They measure the appropriateness of an attention map based on its concentration and attempt to make the attention map highlight a smaller but concentrated region. Specifically speaking, they introduce a weighted average layer to obtain an attention map first. Then, they use average pooling to down-sample its resolution to capture the importance of all the potential relevant regions. After that, they also adopt the Softmax layer to transform the attention map into a probability map. Finally, the maximum probability can be obtained via the global average pooling layer. 

On the basis of the maximum probability, the authors propose a new loss function (named exponential loss function) to measure the appropriateness of the attention heat map which can be written as: 
\begin{equation}
L = \frac{1}{N} e^{\alpha (P^M_{ij} + \beta \mu)}
\end{equation}
where $P^M_{ij}$ is the maximum probability for image $i$ and attribute $j$. $\alpha$ and $\beta$ are hyper-parameters and $\mu = 1/H^2$ is the mean value of the probability map. $H \times H$ is the size of the attention (and probability) map. For the training of the network, the authors first pre-training the CAM network only by minimizing classification loss; then, they adopt joint loss functions to fine-tune the whole network.

\textbf{Summary:} The Visual attention mechanism has been introduced in pedestrian attribute recognition, but the existing works are still limited. How to design new attention models or directly borrow from other domains still needs to be explored in this area.

\subsection{Sequential Prediction-based Models}

In this section, we will give a review of Sequential Prediction based models for person attribute recognition including CNN-RNN\cite{wang2016cnnRNN}, JRL \cite{wang2017attribute}, GRL \cite{ijcaizhao2018grouping}, JCM \cite{liu2018sequenceJCM} and RCRA \cite{zhao2019RCRA}.

\subsubsection{CNN-RNN (CVPR-2016) \cite{wang2016cnnRNN}} 

\begin{figure*}[htb]
\center
\includegraphics[width=6in]{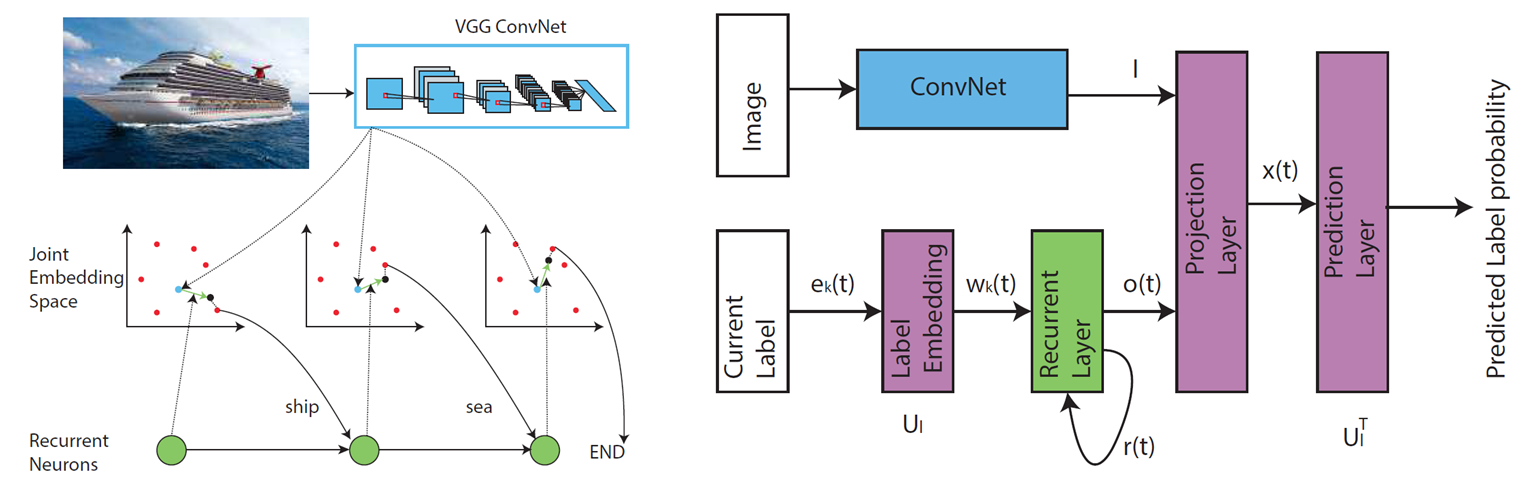}
\caption{The illustration and architecture of the proposed CNN-RNN model for multi-label classification \cite{wang2016cnnRNN}.}
\label{DeepAlgorithms-CNNRNN}
\end{figure*}

Regular multi-label image classification frameworks learn independent classifiers for each category and employ ranking or threshold on the classification results, failing to explicitly exploit the label dependencies in an image. This paper first adopts RNNs to address this problem and combines with CNNs to learn a joint image-label embedding to characterize the semantic label dependency as well as the image-label relevance. As shown in Fig. \ref{DeepAlgorithms-CNNRNN}, the red and blue dots are the label and image embeddings, respectively. The image and recurrent neural output embeddings are summed and denoted with black dots. This mechanism could model the label co-occurrence dependencies in the joint embedding space by sequentially linking the label embeddings. It can compute the probability of a label based on the image embedding $I$ and output of recurrent neurons $x_t$, which can be formulated as: 
\begin{equation}
s(t) = U^T_l x_t 
\end{equation} 
where $x_t = h(U^x_o o(t) + U^x_I I)$, $U^x_o$ and $U^x_I$ are the projection matrices for a recurrent layer of output and image representation, respectively. $U_l$ is the label embedding matrix. $o(t)$ is the output of the recurrent layer at the time step t. 

For the inference of the CNN-RNN model, they attempt to find the sequence of labels that maximize the prior probability: 
\begin{equation}
\begin{small}
\begin{aligned}
l_1, ... , l_k = \arg \max_{l_1, ... , l_k} P(l_1, ... , l_k|I)  \\ 
 = \arg \max_{l_1, ... , l_k} P(l_1|I) \times P(l_2|I, l_1) ... P(l_k|I, l_1, ... , l_{k-1})
\end{aligned}
\end{small}
\end{equation}

They adopt the beam search algorithm \cite{reddy1977speechbeamsearch} for the top-ranked prediction path as their estimation result. The training of the CNN-RNN model can be achieved by cross-entropy loss function and back-propagation through time (BPTT) algorithm. 

The CNN-RNN model is very similar to deep models used in image caption task \cite{vinyals2015showandtell} \cite{Mathews_2018_CVPR}. They all take one image as input and output a series of words under the encoder-decoder framework. The main difference is that the caption model outputs one sentence and the CNN-RNN model generates attributes (but these attributes are also related to each other). Therefore, we can borrow some techniques from the image caption community to help improve the performance of pedestrian attribute recognition.

\subsubsection{JRL (ICCV-2017) \cite{wang2017attribute}} 

\begin{figure}[htb]
\center
\includegraphics[width=3.3in]{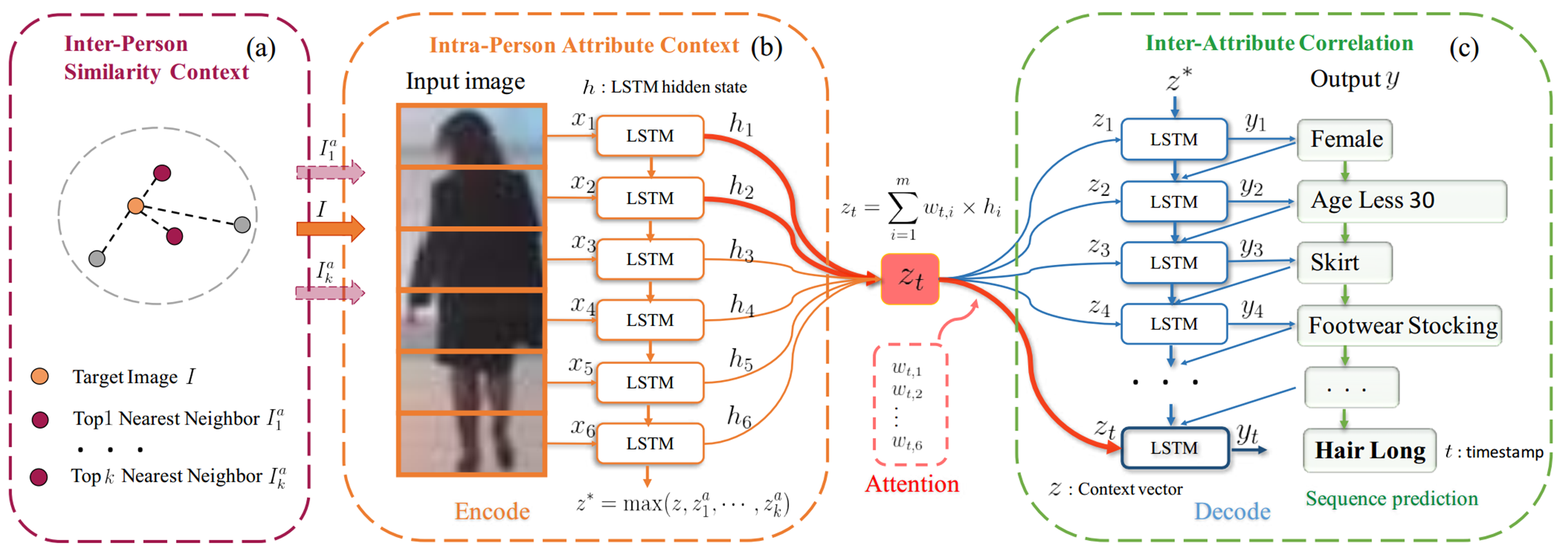}
\caption{The pipeline of Joint Recurrent Learning (JRL) of attribute context and correlation. This figure is adopted from JRL \cite{wang2017attribute}.}
\label{DeepAlgorithms-JRL}
\end{figure}

This paper first analyses the existing learning issues in the pedestrian attribute recognition task, \emph{e.g.} poor image quality, appearance variation, and little annotated data. They propose to explore the interdependency and correlation among attributes and visual context as extra information sources to assist attribute recognition. Hence, the JRL model is proposed for joint recurrent learning of attribute context and correlation, as its name shows. The overall pipeline of JRL can be found in Fig. \ref{DeepAlgorithms-JRL}.

To better mine this extra information for accurate person attribute recognition, the authors adopt the \emph{sequence-to-sequence} model to handle the aforementioned issues. They first divide the given person image $I$ into $m$ horizontal strip regions and form a region sequence $S = (s_1, s_2, ..., s_m)$ in top-bottom order. The obtained region sequences $S$ can be seen as the input sentence in natural language processing, and can be encoded with the LSTM network in a sequential manner. The hidden state $h^{en}$ of the encoder LSTM can be updated based on the regular LSTM update procedure, as shown in Eq. \ref{lstmUpdatefunction}. The final hidden state $h^{en}_{m}$ can be seen as the summary representation $z = h^{en}_{m}$ of the whole person image (named as \emph{context vector}). This feature extraction procedure could model the intra-person attribute context within each person image $I$.

To mine more auxiliary information to handle the appearance ambiguity and poor image quality in a target image. The authors resort to the visually similar exemplar training images and introduce these samples to model the inter-person similarity context constraint. They first search top-$k$ samples that are similar to the target image with CNN features based on the L2 distance metric and compute its own context vector $z^a_i$. Then, all the context vector representations are ensembled as the inter-person context $z^*$ with the max-pooling operation.

In the decoding phase, the decoder LSTM takes both \emph{intra-person attribute context (z)} and \emph{inter-person similarity context $(z^*)$} as input and output variable-length attributes over time steps. The attribute prediction in this paper can also be seen as a generation scheme. To better focus on local regions of person image for specific attributes and obtain more accurate representation, they also introduce the attention mechanism to attend to the intra-person attribute context. For the final attribute estimation order, they adopt the ensemble idea to incorporate the complementary benefits of different orders and thus capture more high-order correlation between attributes in context.

\subsubsection{GRL (IJCAI-2018) \cite{ijcaizhao2018grouping}} 
\begin{figure*}[htb]
\center
\includegraphics[width=6.5in]{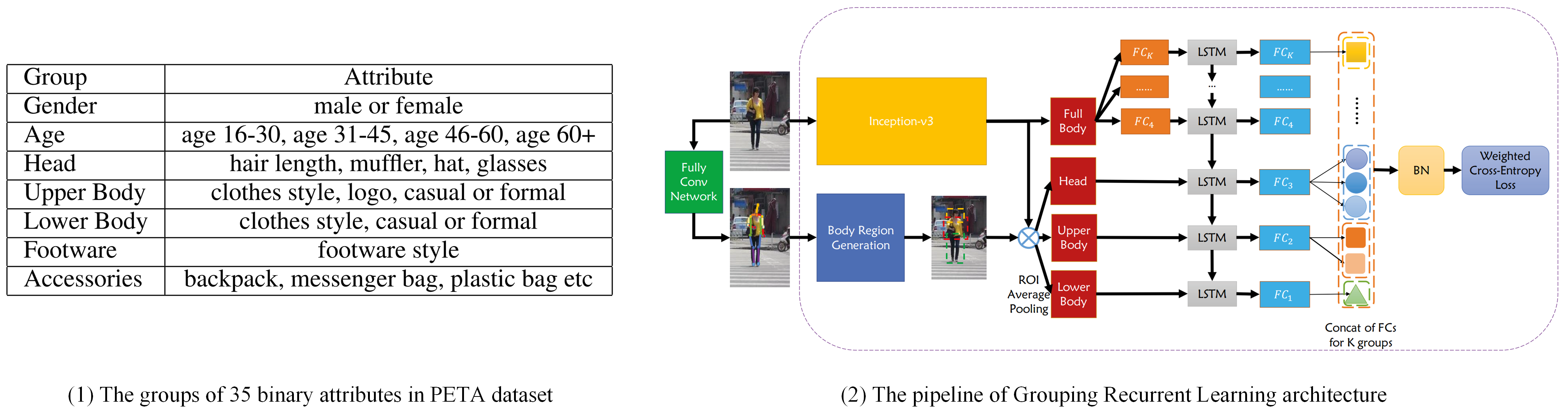}
\caption{The pipeline of Group Recurrent Learning (GRL). This figure is adapted from GRL \cite{ijcaizhao2018grouping}.}
\label{DeepAlgorithms-GRL}
\end{figure*}

GRL is developed based on JRL which also adopts the RNN model to predict the human attributes in a sequential manner. Different from JRL, GRL is formulated to recognize human attributes by group step by step to pay attention to both intra-group and inter-group relationships. As shown in Fig. \ref{DeepAlgorithms-GRL} (1), the author divides the whole attribute list into many groups because the attributes in the intra-group are mutually exclusive and have relations between inter-group. For example, \emph{BoldHair} and \emph{BlackHair} cannot occur on the same person image, but they are both related to the head-shoulder region of a person and can be in the same group to be recognized together. It is an end-to-end single model algorithm with no need for preprocessing and it also exploits more latent intra-group and inter-group dependency among grouped pedestrian attributes. The overall algorithm can be found in Fig. \ref{DeepAlgorithms-GRL} (2).

As shown in Fig. \ref{DeepAlgorithms-GRL} (2), given the human image, they first detect the key points and locate the head, upper body, and lower body regions using the body region generation module. They extract the whole image features with the Inception-v3 network and obtain the local part features using ROI average pooling operation. It is worth noting that all attributes in the same group share the same fully connected feature. Given the global and local features, they adopt LSTM to model the spatial and semantic correlations in attribute groups. The output of each LSTM unit is then fed into a fully connected layer and a prediction vector can be obtained. This vector has the same dimension as the number of attributes in the relevant group.  They also use a batch normalized layer to balance the positive and negative outputs of this network.

\subsubsection{JCM (arXiv-2018) \cite{liu2018sequenceJCM}} 

\begin{figure}[htb]
\center
\includegraphics[width=3in]{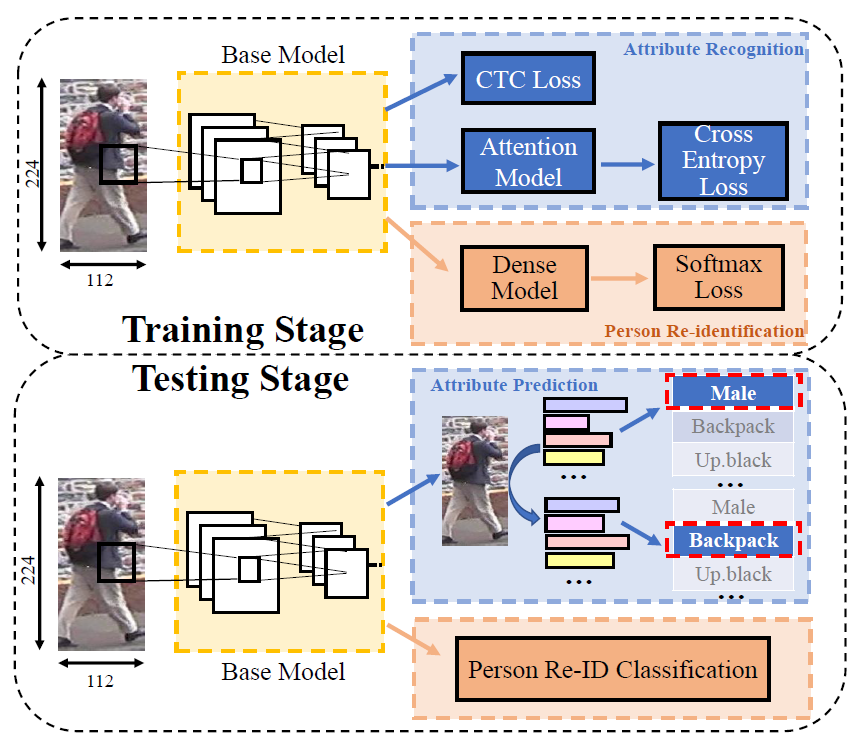}
\caption{The overview of JCM which is proposed in \cite{liu2018sequenceJCM}.}
\label{DeepAlgorithms-JCM}
\end{figure}	

Existing sequential prediction-based person attribute recognition algorithms, such as JRL \cite{wang2017attribute}, and GRL \cite{ijcaizhao2018grouping}, may be easily influenced by different manual division and attribute orders due to the weak alignment ability of RNN. This paper proposes a joint CTC-Attention model (JCM) to conduct attribute recognition, which could predict multiple attribute values with arbitrary length at a time avoiding the influence of attribute order in the mapping table. 

As shown in Fig. \ref{DeepAlgorithms-JCM}, the JCM is actually a multi-task network that contains two tasks, \emph{i.e.} the attribute recognition and person re-identification. They use ResNet-50 as the basic model to extract features for both tasks. For attribute recognition, they adopt the Transformer \cite{vaswani2017attention} as their attention model for the alignment of long attribute sequences. The connectionist temporal classification (CTC) loss \cite{graves2006connectionist} and cross-entropy loss functions are used for the training of the network. For the person re-ID, they directly use two fully connected layers (\emph{i.e.} the dense model) to obtain feature vectors and use the softmax loss function to optimize this branch. 

In the test phase, the JCM could simultaneously predict the person's identity and a set of attributes. They also use beam search for the decoding of attribute sequences. Meanwhile, they extract the features from the CNN in the base model to classify pedestrians for person re-ID tasks.

\subsubsection{RCRA (AAAI-2019) \cite{zhao2019RCRA}} 
\begin{figure}[htb]
\center
\includegraphics[width=3.5in]{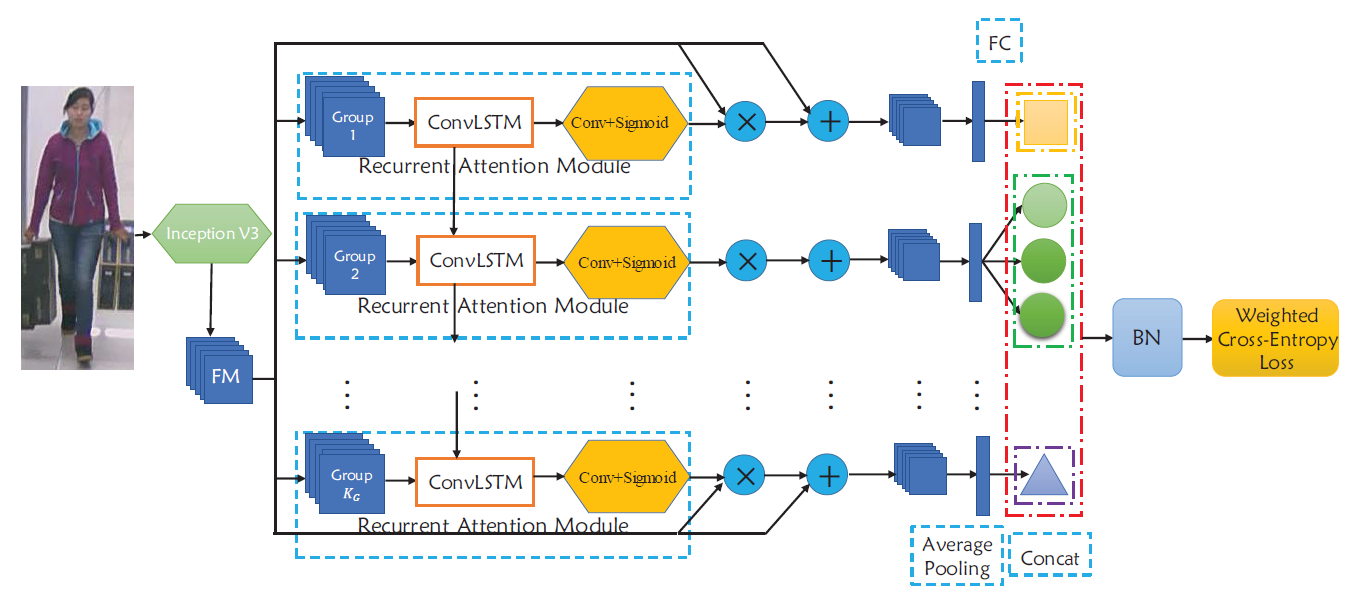}
\caption{The overview of RCRA which is proposed in \cite{zhao2019RCRA}.}
\label{DeepAlgorithms-RCRA}
\end{figure}

This paper proposes two models, \emph{i.e.}, Recurrent Convolutional (RC) and Recurrent Attention (RA) for pedestrian attribute recognition, as shown in Figure \ref{DeepAlgorithms-RCRA}. The RC model is used to explore the correlations between different attribute groups with the Convolutional-LSTM model \cite{xingjian2015convolutionallstm} and the RA model takes advantage of the intra-group spatial locality and inter-group attention correlation to improve the final performance. 

Specifically speaking, they first divide all the attributes into multiple attribute groups, similar to GRL \cite{ijcaizhao2018grouping}. For each pedestrian image, they use CNN to extract its feature map and feed it to the ConvLSTM layer group by group. Then, new feature maps for each time step can be obtained by adding a convolutional network after ConvLSTM. Finally, the features are used for attribute classification on the current attribute group. 

Based on the aforementioned RC model, they also introduce a visual attention module to highlight the region of interest on the feature map. Given the image feature map $F$ and the heat map $H_t$ of attention at each time step $t$, the attended feature map $F_t$ for the current attribute group can be obtained via: 
\begin{equation}
\label{rarcAtteniton} 
F_t = sigmoid(H_t) \otimes F + F
\end{equation}
where $\otimes$ denotes the spatial point-wise multiplication. The attended feature maps are used for final classification. The training of this network is also based on the weighted cross-entropy loss function proposed in WPAL-network \cite{zhou2017weakly}.

\textbf{Summary:} As we can see from this subsection, these algorithms all adopt the sequential estimation procedure. Because the attributes are correlated to each other, they also have various difficulties. Therefore, it is an interesting and intuitive idea to adopt the RNN model to estimate the attributes one by one. Among these algorithms, they integrate different neural networks, attribute groups, and multi-task learning into this framework. Compared with CNN-based methods, these algorithms are more elegant and effective. The disadvantage of these algorithms is the time efficiency due to the successive attribute estimation. In future works, more efficient algorithms for sequential attribute estimation are needed.

\subsection{Loss Function based Models}

In this section, we will review some algorithms with improved loss function, including WPAL \cite{zhou2017weakly}, AWMT \cite{mmhe2017adaptively}.

\subsubsection{WPAL-network (BMVC-2017) \cite{zhou2017weakly}}  

\begin{figure}[htb]
\center
\includegraphics[width=3.3in]{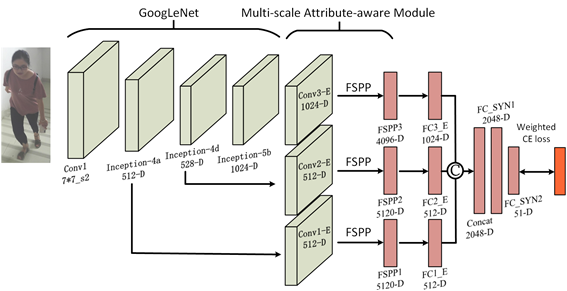}
\caption{The pipeline of WPAL-network. This figure is rewritten based on WPAL-network \cite{zhou2017weakly}.}
\label{DeepAlgorithms-WPAL}
\end{figure}

The WPAL is proposed to simultaneously recognize and locate the person attributes in a weakly-supervised manner (\emph{i.e.} only person attribute labels, no specific bounding box annotation). As shown in Fig. \ref{DeepAlgorithms-WPAL}, GoogLeNet is adopted as their basic network for feature extraction. They fuse features from different layers (\emph{i.e.} the features from $Conv3-E, Conv2-E$, and $Conv1-E$ layers) and feed into the Flexible Spatial Pyramid Pooling layer (FSPP). Compared with regular global max-pooling, the advantage of FSPP can be listed as following two aspects: 1). it can add a spatial constraint to some attributes like hats; 2). The structure lies in the middle stage of the network but not the top, making a correlation between the detector and target class not bound at first but free to be learned during training. The outputs of each FSPP are fed into fully connected layers and output a vector whose dimension is the same as the number of pedestrian attributes.

In the training procedure, the network could simultaneously learn to fit the following two targets: the first one is to learn the correlation between attributes and randomly initialized mid-level detectors, and the second one is to adapt the target mid-level features of detectors to fit the correlated attributes. The learned correlation, the detection results of mid-level features are later used to locate the person's attributes.

In addition, the authors also introduce a novel weighted cross entropy loss function to handle the extremely imbalanced distribution of positive and negative samples of most attribute categories. The mathematical formulation can be written as follows:
\begin{equation}\label{weightedCElossfunction}
  Loss_{wce}  = \sum_{i=1}^{L} \frac{1}{2w_i} * p_i * log(\hat{p}_i) + \frac{1}{2(1-w_i)}(1-p_i) * log(1-\hat{p}_i)
\end{equation}
where $L$ denotes the number of attributes, $p$ is the ground truth attribute vector, $\hat{p}$ is the estimated attribute vector, and $w$ is a weight vector indicating the proportion of positive labels overall attribute categories in the training dataset.

\subsubsection{AWMT (MM-2017) \cite{mmhe2017adaptively}}  
\begin{figure}[htb]
\center
\includegraphics[width=3.5in]{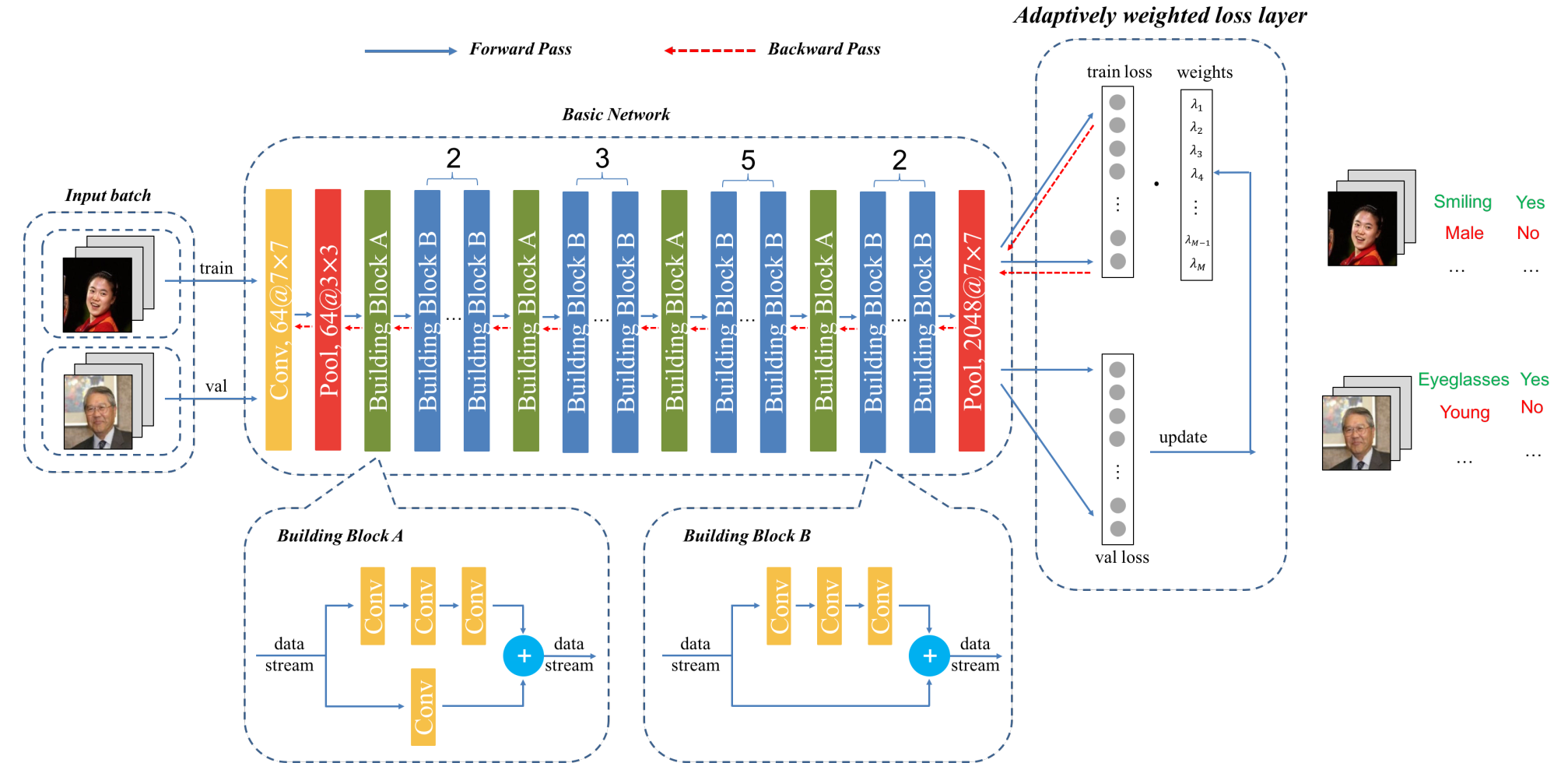}
\caption{The pipeline of AWMT-network. This figure is adopted from AWMT \cite{mmhe2017adaptively}.}
\label{DeepAlgorithms-AWMT}
\end{figure}

As is known to all, the learning difficulty of various attributes is different. However, most existing algorithms ignore this situation and share relevant information in their multi-task learning framework. This will lead to \emph{negative transfer}, in other words, the inadequate brute-force transfer may hurt the learner’s performance when two tasks are dissimilar. AWMT proposes to investigate a sharing mechanism that is possible of \emph{dynamically} and \emph{adaptively} coordinating the relationships of learning different person attribute tasks. Specifically, they propose an adaptively weighted multi-task deep framework to jointly learn multiple person attributes, and a validation loss trend algorithm to automatically update the weights of the weighted loss layer. The pipeline of their network can be found in Fig. \ref{DeepAlgorithms-AWMT}.

As shown in Fig. \ref{DeepAlgorithms-AWMT}, they adopt ResNet-50 as a basic network and take both train and val images as input. The basic network will output its predicted attribute vectors for both train and val images. Hence, the train loss and val loss can be obtained simultaneously. The value loss is used to update the weight vectors $\lambda_j (j = 1, ..., M)$ which is then utilized to weight different attributes learning. The adaptive weighted loss function can be formulated as follows:
\begin{equation}\label{awmtweightedLossfunction}
\Theta = \arg \min_{\Theta} \sum_{j=1}^{M} \sum_{i=1}^{N} <\lambda_j, \mathcal{L}(\psi_j(\textbf{I}_i; \Theta) - \textbf{L}_{ij})>
\end{equation}
where $\Theta$ denotes the parameters of the neural network, $\lambda_j$ is the scale value to weigh the importance of the task of learning $j$-th attributes. $\textbf{I}_i$ denote the $i$-th image in a mini-batch, $\textbf{L}_{ij}$ is the ground truth label of attribute $j$ of image $i$. $\psi_j(\textbf{I}_i; \Theta)$ is the predicted attributes of input image $\textbf{I}_i$ under neural network parameter $\Theta$. $<\cdot>$ is the inner product operation.

The key problem is how to adaptively tune the weight vector $\lambda_j$ in Eq. \ref{awmtweightedLossfunction}. They propose the validated loss trend algorithm to realize this target. The intuition behind their algorithm is that in learning multiple tasks simultaneously, the "important" tasks should be given high weight (\emph{i.e.} $\lambda_j$ ) to increase the scale of loss of the corresponding tasks. But the question is how we can know which task is more "important", in other words, how do we measure the importance of one task?

In this paper, the authors propose to use the generalization ability as an objective measurement. Specifically, they think the trained model of one task with lower generation ability should be set higher weight than those models of the other tasks. The weight vector $\lambda_j$ is updated per $k$ iterations and used to compute the loss of training data and update the network parameter $\Theta$ in the backward pass. Their experiments on several attribute datasets validated the effectiveness of this adaptive weighting mechanism.

\textbf{Summary:} There are few works that focus on designing new loss functions for pedestrian attribute recognition. WPAL-network \cite{zhou2017weakly} considers the unbalanced distribution of data and proposes a weighted cross-entropy loss function according to the proportion of positive labels overall attribute categories in the training dataset. This method seems a little tricky but has been widely used in many PAR algorithms. AWMT \cite{mmhe2017adaptively} proposes an adaptive weighting mechanism for each attribute learning to make the network focus more on handling the ``hard'' tasks. These works fully demonstrate the necessity to design novel loss functions to better train the PAR network.

\subsection{Curriculum Learning-based Algorithms} 
In this subsection, we will introduce the curriculum learning-based algorithm which considers learning the human attribute in a ``easy'' to ``hard'' way, such as MTCT \cite{dong2017multiWACV},  CILICIA \cite{sarafianos2017curriculum}.

\subsubsection{MTCT (WACV-2017) \cite{dong2017multiWACV}} 

\begin{figure*}[htb]
\center
\includegraphics[width=6.5in]{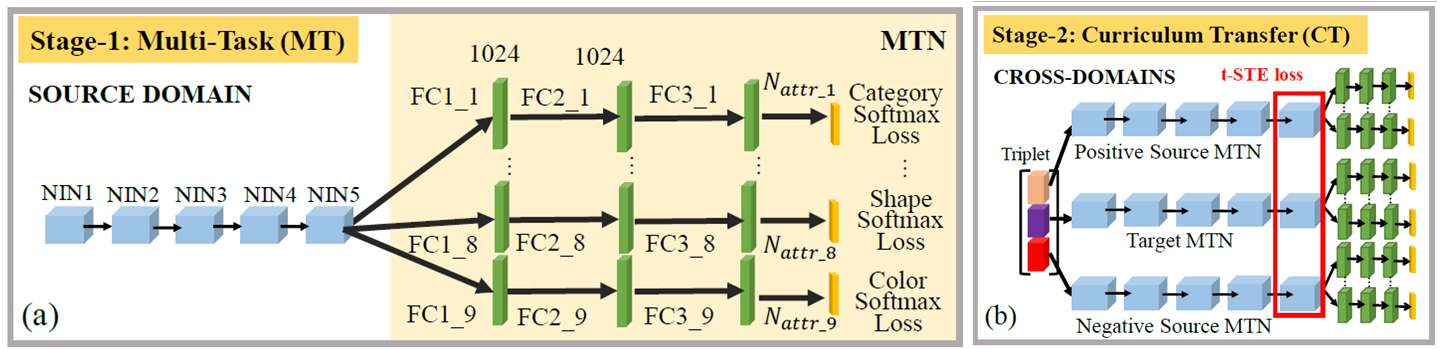}
\caption{The MTCT network design \cite{dong2017multiWACV}.}
\label{DeepAlgorithms-MTCT}
\end{figure*}

This paper proposes a multi-task curriculum transfer network to handle the issue of the lack of manually labeled training data. As shown in Fig. \ref{DeepAlgorithms-MTCT}, their algorithm mainly contains multi-task network and curriculum transfer learning. 

For the multi-task network, they adopt five stacked Network-In-Network (NIN) convolutional units \cite{lin2013networkNIN} and $N$ parallel branches, with each branch representing three layers of fully connected sub-network for modeling one of the $N$ attributes respectively. The Softmax loss function is adopted for the model training. 

Inspired by cognitive studies that suggest a better learning strategy adopted by humans and animals is to start with learning easier tasks before gradually increasing the difficulties of the tasks, rather than blindly learning randomly organized tasks.  Therefore, they adopt a curriculum transfer learning strategy for clothing attribute modeling. Specifically, it consists of two main stages. In the first stage, they use the clean (\emph{i.e.} easier) source images and their attribute labels to train the model. In the second stage, they embed cross-domain image pair information and simultaneously appended harder target images into the model training process to capture harder cross-domain knowledge. They adopt the t-STE (t-distribution stochastic triplet embedding) loss function to train the network which can be described as: 
\begin{equation}
\begin{small}
\begin{aligned}
L_{t-STE} = \sum_{I_t, I_{ps}, I_{ns} \in T} \\ 
log  \frac{(1+\frac{||f_t(I_t) - f_s(I_{ps})||^2}{\alpha})^{\beta}}{ (1+\frac{||f_t(I_t) - f_s(I_{ps})||^2}{\alpha})^{\beta}   +  (1+\frac{||f_t(I_t) - f_s(I_{ns})||^2}{\alpha})^{\beta}} 
\end{aligned}
\end{small}
\end{equation}
where $\beta = - 0.5 * (1+\alpha) $ and $\alpha$ is the freedom degree of the Student kernel. $f_t(\cdot)$ and $f_s(\cdot)$ are the feature extraction function for target and source multi-task networks, respectively.

\subsubsection{CILICIA (ICCV-2017) \cite{sarafianos2017curriculum}} 

\begin{figure}[htb]
\center
\includegraphics[width=3.5in]{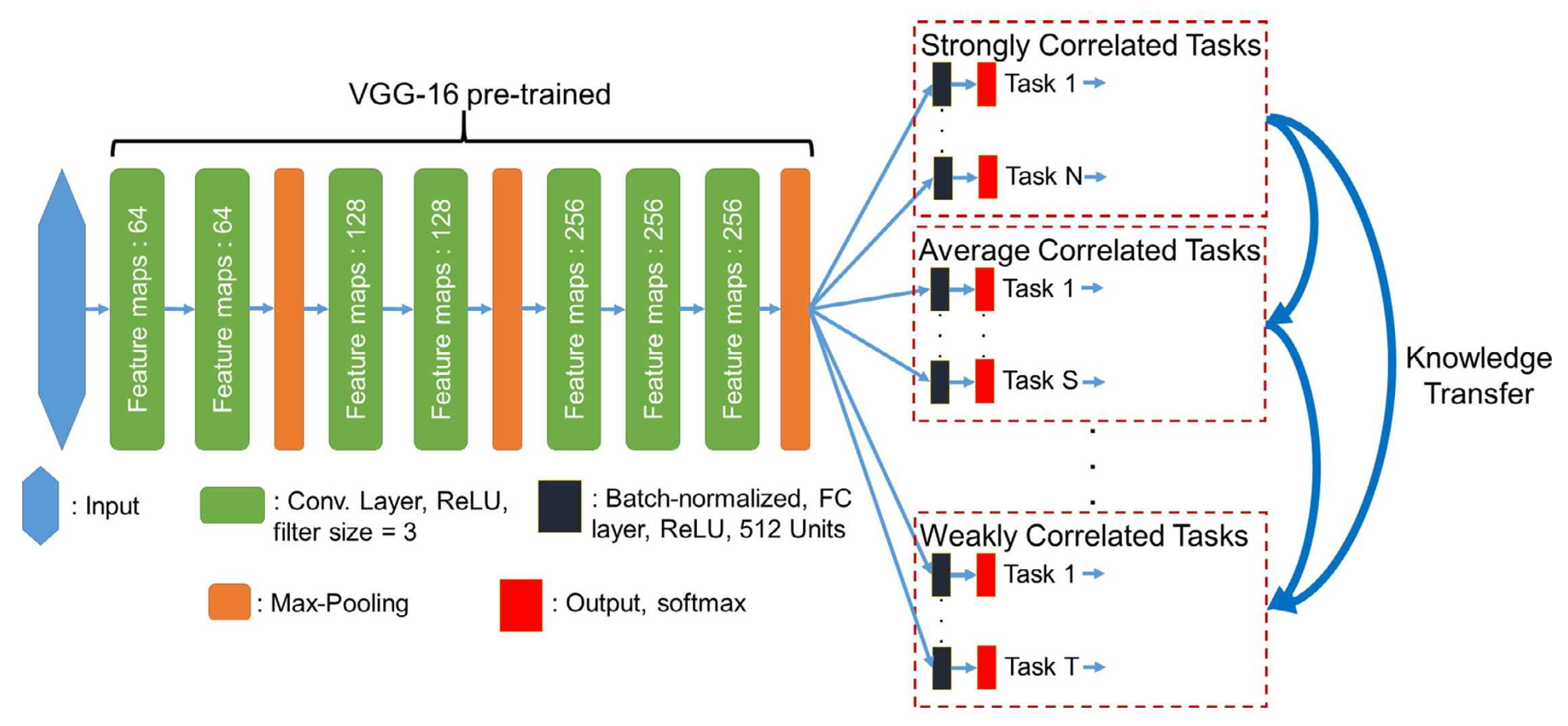}
\caption{The CILICIA network designed in \cite{sarafianos2018curriculumPR} .}
\label{DeepAlgorithms-CILICIA}
\end{figure}

Similar to MTCT \cite{dong2017multiWACV}, CILICIA \cite{sarafianos2017curriculum} also introduce the idea of curriculum learning into person attribute recognition task to learning the attributes from easy to hard. The pipeline of CILICIA can be found in Fig. \ref{DeepAlgorithms-CILICIA}.  They explore the correlations between different attribute learning tasks and divide such correlations into strongly and weakly correlated tasks. Specifically, under the framework of multi-task learning, they use the respective Pearson correlation coefficients to measure the strongly correlated tasks which can be formulated as: 
\begin{equation}
p_i = \sum_{j=1, j \neq i}^{T} \frac{cov(y_{t_i}, y_{t_j})}{\sigma(y_{t_i} \sigma(y_{t_j}))}, i = 1, ... , T 
\end{equation}
where $\sigma(y_{t_i})$ is the standard deviation of the labels $y$ of the task $t_i$. The tasks of $p_i$ with top $50\%$ are strongly correlated with rest and can be divided into strongly correlated groups. The rest tasks belong to a weakly correlated group and will be learned under the guidance of knowledge learned from a strongly correlated group.

For the multi-task network, they adopt the categorical cross-entropy function \cite{zhu2017multiImageVC} between predictions and targets, which can be defined as follows (for a single attribute t):  
\begin{equation}
\label{catrgoricalCEF}
L_t = \frac{1}{N} \sum_{i=1}^{N} \sum_{j=1}^{M} (\frac{1/M_j}{\sum_{n=1}^{M} 1/M_n}) \cdot \mathbbm{1}[y_i = j] \cdot log(p_{i, j})
\end{equation}
where $\mathbbm{1}[y_i = j]$ is one if the target of sample $i$ belong to class $j$, and zero otherwise. $M_j$ is the number of samples belonging to class $j$, and M and N are the number of classes and samples, respectively.

To weight different attribute learning tasks, one intuitive idea is to learn another branch network for weights learning. However, the authors did not see significant improvement with this method. Therefore, they adopt the \emph{supervision transfer} learning technique \cite{zhang2016real} to help attribute learning in weakly correlated group: 
\begin{equation}
L_w = \lambda \cdot L_s + (1 - \lambda) \cdot L^f_w, 
\end{equation}
where $L_w^f$ is the total loss obtained during the forward pass using Eq. \ref{catrgoricalCEF} only over the weakly correlated tasks.

They also propose CILICIA-v2 \cite{sarafianos2018curriculumPR} by proposing an effective method to obtain the groups of tasks using hierarchical agglomerative clustering. It can be of any number and not just only two groups (\emph{i.e.} strong/weakly correlated). More specifically, they employ the computed Pearson correlation coefficient matrix to perform hierarchical agglomerative clustering using the Ward variance minimization algorithm. Ward’s method is biased towards generating clusters of the same size and analyses all possible pairs of joined clusters, identifying which joint produces the smallest within-cluster sum of squared (WCSS) errors. Therefore, we can obtain attribute groups via the WCSS threshold operation. For each group, they compute the learning sequence of clusters by sorting the obtained respective Pearson correlation coefficients only within the clusters. Once the total dependencies for all the clusters are formed, the curriculum learning process can be started in descending order.

\textbf{Summary:} Inspired by the recent progress of cognitive science, the researchers also consider using such ``easy" to ``hard" learning mechanisms for PAR. They introduce existing curriculum learning algorithms into their learning procedure to model the relations between each attribute. Some other algorithms such as self-paced learning \cite{kumar2010selfPL} are also used to model the multi-label classification problem \cite{li2018selfPLforML} or other computer vision tasks \cite{zhang2017cosaliencyDetection}. It is also worthy to introduce more advanced works of cognitive science to guide the learning of PAR.

\subsection{Graphic Model based Algorithms} 
Graphic models are commonly used to model structure learning in many applications. Similarly, there are also some works to integrate these models into the pedestrian attribute recognition task, for example, DCSA \cite{DCSA2012}, A-AOG \cite{park2018attributeAOG}, VSGR \cite{vsgraaai2019}.

\subsubsection{DCSA (ECCV-2012) \cite{DCSA2012}} 

\begin{figure}[htb]
\center
\includegraphics[width=3.5in]{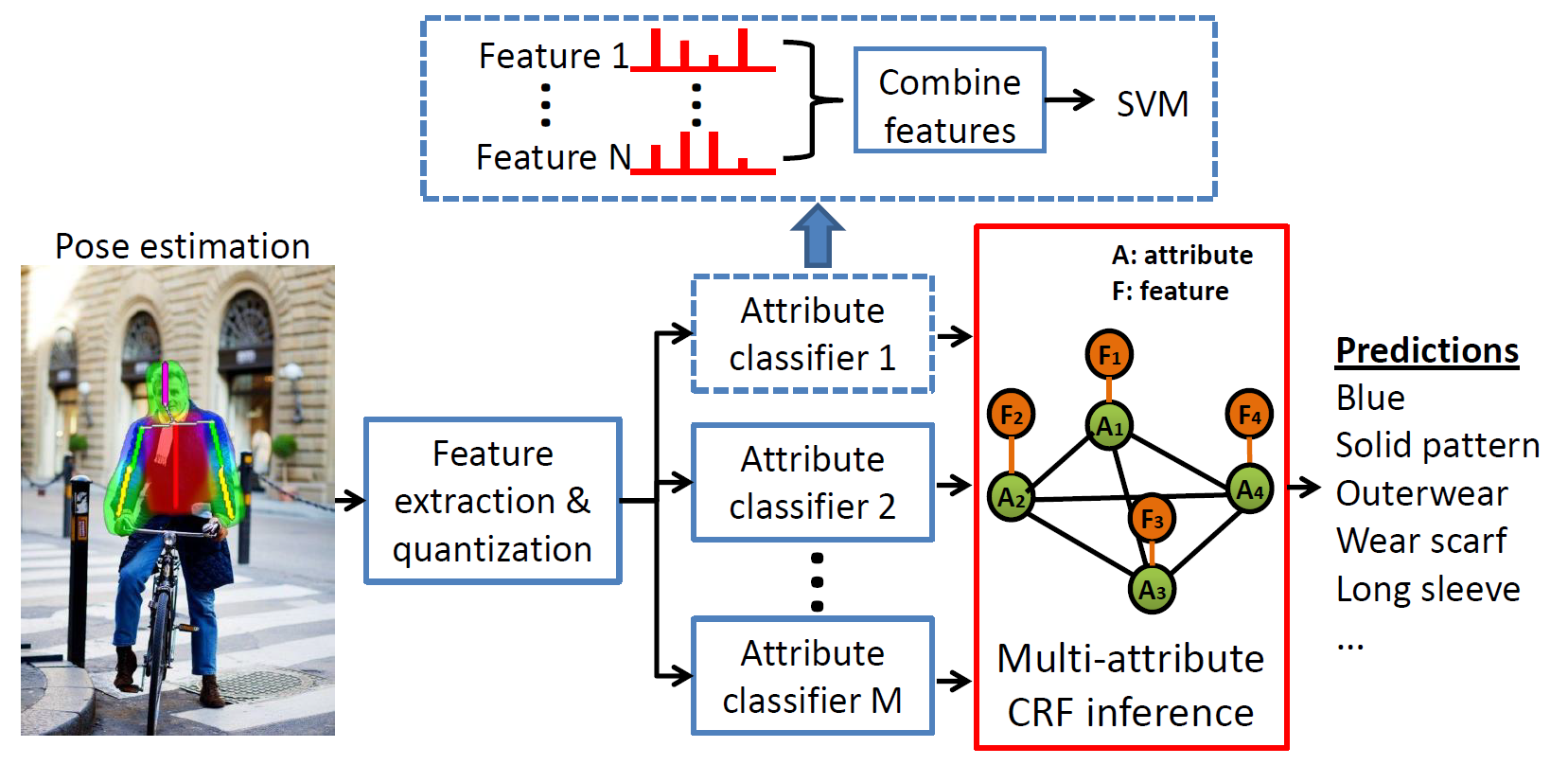}
\caption{The pipelines of DCSA \cite{DCSA2012}.}
\label{DeepAlgorithms-DCSA}
\end{figure}	

In this paper, the authors propose to model the correlations between human attributes using a conditional random field (CRF). As shown in Fig. \ref{DeepAlgorithms-DCSA}, they first estimate the pose information using off-the-shelf algorithms \cite{eichner2010articulated} and locate the local parts of the upper body only (the lower body is ignored because of the occlusion issues). Then, four types of base features are extracted from these regions, including SIFT \cite{lowe2004distinctive}, texture descriptor \cite{varma2005statistical}, color in LAB space, and skin probabilities. These features are fused to train multiple attribute classifiers via SVM. The key idea of this paper is to apply the fully connected CRF to explore the mutual dependencies between attributes. They treat each attribute function as a node of CRF and the edge connecting every two attribute nodes reflects the joint probability of these two attributes.  The belief propagation \cite{tappen2003comparison} is adopted to optimize the attribute label cost.

\subsubsection{A-AOG (TPAMI-2018) \cite{park2018attributeAOG}} 

\begin{figure}[htb]
\center
\includegraphics[width=3.5in]{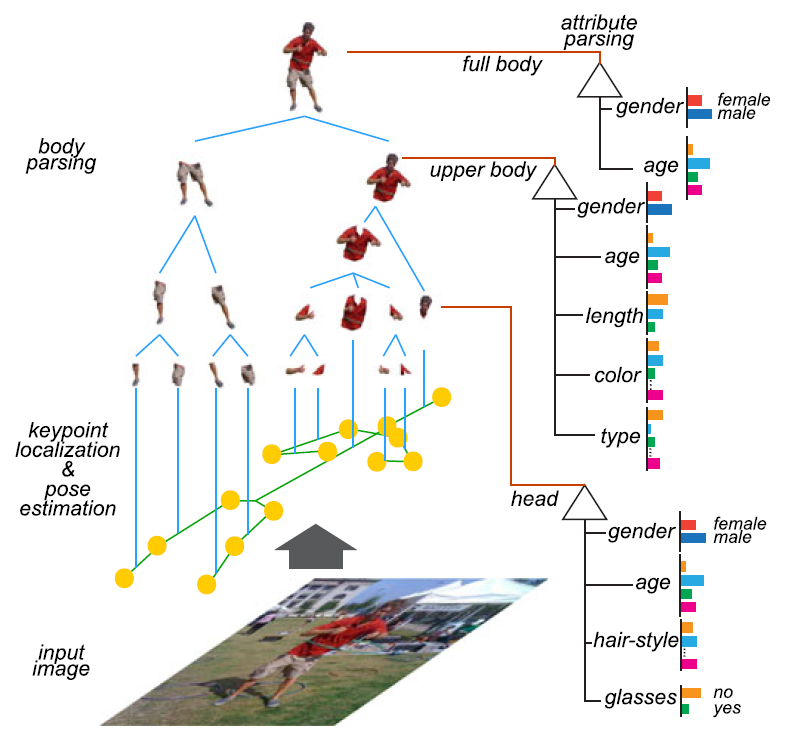}
\caption{An attributed parse graph for a human image \cite{park2018attributeAOG}.}
\label{DeepAlgorithms-AOG}
\end{figure}	

The A-AOG model which is short for attribute And-Or grammar, is proposed to explicitly represent the decomposition and articulation of body parts, and account for the correlations between poses and attributes. This algorithm is developed based on And-Or graph \cite{zhu2007stochasticAOG} and the and-nodes denote decomposition or dependency; the or-nodes represent alternative choices of decomposition or types of parts. Specifically speaking, it mainly integrates the three types of grammar: \emph{phrase structure grammar, dependency grammar}, and an \emph{attribute grammar}. 

Formally, the A-AOG is defined as a five-tuple:
\begin{equation}
A-AOG = < S, V, E, X, \mathcal{P}  >
\end{equation}
where $V$ is the vertex set and it mainly contains a set of and-nodes, or-nodes, and terminal nodes: $V = V_{and} \cup V_{or} \cup V_{T}$; 
$E$ is the edge set and it consists of two subsets $E = E_{psg} \cup E_{dg}$: set of edges with phrase structure grammar $E_{psg}$ and dependency grammar $E_{dg}$. $X = \{x_1, x_2, ... , x_N\}$ is the attribute set associated with nodes in $V$.  $\mathcal{P}$ is the probability model on graphical representation. 

\begin{figure*}[htb]
\center
\includegraphics[width=7in]{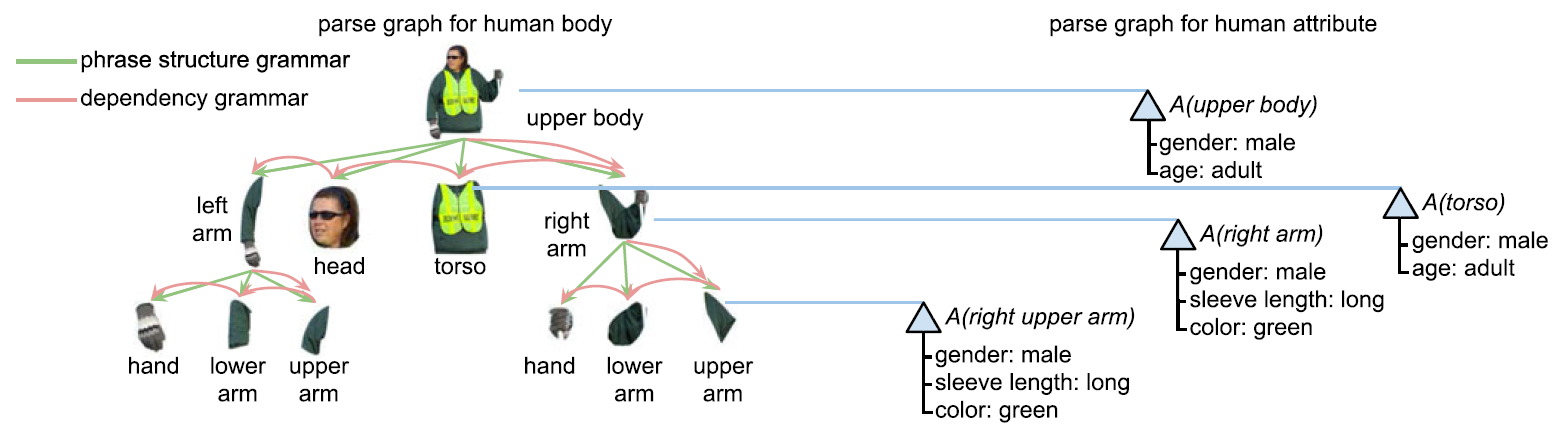}
\caption{A parse graph example derived from the A-AOG \cite{park2018attributeAOG}, which contains parse graphs for human body detection and pose and human attributes.}
\label{DeepAlgorithms-ParseGraph}
\end{figure*}	

According to the aforementioned definitions, the parse graph can be formulated as:
\begin{equation}
pg = (V(pg), E(pg), X(pg)) 
\end{equation}
An example of the parse graph derived from the A-AOG can be found in Fig. \ref{DeepAlgorithms-ParseGraph}. Given an image $I$, the target is to find the most probable parse graph $pg$ from their grammar model. They adopt the Bayesian framework which computes the joint posterior as the product of a likelihood and prior probability, to formulate the probability model over the parse graph as follows:
\begin{equation}
\label{BayesianFrameworkFunction}
\begin{small}
\begin{aligned}
P(pg|I; \lambda) \varpropto P(I|pg; \lambda) P(pg; \lambda) 	\\  
		= \frac{1}{Z} \exp \{ -\mathcal{E}(I|pg; \lambda) - \mathcal{E}(pg; \lambda) \} 
\end{aligned}
\end{small}
\end{equation}
where $\lambda$ is the model parameters. The energy functions $\mathcal{E}$ can be decomposed into a set of potential functions. Both terms in Eq. \ref{BayesianFrameworkFunction} can be decomposed into part and attribute relations, therefore, Eq. \ref{BayesianFrameworkFunction} can be rewritten as: 
\begin{equation}
\label{BayesianFrameworkFunctionV2}
\begin{small}
\begin{aligned}
P(pg|I; \lambda) = \frac{1}{Z} \exp \{ -\mathcal{E}^V_{app} (I|pg; \lambda) - \mathcal{E}^X_{app} (I|pg; \lambda) \\ 
														 -\mathcal{E}^V_{rel}(pg; \lambda) - \mathcal{E}^X_{rel} (pg; \lambda) \} 
\end{aligned}
\end{small}
\end{equation}
where $\mathcal{E}^V_{app} (I|pg; \lambda)$, $\mathcal{E}^X_{app} (I|pg; \lambda)$, $\mathcal{E}^V_{rel}(pg; \lambda)$ and $\mathcal{E}^X_{rel} (pg; \lambda)$ are appearance and relations terms for part and attribute respectively.

Then, the energy terms can be expressed as following scoring functions:
\begin{equation}
\label{BayesianFrameworkFunctionV3}
\begin{small}
\begin{aligned}
S(pg|I) =  - \mathcal{E}^V_{app} (I|pg) - \mathcal{E}^X_{app} (I|pg) -\mathcal{E}^V_{rel}(pg) - \mathcal{E}^X_{rel} (pg) \\ 
= S^V_{app} (I, pg) + S^X_{app} (I, pg) + S^V_{rel} (pg) + S^X_{rel} (pg). 
\end{aligned}
\end{small}
\end{equation}
Therefore, the most probable parse graph $pg^*$ can be found by maximizing the score function \ref{BayesianFrameworkFunctionV3}: 
\begin{equation}
\label{BayesianFrameworkFunctionV4}
\begin{small}
\begin{aligned}
pg^* = \arg \max_{pg} P(I|pg) P(pg)  \\ 
	  = \arg \max_{pg} [S^V_{app} (pg, I) + S^X_{app} (pg, I) + S^V_{rel} (pg) + S^X_{rel} (pg)] \\ 
	  \thickapprox  \arg \max_{pg} [S_{app} (pg, I)  + S_{rel} (pg)] 
\end{aligned}
\end{small}
\end{equation}

They use deep CNN to generate the proposals for each part and adopt a greedy algorithm based on the beam search to optimize an aforementioned objective function. For the detailed learning and inference procedure, please check the original paper \cite{park2015attributedAOG} \cite{park2018attributeAOG}.

\subsubsection{VSGR (AAAI-2019) \cite{vsgraaai2019}} 

\begin{figure}[htb]
\center
\includegraphics[width=3.5in]{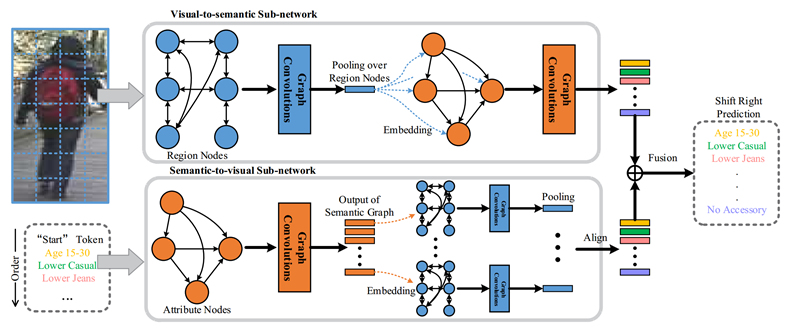}
\caption{The overview of visual-semantic graph reasoning framework (VSGR) which is proposed in \cite{vsgraaai2019}.}
\label{DeepAlgorithms-VSGR}
\end{figure}	

In this paper, the authors propose to estimate the pedestrian attributes via visual-semantic graph reasoning (VSGR). They argue that the accuracy of person attribute recognition is heavily influenced by: 1). only local parts are related with some attributes; 2). challenging factors, such as pose variation, viewpoint, and occlusion; 3). the complex relations between attributes and different part regions. Therefore, they propose to jointly model spatial and semantic relations of region-region, attribute-attribute, and region-attribute with a graph-based reasoning framework. The overall pipeline of their algorithm can be found in Fig. \ref{DeepAlgorithms-VSGR}.  

As shown in Fig. \ref{DeepAlgorithms-VSGR}, this algorithm mainly contains two sub-networks, \emph{i.e.} the visual-to-semantic sub-network and semantic-to-visual sub-network. For the first module, it first divides the human image into a fixed number of local parts $\textbf{X} = (x_1, x_2, ..., x_M)^T$. They construct a graph whose node is the local part and edge is the similarity of different parts. Different from regular relation modeling, they adopt both the similarity relations between parts and topological structures to connect one part with its neighbor regions. The similarity adjacency matrix can be formulated as: 
\begin{equation}
\label{similarityVSGR}
\textbf{A}_{s_a} (i, j) = \frac{\exp(\textbf{F}_s(x_i, x_j))}{\sum_{j=1}^{M} \exp(\textbf{F}_s (x_i, x_j))} 
\end{equation}
where $\textbf{F}_s(x_i, x_j)$ denotes the pairwise similarity between each two-part regions which can also be modeled by a neural network. 

The topological relations between local parts can be obtained via: 
\begin{equation}
\textbf{A}_{s_l}(i, j) = \frac{\exp(-d_{ij} / \Delta)}{\sum_{j=1}^{M} \exp(-d_{ij} / \Delta)}
\end{equation}
where $d_{ij}$ is the pixel distance between two parts and $\Delta$ is the scaling factor.

The two sub-graphs are combined to compute the output of the spatial graph via the following equation: 
\begin{equation}
\textbf{G}_s = \textbf{A}_{s_a} \textbf{X} \textbf{W}_{s_a} + \textbf{A}_{s_l} \textbf{X} \textbf{W}_{s_l} 
\end{equation}
where $\textbf{W}_{s_a}$ and $\textbf{W}_{s_l}$ are weight matrices for two sub-graphs. 

Therefore, the spatial context representation $g_s$ can be obtained via average pooling operation after convolution. After encoding the region-to-region relation, they also adopt a similar operation to model the relations between semantic attributes based on spatial context. The node of the new graph is the attributes, and they transform them into the embedding matrix $\textbf{R} = (r_0, r_1, ..., r_K)$, where $r_0$ denotes the "start" token and each column $r_i$ is an embedding vector. The positional encoding \cite{gehring2017convolutional} is also considered to make use of the attribute order information $\textbf{P} = (p_0, p_1, ..., p_K)$. The embedding matrix and positional encoding are combined together to obtain the semantic representations on an ordered prediction path $\textbf{E} = (e_0, e_1, ..., e_K)$, where $e_k = r_k + p_k$. 

Finally, the spatial and semantic context can be obtained by: 
\begin{equation}
\textbf{C} = \textbf{E} + (\textbf{U}_s g_s) 
\end{equation}
where the $\textbf{U}$ is learnable projection matrix. 
For the edges, they only connect the $i-$th node with nodes whose subscript $\leq$ $i$ to ensure the prediction of the current attribute only has relations with previously known outputs. The edge weights of connected edges can be computed by: 
\begin{equation}
\textbf{F}_{\hat{e}(\textbf{c}_i, \textbf{c}_j)} = \phi_{\hat{e}} (\textbf{c}_i)^T \phi_{\hat{e}'} (\textbf{c}_j)
\end{equation}
where $\phi_{\hat{e}}(*)$ and $\phi_{\hat{e}'}(*)$ are linear transformation functions. The adjacency matrix $\textbf{A}_{\hat{e}}$ can also be obtained by normalizing the connected edge weights along each row. And the convolution operation on the semantic graph can be computed as: 
\begin{equation}
\textbf{G}_{\hat{e}} = \textbf{A}_{\hat{e}}\textbf{C}^T\textbf{W}_{\hat{e}}
\end{equation}

The output representation $\textbf{G}_{\hat{e}}$ can be obtained after conducting convolutions on the semantic graph, and then utilized for sequential attribute prediction. 

The semantic-to-visual sub-network can also be processed in a similar manner and it also outputs sequential attribute prediction. The output of these two sub-networks is fused as the final prediction and can be trained in an end-to-end way.

\textbf{Summary:} Due to the relations that exist in multiple attributes, many algorithms are proposed to mine such information for PAR. Therefore, the Graphic models are the first to think and introduced into the learning pipeline, such as Markov Random Field \cite{li1994markovMRF}, Conditional Random Field \cite{lafferty2001conditionalRF}, And-Or-Graph \cite{zhu2007stochasticAOG} or Graph Neural Networks \cite{henaff2015deepGCN}.  The works reviewed in this subsection are the outputs via the integration of the Graphic models with PAR. Maybe the other Graphic models can also be used for PAR to achieve better recognition performance. Although these algorithms have so many advantages, however, these algorithms seem more complex than others. The efficiency issues also need to be considered in practical scenarios.

\subsection{Other Algorithms} 
This subsection is used to demonstrate algorithms that are not suitable for the aforementioned categories, including PatchIt \cite{sudowe2016patchit}, FaFS \cite{lu2017fullyafs}, GAM \cite{fabbri2017generative}.

\subsubsection{PatchIt (BMVC-2016) \cite{sudowe2016patchit}}

\begin{figure}[htb]
\center
\includegraphics[width=3.5in]{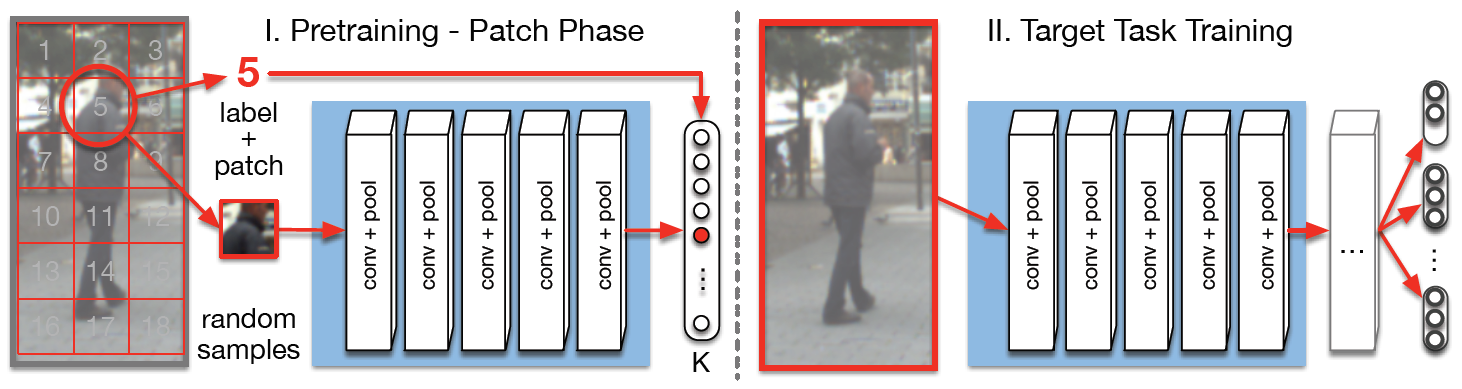}
\caption{The pre-training algorithm introduced from PatchIt \cite{sudowe2016patchit} (left figure) and multi-class classification network for PAR (right figure).}
\label{DeepAlgorithms-PatchIt}
\end{figure}	

Regular ConvNets usually adopt a pre-trained model on an \emph{auxiliary task} for weight initialization. However, it constrains the designed network to as similar to existing architectures, such as AlexNet, VGG, or ResNet. Different from these algorithms, this paper proposes a self-supervised pre-training approach, named PatchTask, to obtain weight initializations for the PAR. Its key insight is to leverage data from the same domain as the target task for pre-training and it only relies on automatically generated rather than human-annotated labels. In addition, it is easier for us to find massive unlabelled data for our task.  

For the PatchTask, the authors define it as a K-class classification problem. As shown in Fig. \ref{DeepAlgorithms-PatchIt}, they first divide the image into multiple non-overlapping local patches, and then, let the network predict the origin of a given patch. They use the PatchTask to obtain an initialization for the convolutional layers of VGG16 and apply it for PAR.

\subsubsection{FaFS (CVPR-2017) \cite{lu2017fullyafs}} 

\begin{figure*}[htb!]
\center
\includegraphics[width=7in]{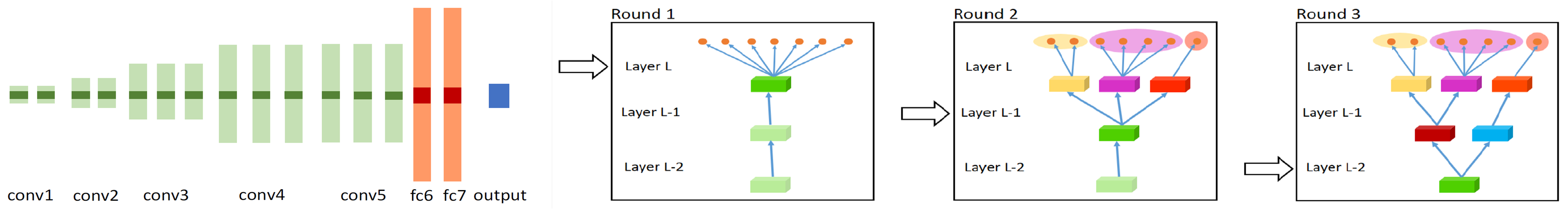}
\caption{(Left) the comparison between the thin model with VGG-16; (Right) Illustration of the widening procedure in \cite{lu2017fullyafs}}
\label{DeepAlgorithms-FaFS}
\end{figure*}	

The target of multi-task learning is to share relevant information across these tasks to help improve final generalization performance. Most hand-designed deep neural networks conduct both shared and task-specific feature learning. Different from existing works, FaFS \cite{lu2017fullyafs} is proposed to design compact multi-task deep learning architecture automatically. This algorithm starts with a thin multi-layer network and dynamically widens it in a greedy manner during training. This will create a tree-like deep architecture by repeating the above widening procedure and similar tasks reside in the same branch until at the top layer. Fig. \ref{DeepAlgorithms-FaFS} (right sub-figure) illustrates this process. Fig. \ref{DeepAlgorithms-FaFS} (left figure) gives a comparison between the thin network and the VGG-16 model. The weight parameters of the thin network are initialized by minimizing the objective function as follows with simultaneous orthogonal matching pursuit (SOMP) \cite{tropp2006algorithmsSOMP}: 
\begin{equation}
A^*, \omega^*(l) = \arg\min_{A\in \mathbb{R}^{d \times d'}, |w| = d'} || W^{p,l} - AW^{p,l}_{w:} ||_F,  
\end{equation} 
where $W^{p,l}$ is the parameters of the pre-trained model at layer $l$ with $d$ rows. $W^{p,l}_{w:}$ denote a truncated weight matrix at only keeps the rows indexed by the set $\omega$. This initialization process is done layer by layer and is applicable for both convolutional and fully connected layers. 

Then, a layer-wise model widening is adopted to widen the thin network, as shown in Fig. \ref{DeepAlgorithms-FaFS} (right sub-figure). This operation is started from the output layer and recursively in a top-down manner towards the lower layers. It is also worth noting that each branch is associated with a subset of tasks. They also separate the similar and dissimilar tasks into different groups according to the probability which is an affinity between a pair of tasks.

\subsubsection{GAM (AVSS-2017) \cite{fabbri2017generative}}

This paper proposes to handle the issue of occlusion and low resolution of pedestrian attributes using deep generative models. Specifically, their overall algorithm contains three sub-networks, \emph{i.e.}, the attribute classification network, the reconstruction network, and the super-resolution network. 

\begin{figure}[htb]
\center
\includegraphics[width=3.5in]{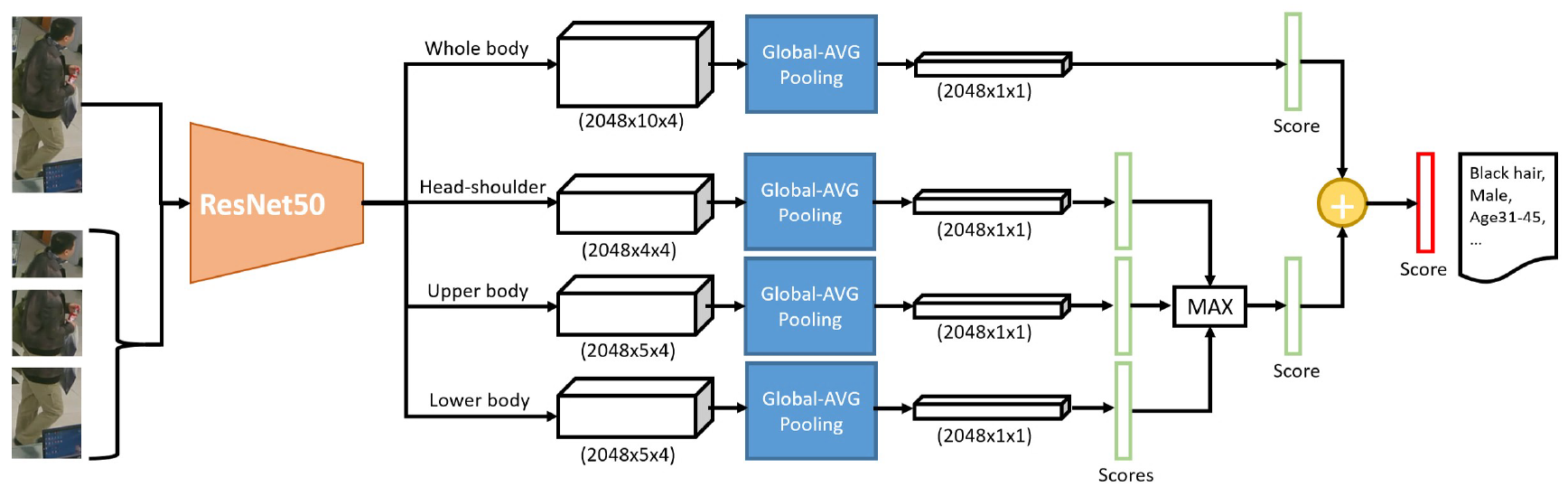}
\caption{The pipeline of GAM-network. This figure is adapted from GAM \cite{fabbri2017generative}.}
\label{DeepAlgorithms-GAM}
\end{figure}

For the attribute classification network, they also adopt joint global and local parts for final attribute estimation, as shown in Fig. \ref{DeepAlgorithms-GAM}. They adopt ResNet50 to extract the deep features and global-average pooling to obtain the corresponding score. These scores are fused as the final attribute prediction score. To handle the occlusion and low-resolution problem, they introduce the deep generative adversarial network \cite{mirza2014conditional} to generate re-constructed and super-resolution images. And use the pre-processed images as input to the multi-label classification network for attribute recognition.

\begin{table*}[htp!]
\center
\newcommand{\tabincell}[2]{\begin{tabular}{@{}#1@{}}#2\end{tabular}}
\caption{A summary of the source code}	
\label{sourceCodeList}
\begin{tabular}{c|c}
\hline
\hline
\textbf{Algorithm}  	&Source Code 	\\
\hline 
DeepMAR \cite{acprli2015DeepMAR}	 &	\url{https://github.com/dangweili/pedestrian-attribute-recognition-pytorch}	\\ 
Wang \emph{et al.} \cite{wang2017multiCVPR} &\url{https://github.com/James-Yip/AttentionImageClass} \\
Zhang \emph{et al.} \cite{li2019richlyRAP2} &\url{https://github.com/dangweili/RAP} \\ 
PatchIt \cite{sudowe2016patchit}   &\url{https://github.com/psudowe/patchit} \\ 
PANDA \cite{zhang2014panda}  &\url{https://github.com/facebookarchive/pose-aligned-deep-networks} \\ 
HydraPlus-Net \cite{liu2017hydraplus}  &\url{https://github.com/xh-liu/HydraPlus-Net}\\ 
WPAL-Net  \cite{zhou2017weakly}  &\url{https://github.com/YangZhou1994/WPAL-network}\\ 
DIAA  \cite{sarafianos2018deep}  &\url{https://github.com/cvcode18/imbalanced_learning}\\ 
\hline
\end{tabular}
\end{table*}

\section{Applications}\label{someApplications}
Visual attributes can be seen as a kind of mid-level feature representation which may provide important information for high-level human-related tasks, such as person re-identification \cite{su2016deepAttributesREID, layne2012towardsAttributeREID, khamis2014jointAttributeREID, li2015clothingAttributesREID, lin2017improvingattributeREID}, pedestrian detection \cite{tian2015pedestrianDetectionAttributes}, person tracking \cite{liu2018stochasticTracking}, person retrieval \cite{wang2013personalRetrivalcolorAtt, feris2014attributePersonSearch}, human action recognition \cite{liu2011recognizingActionAtt}, scene understanding \cite{shao2015deeplySceneUnderstand}. Due to the limited space of this paper, we only review some works in the rest of these subsections.

\begin{figure}[htb]
\center
\includegraphics[width=3.5in]{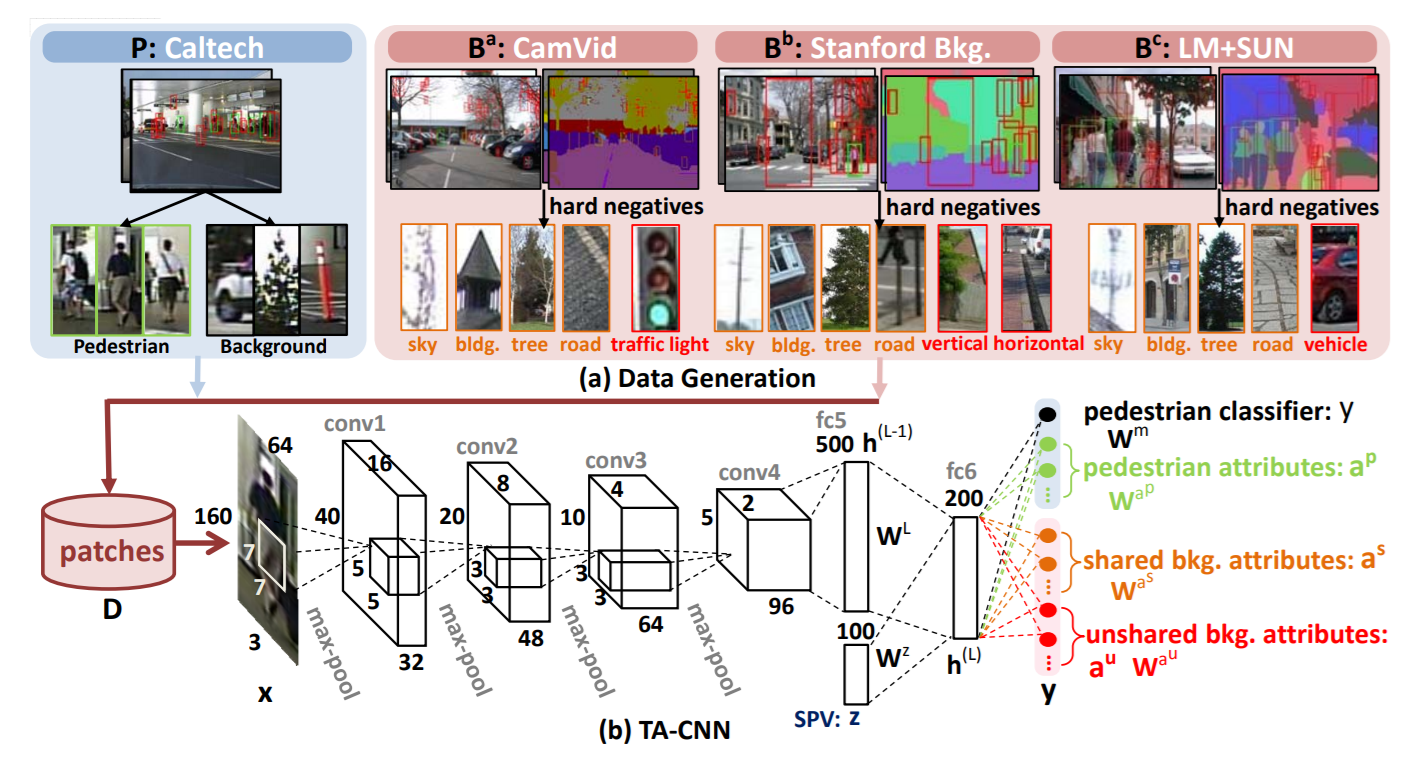}
\caption{The pipeline of pedestrian detection aided by deep learning semantic tasks \cite{tian2015pedestrianDetectionAttributes}.}
\label{Application-detection}
\end{figure}

\textbf{Pedestrian Detection.} Different from regular person detection algorithms which treat it as a single binary classification task, Tian \emph{et al.} propose to jointly optimize person detection with semantic tasks to address the confusion of positive and hard negative samples. They use existing scene segmentation datasets to transfer attribute information to learn high-level features from multiple tasks and dataset sources. Their overall pipeline can be found in Figure \ref{Application-detection}. 

\begin{figure}[htb]
\center
\includegraphics[width=3.5in]{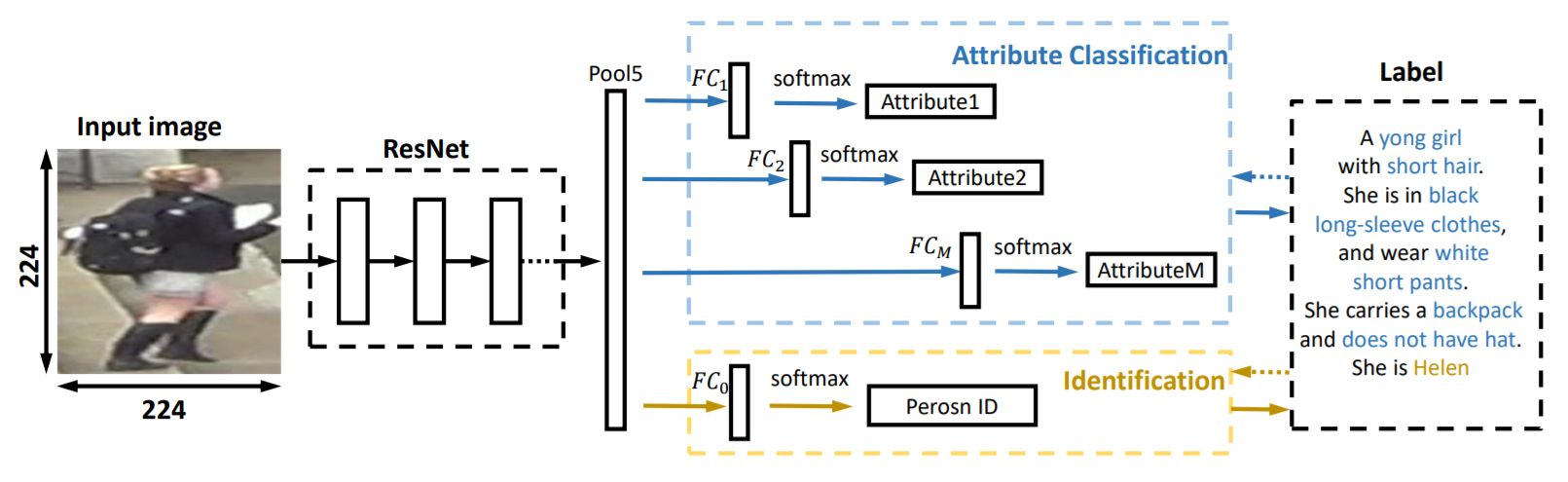}
\caption{The pipeline of APR network in \cite{lin2017improvingattributeREID}.}
\label{Application-REID}
\end{figure}

\textbf{Person Re-identification.} 
As noted in \cite{lin2017improvingattributeREID}, person re-identification and attribute recognition share a common target at the pedestrian description. PAR focuses on local information mine while person re-identification usually captures the global representations of a person. As shown in Figure \ref{Application-REID}, Lin \emph{et al.} propose the multi-task network to estimate the person attributes and person ID simultaneously. Their experiments validated the effectiveness of more discriminative representation learning. 

\begin{figure}[htb]
\center
\includegraphics[width=3.5in]{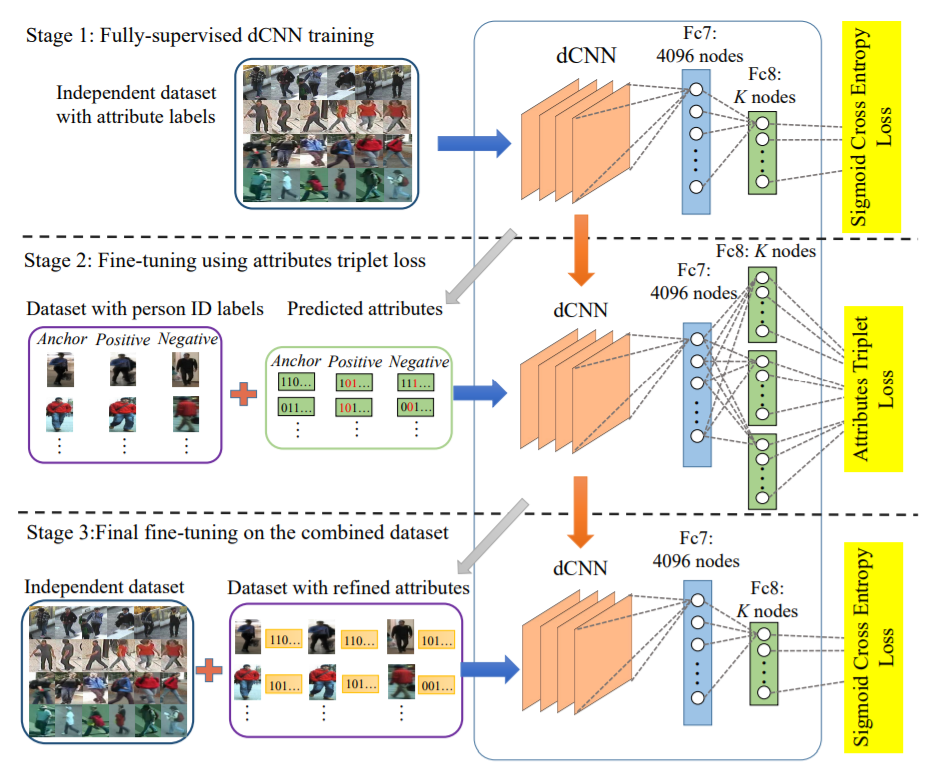}
\caption{The pipeline of Semi-supervised Deep Attribute Learning (SSDAL). \cite{su2016deepAttributesREID}.}
\label{Application-REID2}
\end{figure}

Su \emph{et al.} also propose to integrate the mid-level human attributes into the person re-identification framework in \cite{su2016deepAttributesREID}. They train the attribute model in a semi-supervised manner and mainly contain three stages, as shown in Figure \ref{Application-REID2}. They first pre-train the deep CNN on an independent attribute dataset, then, fine-tuned on another dataset only annotated with person IDs. After that, they estimate attribute labels for the target dataset using the updated deep CNN model. They can achieve good results on multiple-person re-ID datasets using deep attributes with simple Cosine distance. Sameh Khamis \emph{et al.} \cite{khamis2014jointAttributeREID} propose to integrate a semantic aspect into regular appearance-based methods. They jointly learn a discriminative projection to a joint appearance-attribute subspace, which could effectively leverage the interaction between attributes and appearance for matching. Li \emph{et al.} also present a comprehensive study on clothing attributes assisted person re-ID in \cite{li2015clothingAttributesREID}. They first extract the body parts and their local features to alleviate the pose-misalignment issues. Then, they propose a latent SVM-based person re-ID approach to model the relations between low-level part features, middle-level clothing attributes, and high-level re-ID labels of person pairs. They treat the clothing attributes as real-value variables instead of using them as discrete variables to obtain better person-ID performance.

\section{Future Research Directions}	\label{conAndFuture}

In the rest of this section, we discuss several interesting directions for future work of pedestrian attribute recognition. We also list some released source codes of PAR in Table \ref{sourceCodeList}. 

\subsection{More Accurate and Efficient Part Localization Algorithm} 
Human beings can recognize detailed attribute information in a very efficient way because we can focus on specific regions in a glimpse and reason the attribute based on local and global information. Therefore, it is an intuitive idea to design algorithms that can detect the local parts for accurate attribute recognition as we humans do. 

\begin{figure}[htb]
\center
\includegraphics[width=3in]{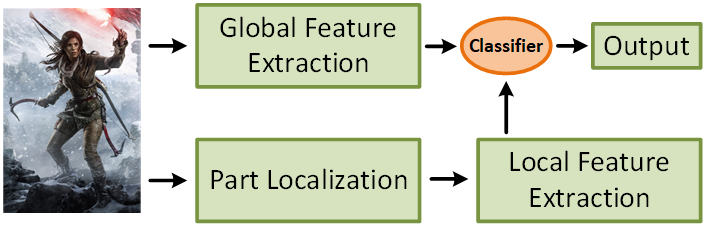}
\caption{The framework of joint local and global feature extraction for person attribute recognition.}
\label{futureworkParts}
\end{figure}

According to section \ref{partbasedPAR}, it is easy to find that researchers are indeed more interested in mining local parts of the human body. They use manually annotated or detected human body or pose information for part localization. The overall framework of part-based attribute recognition can be found in Fig. \ref{futureworkParts}. There are also some algorithms that attempt to propose a unified framework in a weakly supervised manner to jointly handle attribute recognition and localization. We think this will also be a good and useful research direction for pedestrian attribute recognition. 

\subsection{Deep Generative Models for Data Augmentation} 
In recent years, the deep generative models have achieved great progress and many algorithms are proposed, such as: pixel-CNN \cite{van2016conditional}, pixel-RNN \cite{van2016pixelRNN}, VAE \cite{doersch2016tutorial}, GAN \cite{goodfellow2014generative}. Recent works like the progressive GAN \cite{karras2017progressive} and bigGAN \cite{brock2018largebigGAN} even make people feel shocked about the image generated by these algorithms. One intuitive research direction is how can we use deep generative models to handle the issues of low-quality person images or unbalanced data distribution. 

There are already many types of research focusing on image generation with the guidance of text, attribute, or pose information \cite{ma2017poseGAN, qian2018poseGAN, yan2016attribute2image, fu2016texttoimage, zhang2017stackgan}. The generated images can be used in many other tasks for data augmentation, for example, object detection \cite{wang2017afastrcnn}, person re-identification \cite{zheng2017unlabeledREIDgan}, visual tracking \cite{wang2018sint++}, \emph{et al.} GAM \cite{fabbri2017generative} also attempt to generate high-resolution images for person attributes recognition. It is also worth designing new algorithms to generate pedestrian images according to given attributes to augment the training data.

\subsection{Further Explore the Visual Attention Mechanism} 

Visual attention has drawn more and more researchers' attention in recent years \cite{mnih2014recurrent}. It is still one of the most popular techniques used nowadays and is integrated with every kind of deep neural network in many tasks. Just as noted in \cite{mnih2014recurrent}, one important property of human perception is that one does not tend to process a whole scene in its entirety at once. Instead, humans focus attention selectively on parts of the visual space to acquire information when and where it is needed and combine information from different fixations over time to build up an internal representation of the scene \cite{rensink2000dynamic}, guiding future eye movements and decision making. It also substantially reduces the task complexity as the object of interest can be placed in the center of the fixation and irrelevant features of the visual environment (``clutter") outside the fixated region are naturally ignored. 

Many existing attention-based pedestrian attribute recognition algorithms focus on feature or task weighting using a trainable neural network.
Although it indeed improved the overall recognition performance, however, how to accurately and efficiently locate the attention regions is still an open research problem. Designing novel attention mechanisms or borrow from other research domains, such as NLP (natural language processing), for pedestrian attribute recognition will be an important research direction in the future.

\subsection{New Designed Loss Functions} 
In recent years, there are many loss functions have been proposed for deep neural network optimization, such as (Weighted) Entropy Loss, Contrastive Loss, Center Loss, Triplet Loss, and Focal Loss. Researchers also design new loss functions for the PAR, such as WPAL \cite{zhou2017weakly}, AWMT \cite{mmhe2017adaptively}, to further improve their recognition performance. It is a very important direction to study the influence of different loss functions for PAR.

\subsection{Explore More Advanced Network Architecture} 
Existing PAR models adopt off-the-shelf pre-trained networks on large-scale datasets (such as ImageNet), as their backbone network architecture. Seldom of them consider the unique characteristics of PAR and design novel networks. Some novel networks have been proposed in recent years, such as capsule network \cite{sabour2017dynamic} \cite{hinton2018matrix}, and External Memory Network \cite{graves2016hybrid}. However, there are still no attempts to use such networks for PAR. There are also works \cite{Esube2019DBAPAR} demonstrate that the deeper the network architecture the better recognition performance we can obtain. Nowadays, Automatic Machine Learning solutions (AutoML) draw more and more attention \cite{liu2019AutoDeepLab} \cite{elsken2018neural} \cite{he2018amc} and many development tools are also released for development, such as AutoWEKA \cite{ThoHutHooLey13AutoWEKA}, Auto-sklearn \cite{NIPS2015_5872}. Therefore, it will be a good choice to design specific networks for person attribute recognition in future works with the aforementioned approaches.

\subsection{Prior Knowledge guided Learning} 
Different from regular classification task, pedestrian attribute recognition always has its own characteristics due to the preference of human beings or natural constraints. It is an important research direction to mining the prior or common knowledge for the PAR. For example, we wear different clothes in various seasons, temperatures or occasions. On the other hand, some researchers attempt to use history knowledge (such as Wikipedia \footnote{\url{en.wikipedia.org}}) to help improve their overall performance, such as image caption \cite{Wu2017ImageCaptionKnledge, Peng2015ExplicitKnowVQA}, object detection \cite{Jiang2018knowledgeObjectDetection}. Therefore, how to use this information to explore the relations between personal attributes or help the machine learning model to further understand the attributes is still an unstudied problem.

\subsection{Multi-modal Pedestrian Attribute Recognition} 
Although existing single-modal algorithms already achieve good performance on some benchmark datasets as mentioned above. However, as is known to all, the RGB image is sensitive to illumination, bad weather (such as rain, snow, fog), night time, \emph{et al}. It seems impossible for us to achieve accurate pedestrian attribute recognition all day and in all weather. However, the actual requirement of intelligent surveillance needs far more than this target. How can we bridge this gap?

\begin{figure}[htb]
\center
\includegraphics[width=3.5in]{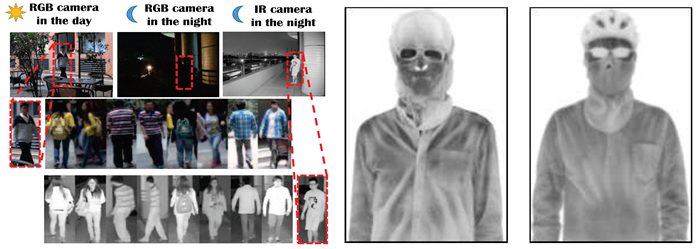}
\caption{The example RGB and thermal infrared images adopted from \cite{wu2017rgbtreid, kresnaraman2016human} .}
\label{futurework-RGBT}
\end{figure}

One intuitive idea is to mine useful information from other modalities, such as thermal or depth sensors, to integrate with RGB sensors. There are already many works that attempt to fuse these multi-modal data and improve their final performance significantly, such as RGB-Thermal tracking \cite{li2017grayscale, li2016learning, li2018rgbtnewbenchmark}, moving object detection \cite{li2017weighted}, person re-identification \cite{wu2017rgbtreid}, RGB-Depth object detection \cite{gupta2014learning, song2016deep}, segmentation \cite{silberman2012indoor}.  We think the idea of multi-modal fusion could also help improve the robustness of pedestrian attribute recognition. As shown in Fig. \ref{futurework-RGBT}, these thermal images can highlight the contour of humans and some other wearing or carrying objects as denoted in \cite{kresnaraman2016human, kresnaraman2017headgear}.

\subsection{Video-based Pedestrian Attribute Recognition} 
Existing pedestrian attribute recognition is based on a single image, however, in practical scenarios we often obtain the video sequence captured by cameras. Although running the existing algorithm on each video frame can be an intuitive and easy strategy, efficiency maybe the bottleneck for practical applications. Chen \emph{et al.} propose a video-based PAR dataset \cite{bourdev2019videoPAR} by re-annotating the MAR dataset \cite{zheng2016mars} which is originally constructed for video-based person re-identification. Generally speaking, image-based attribute recognition can only make use of the spatial information from the given image, which increases the difficulty of PAR due to the limited information. In contrast, given the video-based PAR, we can jointly utilize the spatial and temporal information. The benefits can be listed as follows: 1). we can extend the attribute recognition into a more general case by defining more dynamic person attributes, such as ``running man'';  2). The motion information can be used to reason the attributes that may be hard to recognize in a single image; 3). The general person attributes learned in videos can provide more helpful information for other video-based tasks, such as video caption, and video object detection. Therefore, how to recognize human attributes in practical video sequences efficiently and accurately is a worthy studying problem.

\subsection{Joint Learning of Attribute and Other Tasks} 
Integrating the person attribute learning into the pipeline of other person-related tasks is also an interesting and important research direction. There are already many algorithms proposed by consider the person attributes into corresponding tasks, such as: attribute-based pedestrian detection \cite{tian2015pedestrian}, visual tracking \cite{danelljan2014adaptiveColorAtt}, person re-identification \cite{su2016deepreid, layne2012towardsattri, li2015clothingreID, liu2018sequenceJCM} and social activity analysis \cite{fu2012attributesocalA}. In future works, how to better explore the fine-grained person attributes for other tasks and also use other tasks for better human attribute recognition is an important research direction.

\section{Conclusion}\label{conclusions} 
In this paper, we give a review of pedestrian attribute recognition (\emph{i.e.} PAR) from traditional approaches to deep learning based algorithms in recent years. To the best of our knowledge, this is the first review paper on pedestrian attribute recognition. Specifically speaking, we first introduce the background information (the definition and challenging factors) about PAR. Then, we list existing benchmarks proposed for PAR, including popular datasets and evaluation criteria. After that, we review the algorithms that may used for PAR from two aspects, \emph{i.e.} the multi-task learning and multi-label learning. Then, we give a brief review of PAR algorithms, we first review some popular neural networks which has been widely used in many other tasks; then, we analyze the deep algorithms for PAR from different views, including global-based, part-based, visual-attention-based, sequential prediction based, new designed loss function based, curriculum learning based, graphic model-based and other algorithms. Then, we give a short introduction about the previous works on combining person-attribute learning and other human-related tasks. Finally, we conclude this survey paper and point out some possible research directions from nine aspects of the PAR.

\ifCLASSOPTIONcaptionsoff
  \newpage
\fi

{
\bibliographystyle{IEEEtran}
\bibliography{reference}
}

\vspace{-0.5cm}
\begin{IEEEbiography} [{\includegraphics[width=1in,height=1.25in,clip,keepaspectratio]{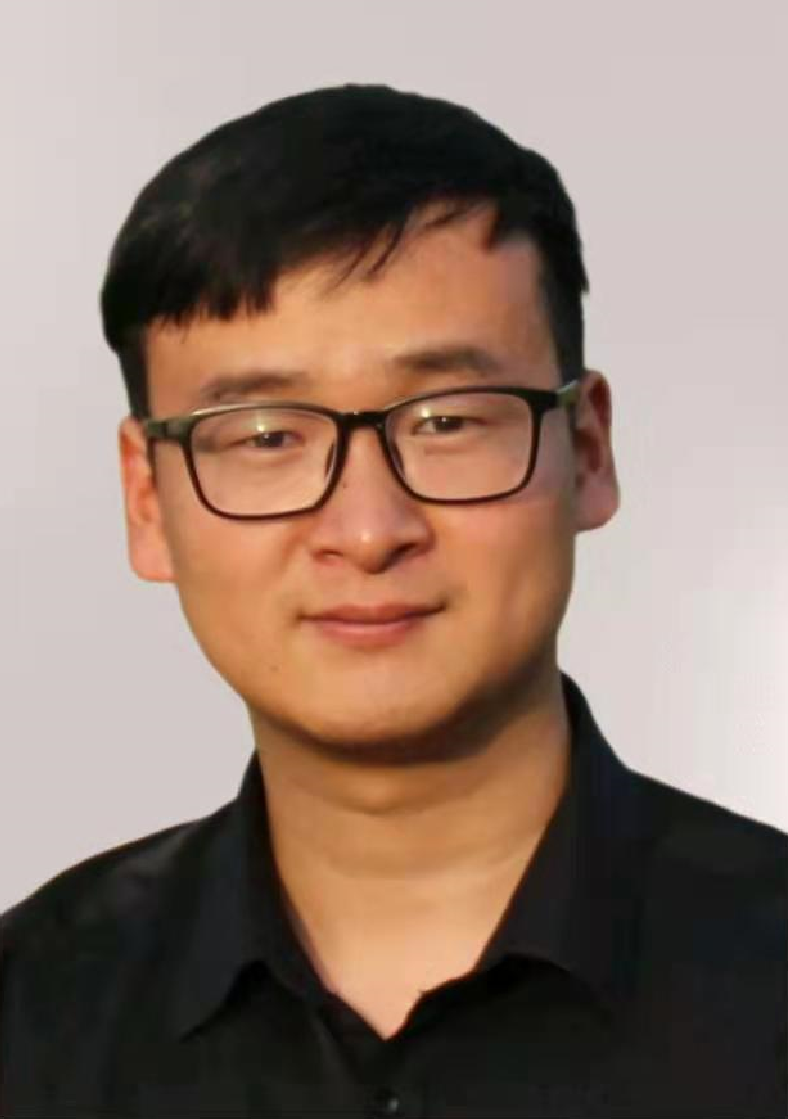}}]
{Xiao Wang} (Member, IEEE) received the B.S. degree in West Anhui University, Luan, China, in 2013. He received the Ph.D. degree in computer science in Anhui University, Hefei, China, in 2019. From 2015 and 2016, he was a visiting student with the School of Data and Computer Science, Sun Yat-sen University, Guangzhou, China. He also has a visiting at UBTECH Sydney Artificial Intelligence Centre, the Faculty of Engineering, the University of Sydney, in 2019. He finished the postdoc research in Peng Cheng Laboratory, Shenzhen, China, from April, 2020 to April, 2022. He is now an Associate Professor at School of Computer Science and Technology, Anhui University, Hefei, China. His current research interests mainly about Computer Vision, Event-based Vision, Machine Learning, and Pattern Recognition. He serves as a reviewer for a number of journals and conferences such as IEEE TCSVT, TIP, IJCV, CVIU, PR, CVPR, ICCV, AAAI, ECCV, ACCV, ACM-MM, and WACV. \textbf{Homepage}: \url{https://wangxiao5791509.github.io/}
\end{IEEEbiography}

\begin{IEEEbiography} 
[{\includegraphics[width=1in,height=1.25in,clip,keepaspectratio]{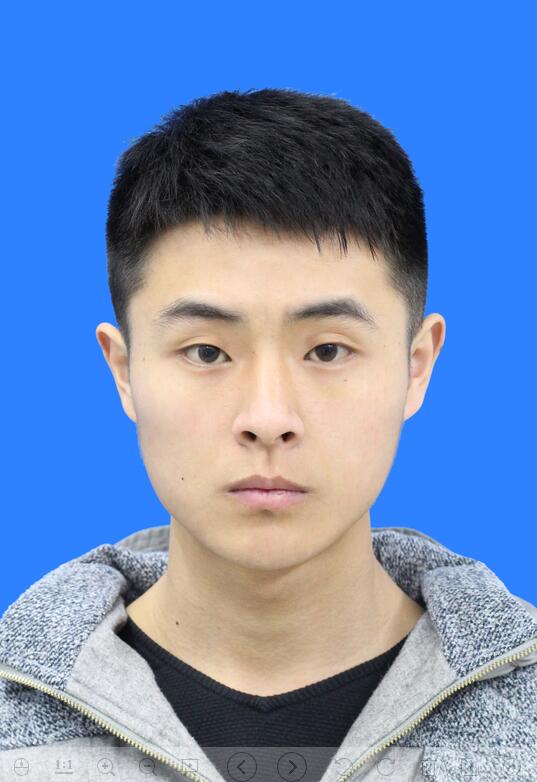}}]
{Shaofei Zheng} received the B.S. degree in Anhui Polytechnic University, Wuhu, China, in 2015. He is currently pursuing the Master degree in computer science in Anhui University. His current research interests mainly about computer vision and machine learning.
\end{IEEEbiography}

\begin{IEEEbiography}
[{\includegraphics[width=1in,height=1.25in,clip,keepaspectratio]{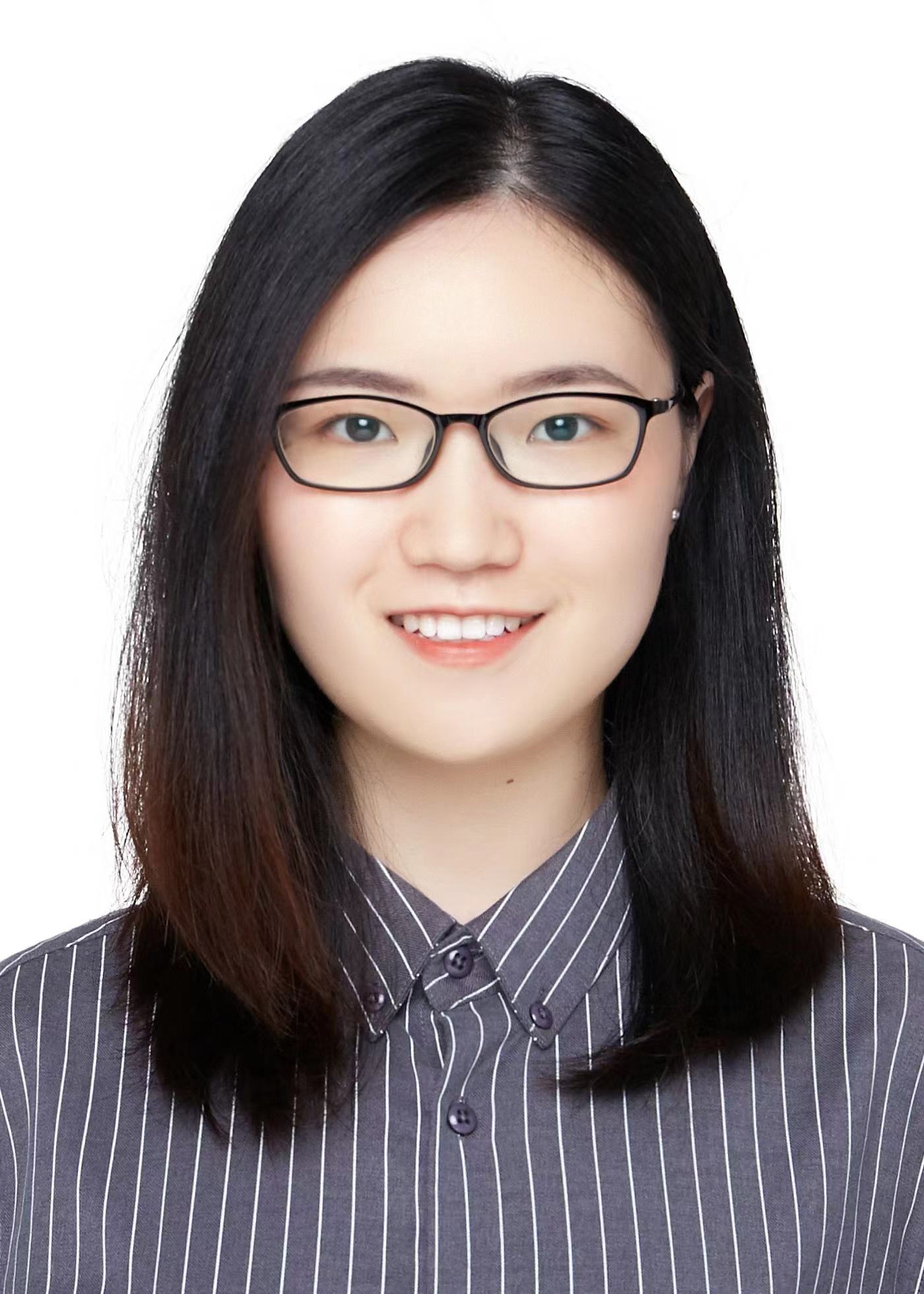}}]
{Rui Yang} received the B.S. degree in Anhui University of Technology, Ma'anshan, China, in 2018. She is currently pursuing the Master degree in computer science in Anhui University. Her current research interests mainly about computer vision and machine learning.
\end{IEEEbiography}

\begin{IEEEbiography} [{\includegraphics[width=1in,height=1.25in,clip,keepaspectratio]{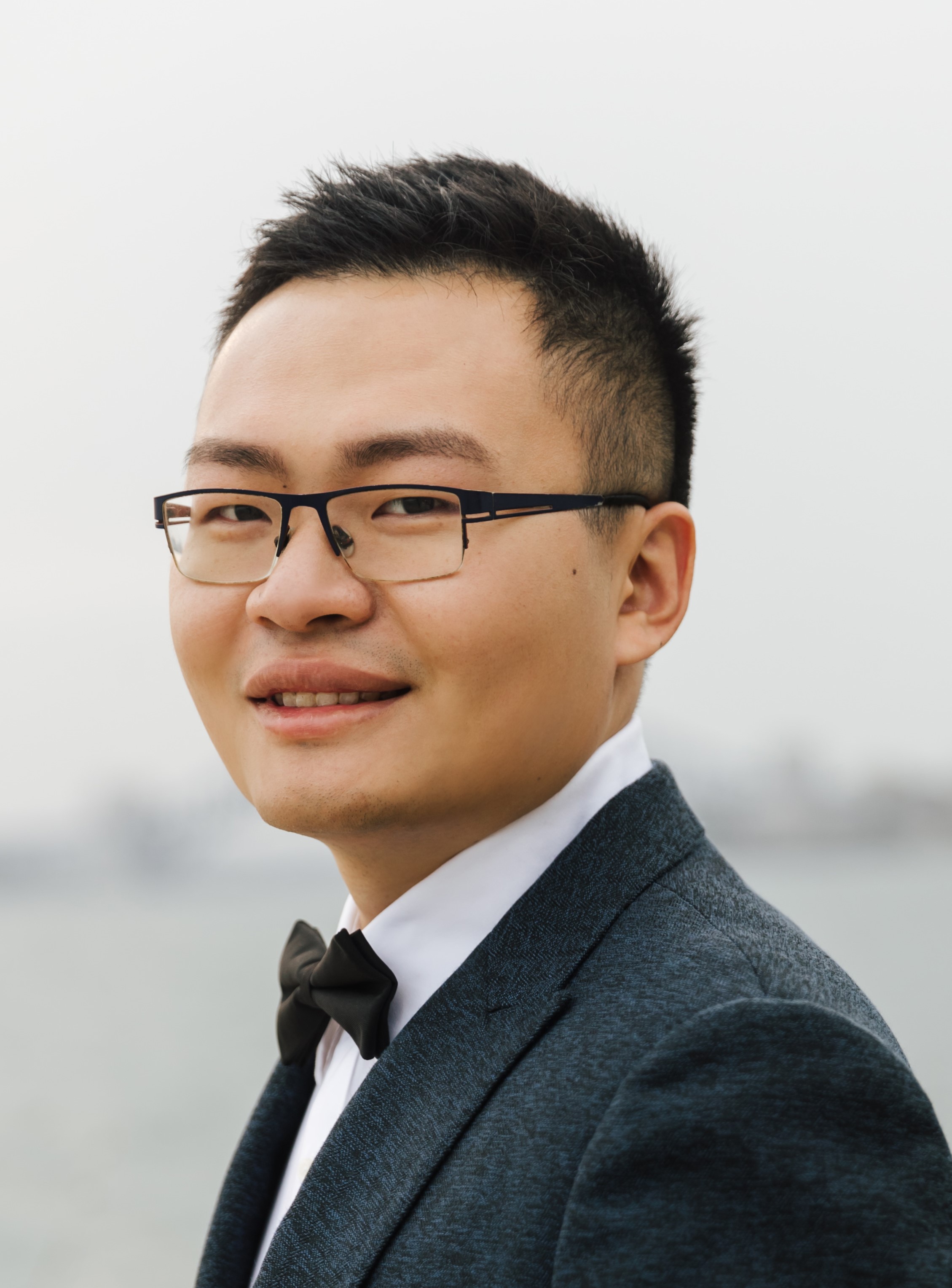}}]
{Zhe Chen} (Member, IEEE) received the B.S. degree in Computer Science from University of Science and Technology of China in 2014 and then received the Ph.D. degree from the University of Sydney in 2019. His research interests include object detection, computer vision applications, and deep learning. His studies were published in high-quality conferences and journals like CVPR, ICCV, ECCV, IJCAI, AAAI, TIP, IJCV, and so on. He has received more then 3300 citations on the Google scholar. He also serves as a reviewer for a number of top journals like T-PAMI, TIP, TCSVT, T-CYB, etc.
\end{IEEEbiography}

\begin{IEEEbiography} 
[{\includegraphics[width=1in,height=1.25in,clip,keepaspectratio]{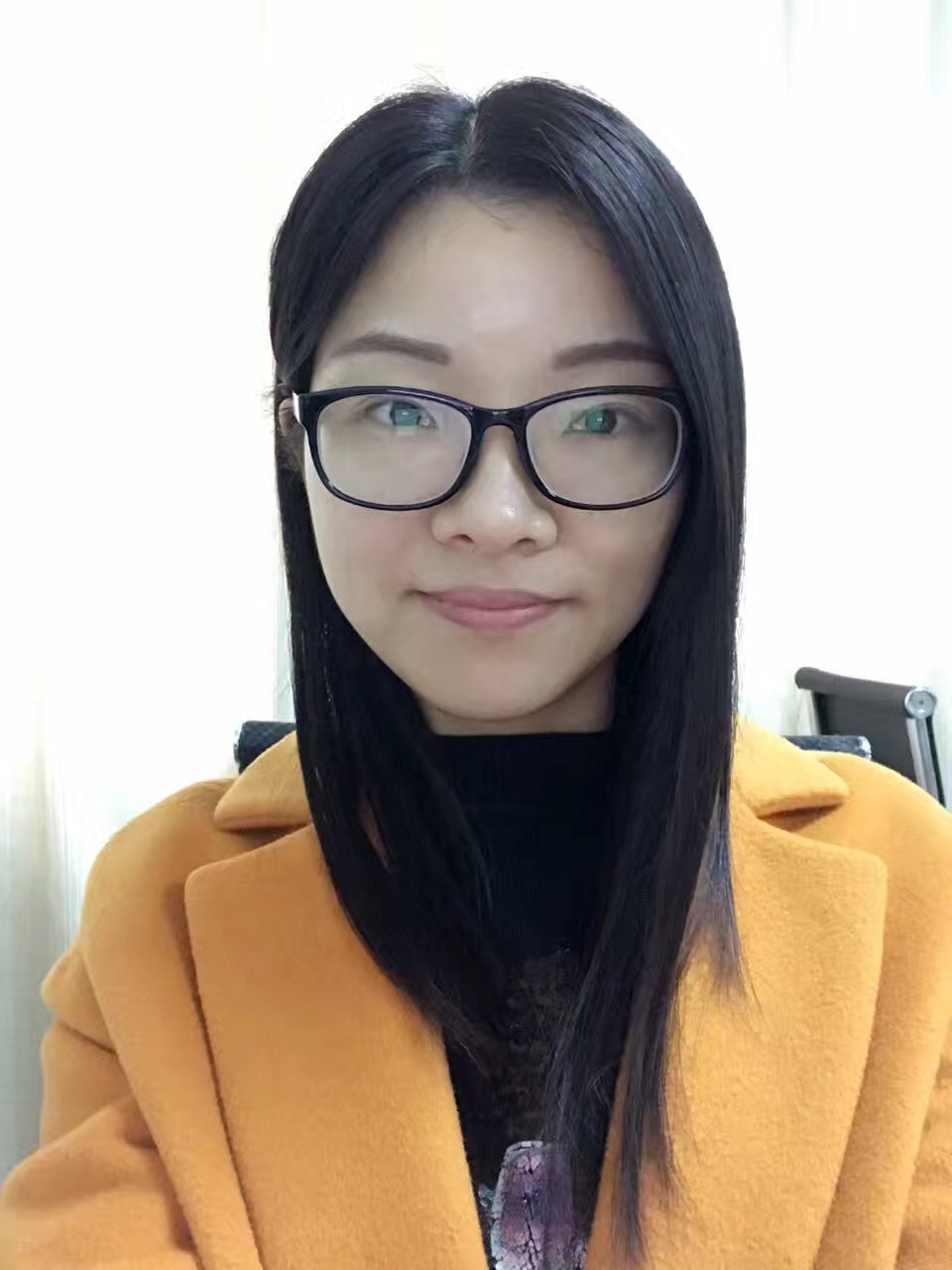}}]
{Aihua Zheng} received her B. Eng. degrees and finished her Master-Doctor combined program in computer science and technology from Anhui University of China in 2006 and 2008, respectively. And received her Ph.D. degree in computer science from the University of Greenwich UK in 2012. She is currently a Lecturer in Anhui University. Her main research areas are visual-based signal processing and pattern recognition. 
\end{IEEEbiography}

\begin{IEEEbiography}
[{\includegraphics[width=1in,height=1.25in,clip,keepaspectratio]{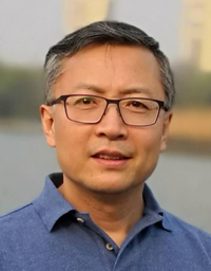}}]
{Bin Luo} received the B.Eng. degree in electronics and the M.Eng. degree in computer science from Anhui University, Hefei, China, in 1984 and 1991, respectively, and the Ph.D. degree in computer science from the University of York, York, U.K., in 2002. He has authored over 200 papers in journals, edited books, and refereed conferences. He is currently a Professor with Anhui University. He also chairs the IEEE Hefei Subsection. He served as a peer reviewer of international academic journals, such as the IEEE TRANSACTIONS ON PATTERN ANALYSIS AND MACHINE INTELLIGENCE, Pattern Recognition, Pattern Recognition Letters, the International Journal of Pattern Recognition and Artificial Intelligence, Knowledge and Information Systems, and Neurocomputing. His current research interests include random graph-based pattern recognition, image and graph matching, graph spectral analysis, and video analysis.
\end{IEEEbiography}

\begin{IEEEbiography}
[{\includegraphics[width=1in,height=1.25in,clip,keepaspectratio]{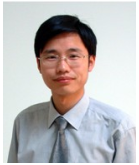}}]
{Jin Tang}received the B.Eng. Degree in automation and the Ph.D. degree in computer science from Anhui University, Hefei, China, in 1999 and 2007, respectively. He is currently a Professor with the School of Computer Science and Technology, Anhui University. His current research interests include computer vision, pattern recognition, and machine learning.
\end{IEEEbiography}

\end{document}